\pdfoutput=1
\documentclass[11pt]{article}

\usepackage[final]{acl}
\usepackage{times}
\usepackage{latexsym}

\usepackage[T1]{fontenc}
\usepackage[utf8]{inputenc}

\usepackage{microtype}

\usepackage{inconsolata}

\usepackage{graphicx}
\usepackage{amsmath}
\usepackage{enumerate}
\usepackage{array}
\usepackage{booktabs}
\usepackage{multirow}
\usepackage{subcaption}
\usepackage{tcolorbox}
\usepackage{listings}
\usepackage{amssymb}
\usepackage{amsthm}
\usepackage{tikz}
\usepackage{algorithm}
\usepackage{bm}
\usepackage{tabularx}
\usepackage{newfloat}
\usepackage{longtable}
\usepackage{subcaption}
\usepackage{colortbl}
\usepackage{xcolor}
\usepackage{algpseudocode}
\usepackage{xifthen}

\title{FRAME: Boosting LLMs with \\A Four-Quadrant Multi-Stage Pretraining Strategy}

\author{
    Xuemiao Zhang\textsuperscript{1,4}$^{\ast}$, \ 
    Feiyu Duan\textsuperscript{2,4}$^{\ast}$, \ 
    Liangyu Xu\textsuperscript{4}\thanks{Equal contribution.}, \ 
    \textbf{Yongwei Zhou\textsuperscript{4}}, \\ \ 
    \textbf{Sirui Wang\textsuperscript{3,4}\thanks{Corresponding author.}}, \ 
    \textbf{Rongxiang Weng\textsuperscript{4}}, \
    \textbf{Jingang Wang\textsuperscript{4}}, \ 
    \textbf{Xunliang Cai\textsuperscript{4}}
    \\
    \textsuperscript{1} Peking University\quad
    \textsuperscript{2} Beihang University\quad
    \textsuperscript{3} Tsinghua University\quad
    \textsuperscript{4} Meituan \\
    \texttt{zhangxuemiao@pku.edu.cn}\quad 
    \texttt{duanfeiyu@buaa.edu.cn} \quad \texttt{ywzhouphd2018@gmail.com}\\
    \texttt{\{xuliangyu02, wangsirui, wangjingang02, caixunliang\}@meituan.com} \\
  } 

\begin{document}
\maketitle
\begin{abstract}
Large language models (LLMs) have significantly advanced human language understanding and generation, with pretraining data quality and organization being crucial to their performance. Multi-stage pretraining is a promising approach, but existing methods often lack quantitative criteria for data partitioning and instead rely on intuitive heuristics. In this paper, we propose the novel \textbf{F}our-quad\textbf{RA}nt \textbf{M}ulti-stage pr\textbf{E}training strategy (FRAME), guided by the established principle of organizing the pretraining process into four stages to achieve significant loss reductions four times. This principle is grounded in two key findings: first, training on high Perplexity (PPL) data followed by low PPL data, and second, training on low PPL difference (PD) data followed by high PD data, both causing the loss to drop significantly twice and performance enhancements. By partitioning data into four quadrants and strategically organizing them, FRAME achieves a remarkable \textbf{16.8\%} average improvement over random across MMLU and CMMLU for the 3B model, effectively boosting LLM performance.
\end{abstract}

\section{Introduction}
% Large language models (LLMs) have achieved significant success in advancing the understanding and generation of human language \cite{touvron2023llama,dubey2024llama,islam2024gpt}. The quality and organization of pretraining data are crucial as they directly impact the performance of LLMs. Studies have revealed that these models require different data at various stages of the pretraining process \cite{yu2024mates}, which provides important insights for developing more efficient pretraining data organization strategies.
LLMs have significantly advanced human language understanding and generation \cite{touvron2023llama,dubey2024llama,islam2024gpt}. The quality and organization of pretraining data are crucial as they directly impact LLM performance. Studies show that LLMs require different data at various pretraining stages \cite{yu2024mates}, offering key insights for developing more efficient pretraining data organization strategies.

Recent studies partition the pretraining process into multiple stages, allowing models to learn from data with distinct characteristics at each stage, which enhances pretraining efficiency\cite{liu2021multistagepretrainingsimplifiedmultimodal,yildiz2024investigating,anonymous2025multiagent}. For instance, MSP introduces specific tasks and data at each phase, allowing models to learn language structures and semantics from simple to complex progressively \cite{liu2021multistagepretrainingsimplifiedmultimodal}. However, these methods don't provide quantitative criteria for partitioning data across stages, often relying on intuitive heuristics. The limitation underscores the need for more systematic approaches to optimize multi-stage pretraining.
\begin{figure}[t]
    \centering
    \begin{subfigure}{0.49\linewidth}
        \centering
        \includegraphics[width=\linewidth]{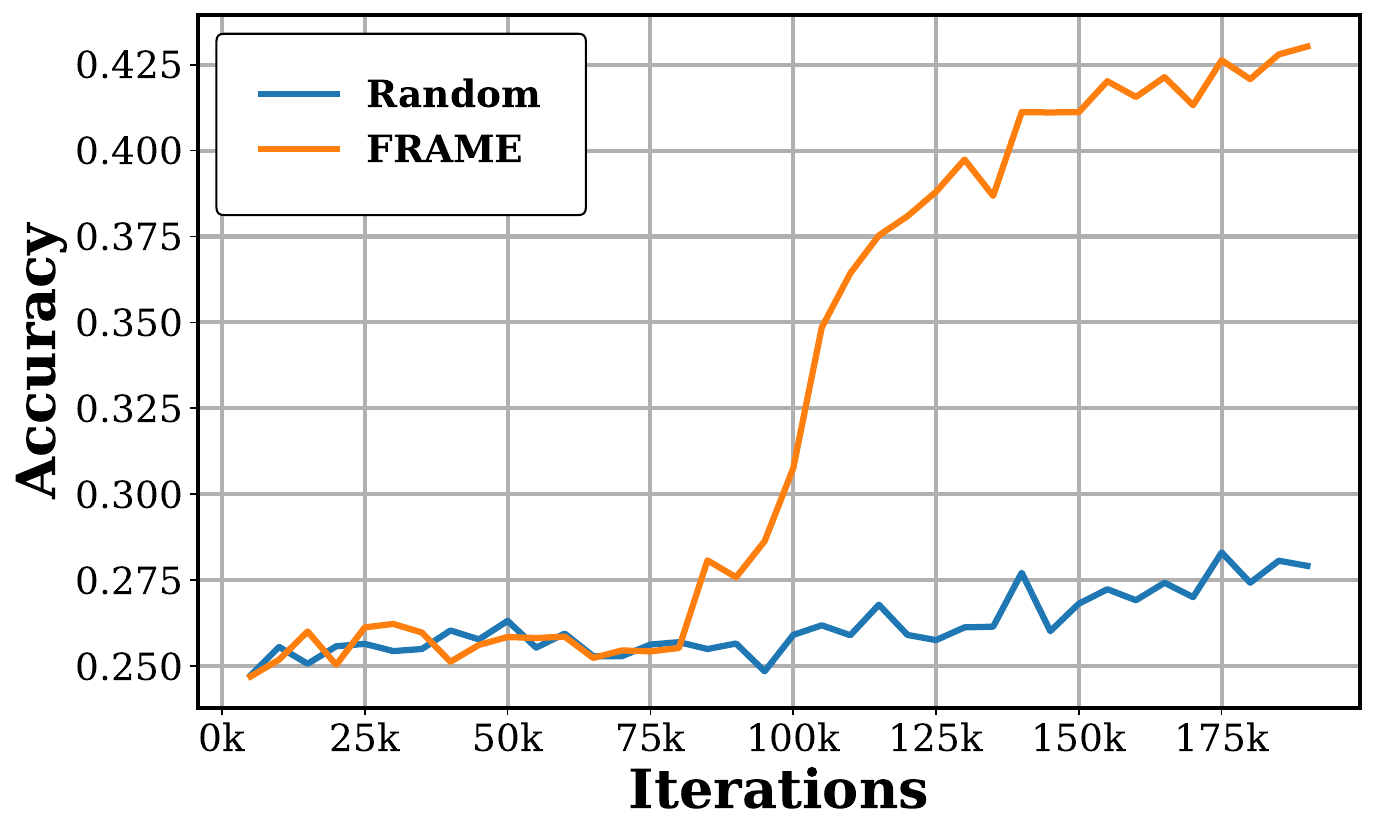}
        \caption{Accuracy on MMLU.}
        \label{fig:arc-e}
    \end{subfigure}
    \hfill
    \begin{subfigure}{0.49\linewidth}
        \centering
        \includegraphics[width=\linewidth]{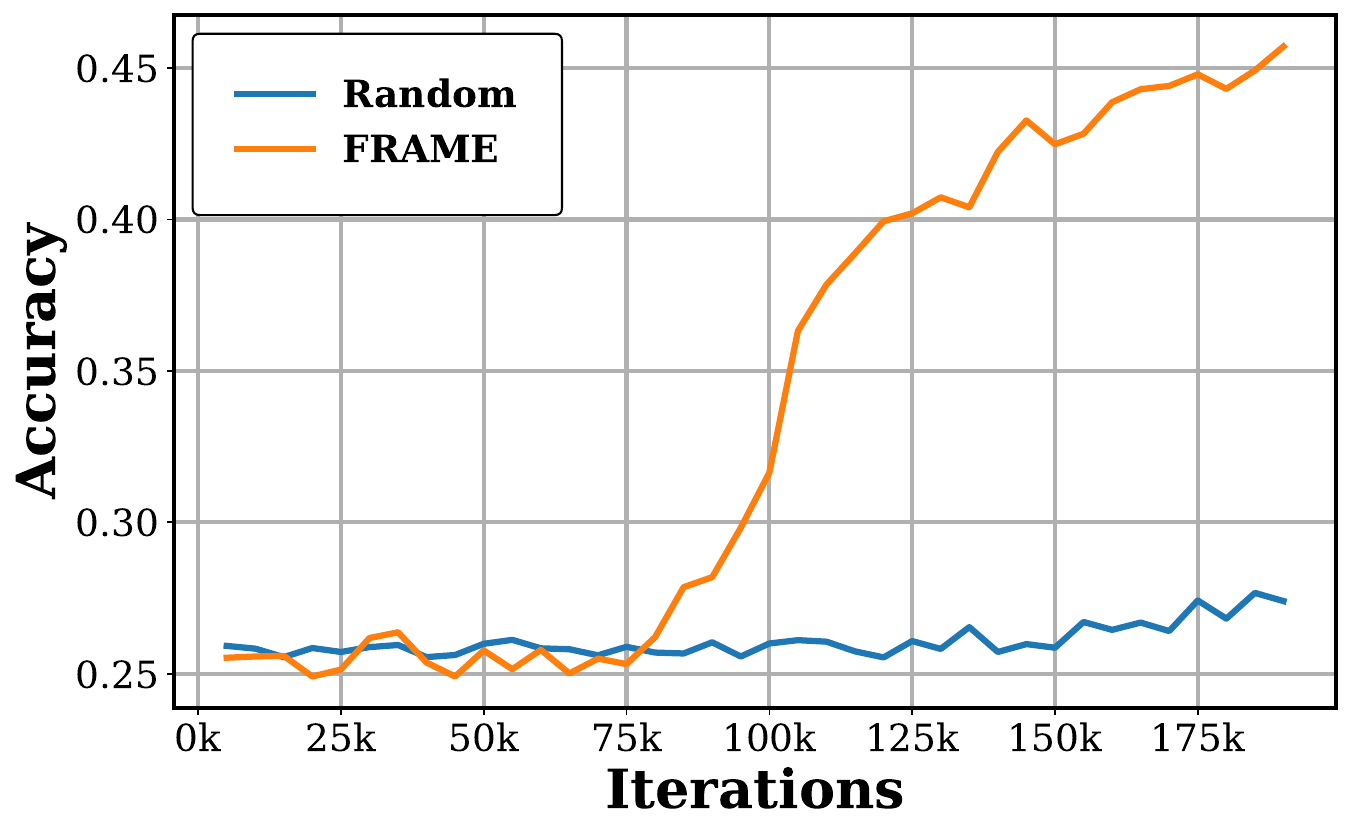}
        \caption{Accuracy on CMMLU.}
        \label{fig:arc-c}
    \end{subfigure}
    \\
    \begin{subfigure}{0.49\linewidth}
        \centering
        \includegraphics[width=\linewidth]{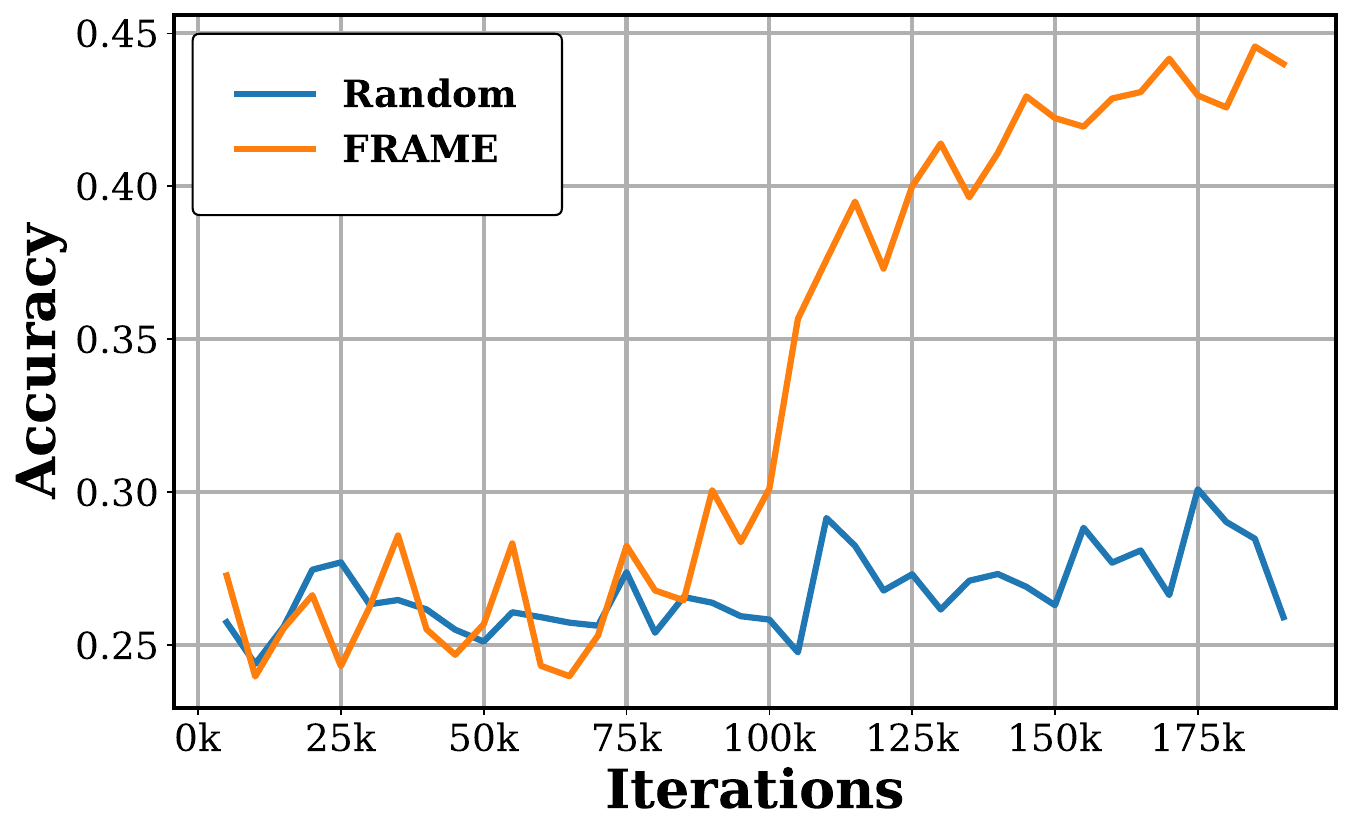}
        \caption{Accuracy on CEVAL.}
        \label{fig:sciq}
    \end{subfigure} 
    \hfill
    \begin{subfigure}{0.49\linewidth}
        \centering
        \includegraphics[width=\linewidth]{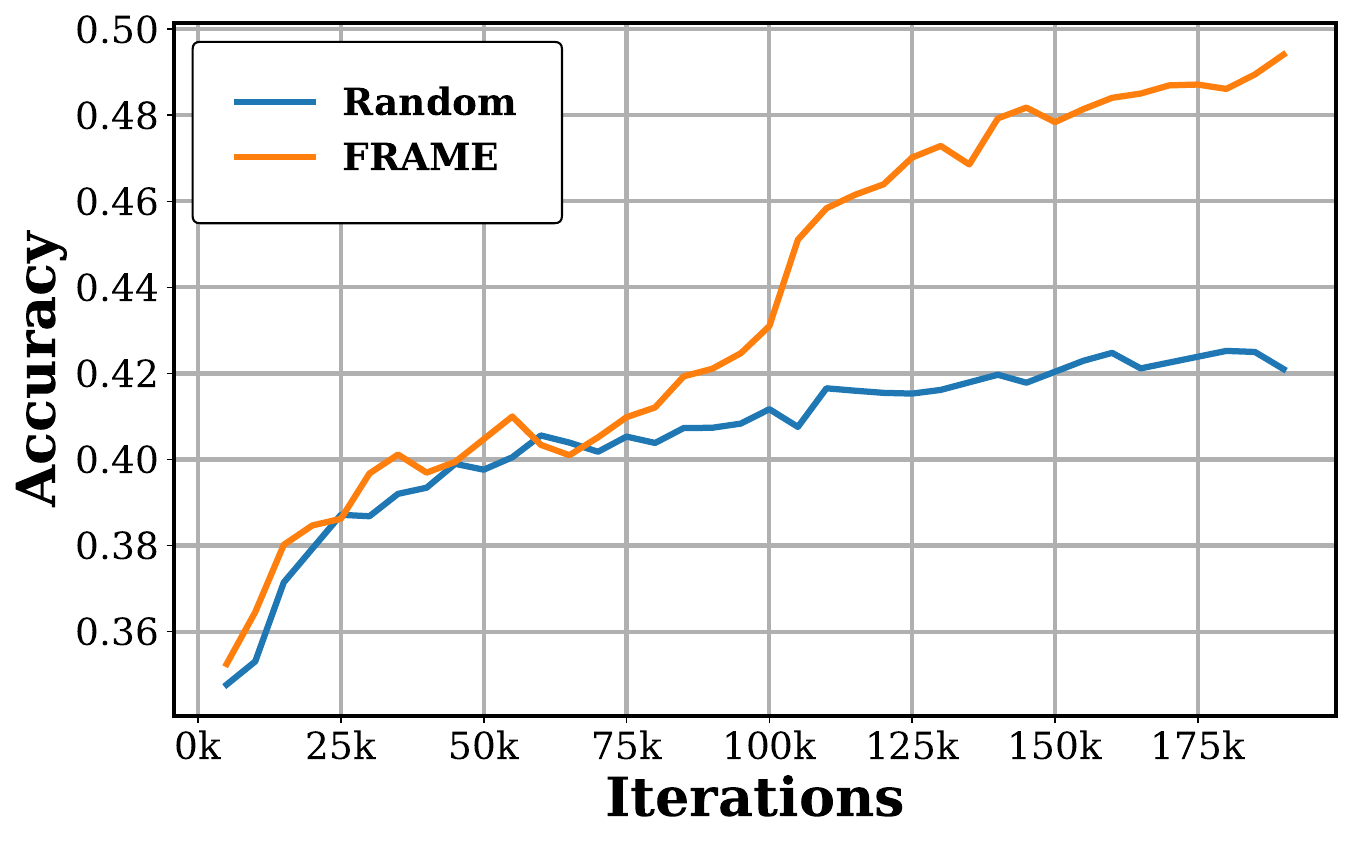}
        \caption{Accuracy on average.}
        \label{fig:piqa}
    \end{subfigure}
    \caption{Few-shot downstream performance on various benchmarks with respect to pretraining iterations, for 3B models trained on 1T tokens. The average performance is based on 8 downstream tasks. FRAME achieves 15.3\% improvement over Random across MMLU, and 18.2\% on CMMLU.}
    \label{fig:AVG}
\end{figure}

In this paper, we propose a novel \textbf{F}our-quad\textbf{RA}nt \textbf{M}ulti-stage pr\textbf{E}training strategy (FRAME) to boost LLMs' performance. This strategy is systematically guided by the principle of organizing the pretraining process into four stages to achieve significant loss drops four times. The principle is derived from two key findings based on quantifiable data metrics.

% Specifically, we intuitively use the metric Perplexity (PPL) to partition the data into two parts: high PPL and low PPL. Our first key finding reveals that training on high PPL data first, followed by low PPL data, results in the loss dropping significantly twice and boosts model performance. Inspired by PDPC \cite{zhang2025preferencecurriculumllmspretrained}, we also introduce the PPL difference (PD) between strong and weak models as another metric for data partitioning. Our second key finding reveals that training on low PD data first, followed by high PD data, results in pretraining loss significantly dropping twice, thereby boosting model performance.
Specifically, we intuitively use the metric Perplexity (PPL) to partition the data into two parts: high PPL and low PPL. Our first key finding reveals that training on high PPL data first, followed by low PPL data, leads to the loss dropping significantly twice and boosts model performance. Inspired by PDPC \cite{zhang2025preferencecurriculumllmspretrained}, we also introduce the PPL difference (PD) between strong and weak models as another metric for data partitioning. Our second key finding shows that training on low PD data first, followed by high PD data, similarly causes the pretraining loss to drop significantly twice, thereby boosting model performance.

Based on these two major findings, we establish the principle of \textbf{organizing the pretraining process into four stages to achieve significant loss reductions four times}. Specifically, we partition the data into four quadrants based on PPL and PD: Quadrant 1 ($Q_1$) with low PD and low PPL, Quadrant 2 ($Q_2$) with high PD and low PPL, Quadrant 3 ($Q_3$) with low PD and high PPL, and Quadrant 4 ($Q_4$) with high PD and high PPL. Through further analysis and experiments, we determine that the optimal strategy is to reorganize the pretraining data in the $Q3\to Q4\to Q1\to Q2$ sequence. To ensure smooth data transitions between stages, we also implement a smoothing process for gradual stage transitions. All data is processed offline to ensure the continuity of model training is not disrupted. Practical evidence shows that FRAME results in loss dropping significantly four times, enhances the models' emergent abilities, and boosts their performance. As shown in Figure \ref{fig:AVG}, FRAME achieves a significant average improvement of 15.3\% on MMLU and 18.2\% on CMMLU over random selection, respectively.

\begin{figure}[t]
    \centering
    \includegraphics[width=1.0\linewidth]{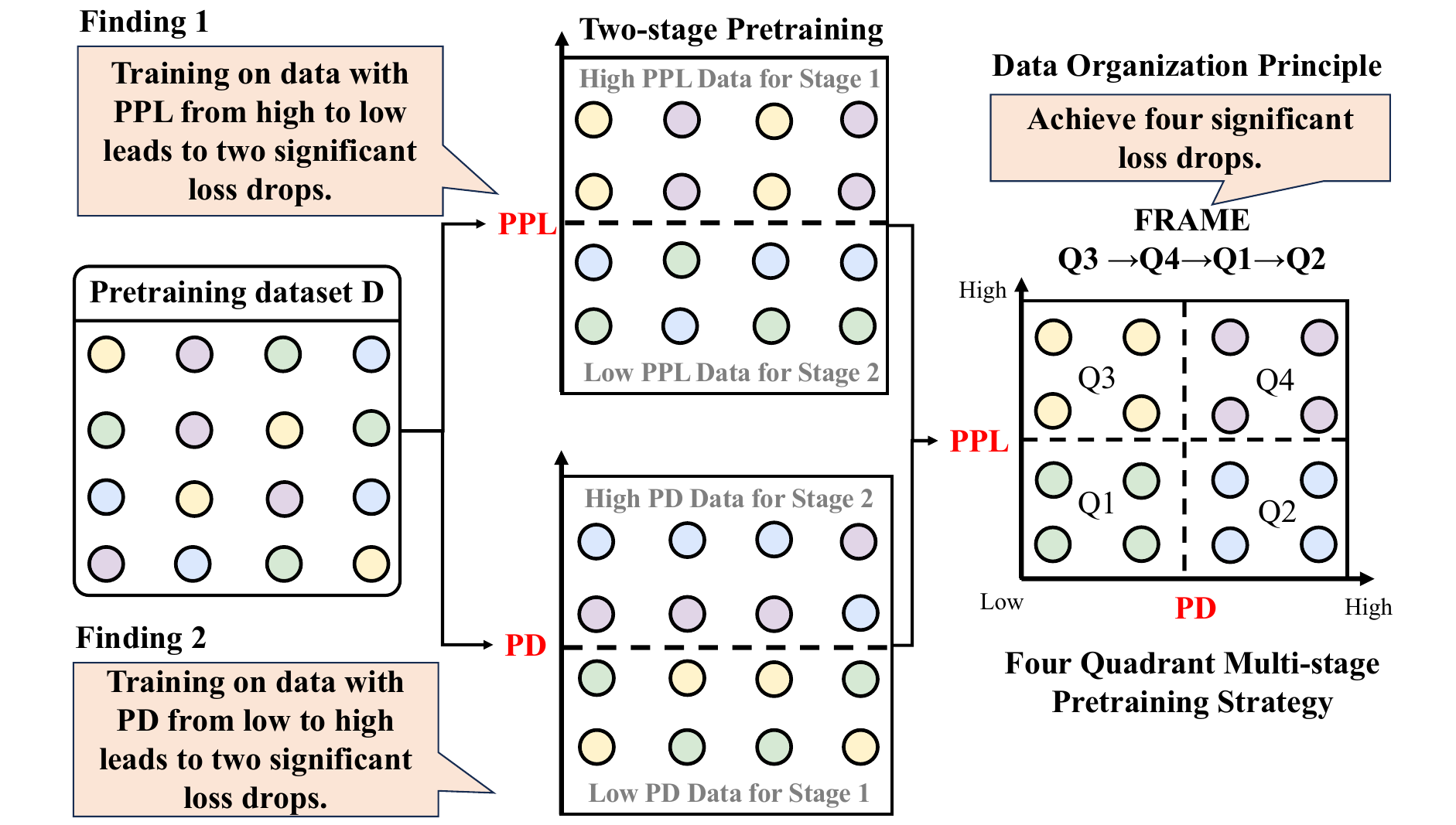}
    \caption{The overall framework of FRAME.}
    \label{fig:framework}
\end{figure}

In summary, our contributions are as follows: (1) We identify two key findings: first, training on high PPL data followed by low PPL data, and second, training on low PD data followed by high PD data, both causing two significant loss reductions and performance enhancements. (2) Guided by the principle of continuous loss reduction, we propose FRAME, a novel strategy that partitions data into four quadrants and organizes the pretraining process into four stages, resulting in loss dropping significantly four times and boosting model performance. (3) Experiments on 3B model, trained on 1T tokens, demonstrate a \textbf{16.8\%} improvement over uniform sampling in average performance across MMLU and CMMLU.

\section{Four Quadrant Multi-stage Pretraining Strategy}
% Determining the order of pretraining data involves organizing them at each training stage\cite{weinshall2018curriculum}. 
In this section, we first evaluate the effectiveness of PPL and PD as metrics for data organization in two-stage training (Sections \ref{subsec:Perplexity-based two stage learning} and \ref{subsec:PD-based two stage learning}). Based on these insights, we propose FRAME with both metrics(Section \ref{subsec:PPL-PD based four area framework}), as shown in Figure \ref{fig:framework}.

\subsection{Two-stage pretraining guided by PPL} 
\label{subsec:Perplexity-based two stage learning}

Research has explored pre-determining the sequence of pretraining data points based on their characteristics, which helps optimize models to globally optimal solutions \cite{pattnaik2024curry, soviany2022curriculum}. A critical aspect of this process is selecting appropriate metrics to organize data effectively, thereby minimizing training loss. Characteristics for text data include length, rare word frequency, and syntactic structure \cite{campos2021curriculum}. Although these heuristic methods seem reasonable from a human cognitive perspective, they may not necessarily align with the specific requirements of the model. Thus, data characteristics should be determined using metrics that are perceptible to the model and align with the standards of the target tasks \cite{xu2020curriculum}. In pretraining tasks, PPL closely aligns with the self-supervised learning objective (language modeling) and effectively evaluates model-data fit, making it an appropriate metric for organizing data.

\paragraph{Experimental Setting} We extract 500B tokens from a bilingual dataset, with both English and Chinese corpora \footnote{For the details of the dataset, please refer to Section \ref{sec:settings}}. We train a 1.3B reference model (RM) on the subset using a random sequence and compute PPL of the subset using the RM. Based on the median PPL of the dataset, we partition the training data into two equal subsets: \( \text{A}_{\text{PPL}}^{\text{low}} \) and \( \text{A}_{\text{PPL}}^{\text{high}} \). The data within each subset is uniformly distributed. We conduct two-stage training on the 3B model in the sequences \( \text{A}_{\text{PPL}}^{\text{low}} \rightarrow \text{A}_{\text{PPL}}^{\text{high}} \) and \( \text{A}_{\text{PPL}}^{\text{high}} \rightarrow \text{A}_{\text{PPL}}^{\text{low}} \), and compare the results with those from the random training model. We evaluate model performance using MMLU\cite{hendrycks2020measuring} and CMMLU\cite{li2023cmmlu}.

\begin{figure}[htbp]
    \centering
    \begin{subfigure}{0.47\linewidth}
        \centering
        \includegraphics[width=\linewidth]{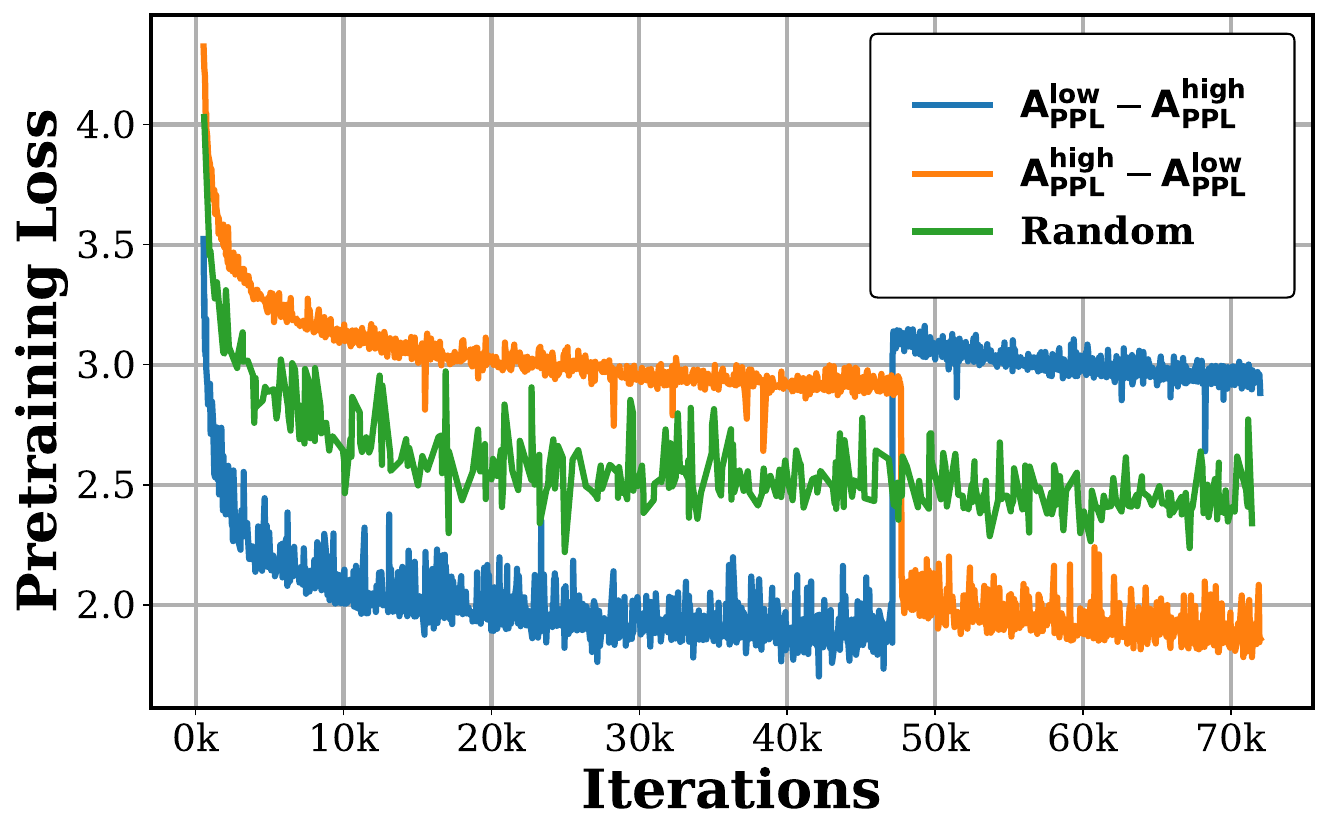}
        \caption{PPL training loss.}
        \label{fig:PPL_training_loss}
    \end{subfigure}
    \hspace{0.02\linewidth}
    \begin{subfigure}{0.47\linewidth}
        \centering
        \includegraphics[width=\linewidth]{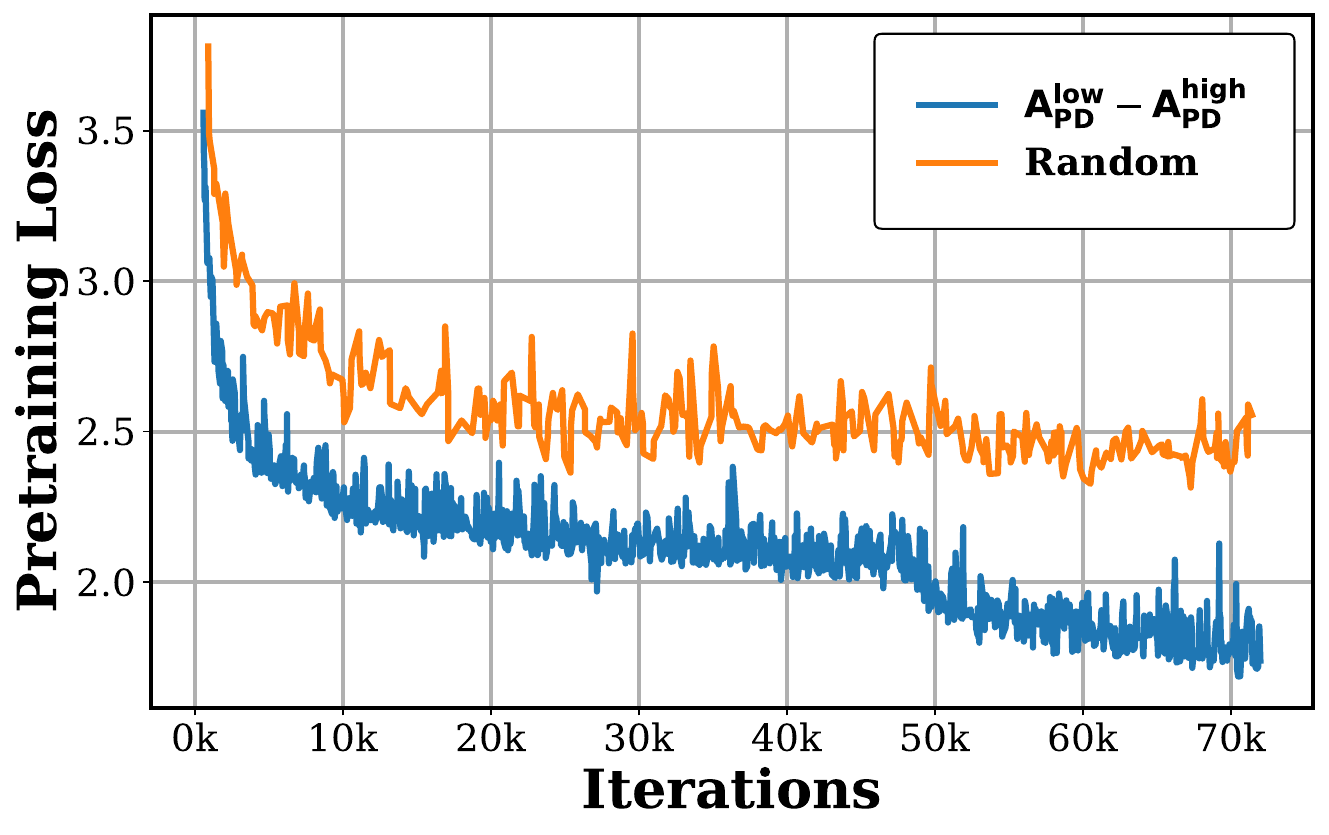}
        \caption{PD training loss.}
        \label{fig:PD_training_loss}
    \end{subfigure}
    \caption{Two-stage pretraining losses based on PPL and PD, respectively.}
    \label{fig:training_loss_comparison}
\end{figure}
\begin{table}[t]
    \centering
    \small
    \begin{tabular}{c|cc}
    \toprule
        \textbf{Methods} & \textbf{MMLU} & \textbf{CMMLU} \\
    \midrule
        Random  & 24.8 & 25.6 \\ 
        $\text{A}_{\text{PPL}}^{\text{low}} \to \text{A}_{\text{PPL}}^{\text{high}}$ & 26.0 & 26.0 \\ 
        $\text{A}_{\text{PPL}}^{\text{high}} \to \text{A}_{\text{PPL}}^{\text{low}}$ & \textbf{39.6} & \textbf{42.6} \\ 
    \bottomrule
    \end{tabular}
    \caption{Accuracy on MMLU and CMMLU for two-stage pretraining based on PPL with 3B models.}
    \label{fig:PPL training results}
\end{table}
% \begin{figure}[t]
%     \centering
%     \includegraphics[width=1.0\linewidth]{figure/ppl_train_loss_curve.pdf}
%     \caption{PPL training loss.}
%     \label{fig:PPL training loss}
% \end{figure}
% \begin{figure}[t]
%     \centering
%     \includegraphics[width=1.0\linewidth]{figure/pd_train_loss_curve.pdf}
%     \caption{PD training loss.}
%     \label{fig:PD training loss}
% \end{figure}

\paragraph{Results} Figure \ref{fig:PPL_training_loss} illustrates the changes in training loss over time. It is observed that training on high PPL data followed by low PPL data results in significant loss reductions occurring twice, ultimately achieving a lower loss level. Conversely, the reverse setting maintains a higher loss, which remains above that of the Random setting. Table \ref{fig:PPL training results} shows the model's benchmark accuracy for the settings \( \text{A}_{\text{PPL}}^{\text{low}} \rightarrow \text{A}_{\text{PPL}}^{\text{high}} \) and \( \text{A}_{\text{PPL}}^{\text{high}} \rightarrow \text{A}_{\text{PPL}}^{\text{low}} \). Training on high PPL data followed by low PPL data yields significantly higher performance than random training. In contrast, in the \( \text{A}_{\text{PPL}}^{\text{low}} \rightarrow \text{A}_{\text{PPL}}^{\text{high}} \) setting, the model shows only slight improvement. This leads to our \textbf{first key finding: training first on high PPL data followed by low PPL data can cause the loss to drop significantly twice, ultimately boosting model performance.}

\subsection{Two-stage pertaining guided by PD}
\label{subsec:PD-based two stage learning}
Subsequent analysis (detailed in Section \ref{subsec:analysis}) reveals that relying solely on PPL as a metric presents issues, as it is not stable across diverse data domains. Inspired by PDPC \cite{zhang2025preferencecurriculumllmspretrained}, PD between strong and weak models can also reflect the difficulty of samples for the models. Consider two models,  the weak model $M_w$ and the strong model $M_s$, both trained on an identical dataset $D$. For any sample \( x \), we calculate PD as follows:
\begin{equation}
\begin{aligned}
    &PD(x) = \frac{PPL_{M_w}(x) - PPL_{M_s}(x)}{PPL_{M_w}(x)} \\
    % &PPL_{M_{*}}(x) = exp{(-\frac{1}{L_x}\sum_{t=1}^{L_x}\log P(x_t|x_{<t}))}
\end{aligned}
\end{equation}
where \( PPL_{M_w}(x) \) and \( PPL_{M_s}(x) \) are the perplexity of the sample \( x \) calculated using \( M_w \) and \( M_s \), respectively. A low PD indicates similar learning efficiency for both models, while a high PD suggests the sample is more challenging for the weak model.

If checkpoints from earlier and later training stages of the same model are viewed as weak and strong models (with the same parameters but improved performance due to more data in later stages), then data with low PD values pose similar difficulty for both early and late stages, while data with high PD values are more challenging for the model's early checkpoints. In light of the above analysis, PD emerges as a model-aware difficulty metric that is well-suited for organizing text data.

The distributions of PPL and PD on different domains are analyzed in Appendix \ref{sec:Data Distribution}, PD exhibits a relatively consistent distribution across varied domains, following a normal distribution with an approximate mean value of 0.3. Compared to PPL, PD offers a better advantage in maintaining data diversity throughout each stage of training.

\paragraph{Experimental Setting}
% We adopt the same experimental setup as described in Section \ref{subsec:Perplexity-based two stage learning}. Additionally, we trained a reference model with 100M parameters and annotated the dataset with PPL. Subsequently, we calculated the \( PD \) for each sample using the the 100M and 1.3B reference models, denoted as \( PD(100M-1.3B) \).
We train a 100M parameter RM and then calculate the \( \text{PD} \) for each sample using both the 100M and 1.3B RMs, referred to as $\text{PD(100M-1.3B)}$. Using the median value of \(\text{PD}\) across all data, we divide the training dataset into two subsets with equal token counts: \( \text{A}_{\text{PD}}^{\text{low}} \) and \( \text{A}_{\text{PD}}^{\text{high}} \). 
Based on PDPC's finding that a low-to-high PD ordering achieves better results, we conduct a two-stage training process with \( \text{A}_{\text{PD}}^{\text{low}} \) first, followed by \( \text{A}_{\text{PD}}^{\text{high}} \), and compare the results to the random setting. We use the evaluation methods described in Section \ref{subsec:Perplexity-based two stage learning}.
% \begin{table}[t]
%     \centering
%     \small
%     \begin{tabular}{cccc}
%     \toprule
%         \textbf{Methods} & \textbf{Steps} & \textbf{MMLU} & \textbf{CMMLU} \\ 
%     \midrule
%         Random & \multirow{2}{*}{30K} & 25.43 & 25.88 \\ 
%         $\text{A}_{\text{PD}}^{\text{low}} \to \text{A}_{\text{PD}}^{\text{high}}$ & ~ & \textbf{26.65} & \textbf{26.19} \\ 
%     \midrule
%         Random & \multirow{2}{*}{60K} & 25.94 & \textbf{25.84} \\ 
%         $\text{A}_{\text{PD}}^{\text{low}} \to \text{A}_{\text{PD}}^{\text{high}}$ & ~ & \textbf{26.23} & 25.10 \\
%     \midrule
%         Random & \multirow{2}{*}{95K} & 24.84 & 25.57 \\ 
%         $\text{A}_{\text{PD}}^{\text{low}} \to \text{A}_{\text{PD}}^{\text{high}}$ & ~ & \textbf{26.89} & \textbf{27.20} \\ 
%     \bottomrule
%     \end{tabular}
%     \caption{Accuracy on MMLU and CMMLU for two-
% stage pretraining based on PD with 3B models.}
%     \label{fig:PD training results}
% \end{table}
\begin{table}[t]
    \centering
    \small
    \begin{tabular}{c|cc}
    \toprule
        \textbf{Methods} & \textbf{MMLU} & \textbf{CMMLU} \\ 
    \midrule
        Random & 24.8 & 25.6 \\ 
        $\text{A}_{\text{PD}}^{\text{low}} \to \text{A}_{\text{PD}}^{\text{high}}$  & \textbf{26.9} & \textbf{27.2} \\ 
    \bottomrule
    \end{tabular}
    \caption{Accuracy on MMLU and CMMLU for two-
stage pretraining based on PD with 3B models.}
    \label{fig:PD training results}
\end{table}
\paragraph{Results} Figure \ref{fig:PD_training_loss} shows the training loss changes under the \( \text{A}_{\text{PD}}^{\text{low}} \rightarrow \text{A}_{\text{PD}}^{\text{high}} \) setup. The loss initially drops rapidly during the low PD phase, then stabilizes, and decreases further with high PD data, eventually falling below the Random model's loss. It suggests that the first phase sets a beneficial optimization path for the second, helping avoid local optima. Table \ref{fig:PD training results} shows the model's accuracy under the \( \text{A}_{\text{PD}}^{\text{low}} \rightarrow \text{A}_{\text{PD}}^{\text{high}} \) setting. Notably, it exceeds the Random setting by 2.1\% on MMLU and 1.6\% on CMMLU, which validates PD as an effective metric. Further, this leads to our \textbf{second key finding: training first on low PD data followed by high PD data can cause the loss to drop significantly twice, ultimately boosting model performance.} 

\subsection{Four-Quadrant Guided Training Strategy}
\label{subsec:PPL-PD based four area framework}
Inspired by the two key findings about PPL and PD, we establish the principle of \textbf{organizing pretraining data to achieve significant reductions in training loss}. Based on this principle, we introduce a novel \textbf{F}our quad\textbf{RA}nt \textbf{M}ulti-stage pr\textbf{E}training strategy (FRAME), which uses PPL and PD to partition data and reorganize the data sequence to make the training loss drop significantly. The core process of FRAME is shown in Algorithm \ref{alg:frames}. 

\begin{figure}[t]
    \centering
    \begin{subfigure}{0.42\linewidth}
        \centering
        \includegraphics[width=\linewidth]{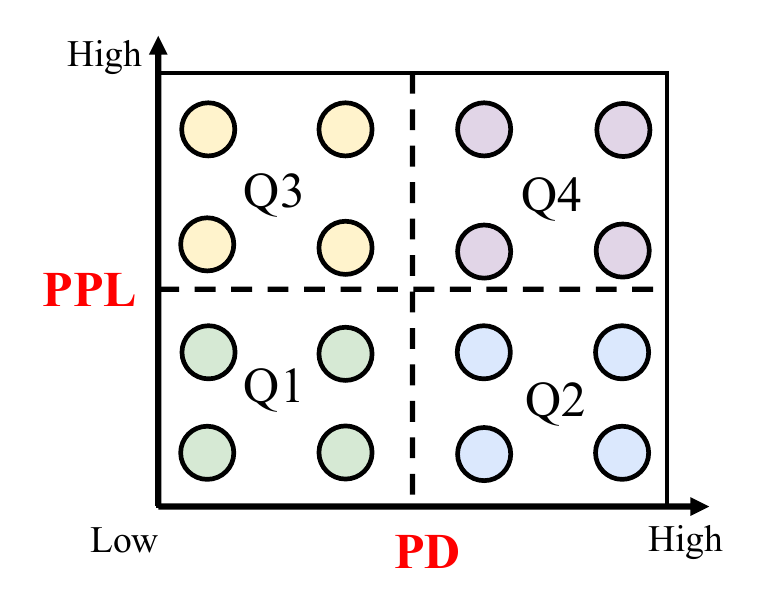}
     \caption{Four Quadrants.}
     \label{fig:Four Quadrant}
    \end{subfigure}
    \hspace{0.02\linewidth}
    \begin{subfigure}{0.47\linewidth}
        \centering
        \includegraphics[width=\linewidth]{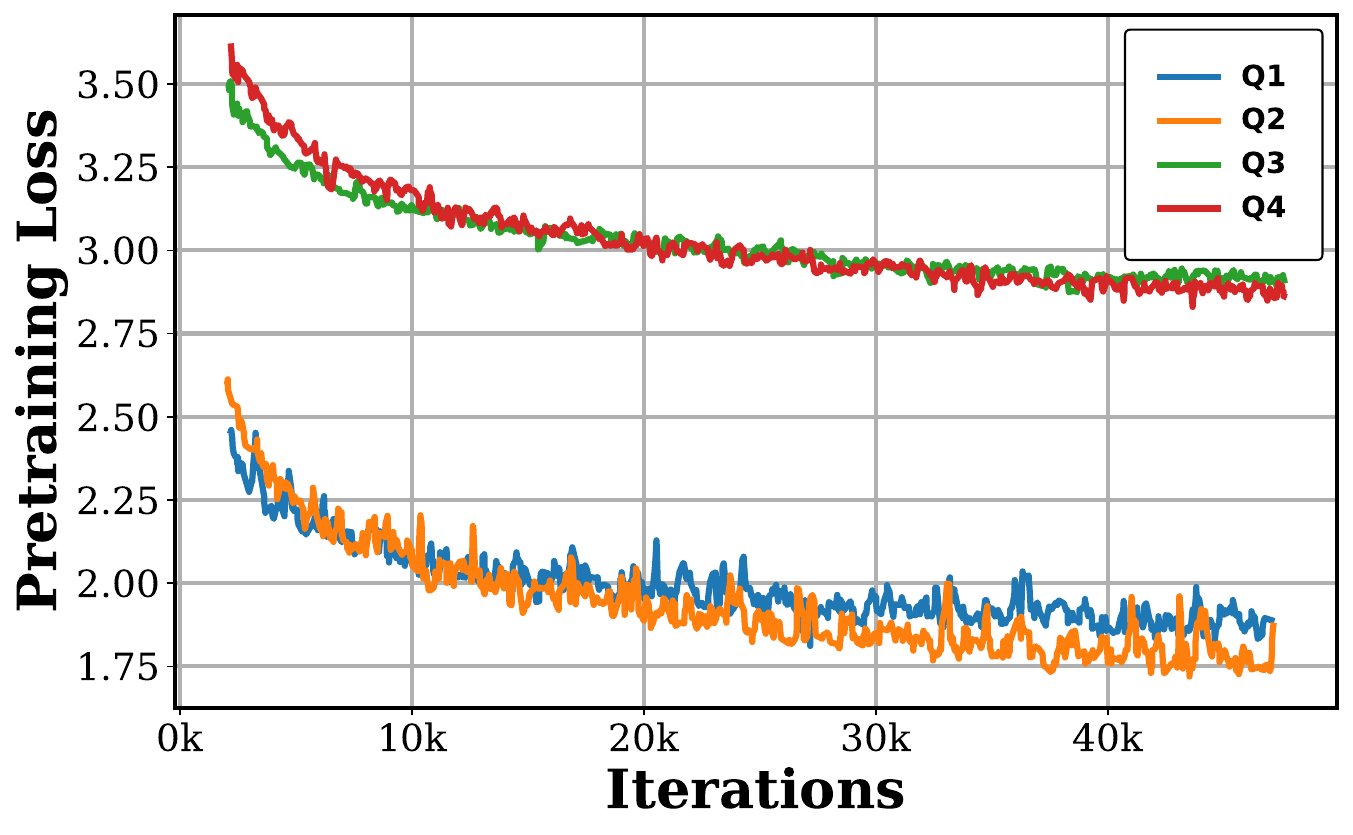}
        \caption{Losses of four quadrants.}
        \label{fig:Single_quadrant_training_loss}
    \end{subfigure}
    \caption{Four-quadrant partitioning and pretraining losses of four quadrants.}
    \label{fig:main_training_loss_comparison}
\end{figure}

Specifically, we train two RMs on the target training set \( D \), with the strong model \( M_s \) having more parameters than the weak model \( M_w \). Both models are trained on data from the same distribution and under identical settings. We compute PPL using \( M_s \) and PD using both \( M_s \) and \( M_w \) for each data point in $D$. We use the two metrics to partition \( \mathcal{D} \) into four quadrants, as shown in Figure \ref{fig:Four Quadrant}. Our main goal is to ensure that the token numbers of the four quadrants are roughly equal. Therefore, we first determine the PPL threshold to divide the entire dataset into two parts with equal token numbers. Then, for both the high PPL and low PPL parts, we separately find each of their respective PD thresholds to further divide them into two sub-parts with equal token numbers, resulting in four quadrants. During the experiments, there are only slight differences in the PD thresholds for the two PPL subsets, and these dierences did not affect the final
experimental results.

The data from four quadrants can be described as follows: {Quadrant \( \mathbf{Q_1} \)} contains data that both \( M_s \) and \( M_w \) can fit well; {Quadrant \( \mathbf{Q_2} \)} contains data that \( M_s \) learns well, but \( M_w \) struggles with. Data from this subset is challenging and requires a higher model capacity to understand them; {Quadrant \( \mathbf{Q_3} \)} contains data poorly learned by both \( M_s \) and \( M_w \); {Quadrant \( \mathbf{Q_4} \)} contains data poorly learned by \( M_s \) and even worse by \( M_w \).

% In accordance with the previous experimental insights, we need to adhere to the following constraints:
Based on the two key findings about PPL and PD, we should follow these constraints:
\begin{equation}
    \forall x \in S_i, \forall y \in S_{i+1}, \quad \text{PPL}(x) \geq \text{PPL}(y)
    \label{eq:ppl}
\end{equation}
\begin{equation}
    \forall x \in S_i, \forall y \in S_{i+1}, \quad \text{PD}(x) \leq \text{PD}(y)
\label{eq:pd}
\end{equation}
where \( S_i \) represents stage \( i \) of the training process. Equation (\ref{eq:ppl}) ensures that the PPL of data in stage \( i \) is not less than that in stage \( i+1 \). Equation (\ref{eq:pd}) ensures that the PD of data in stage \( i \) is not greater than that in stage \( i+1 \).

To efficiently organize data from the four quadrants while adhering to constraints, we use a double-loop approach: first, dividing the training into two main phases based on a specific constraint, then further splitting each main phase into two sub-phases according to another constraint. The approach yields two distinct four-stage training strategies: the first strategy follows the sequence \(Q_3 \rightarrow Q_4 \rightarrow Q_1 \rightarrow Q_2\), while the second adopts \(Q_3 \rightarrow Q_1 \rightarrow Q_4 \rightarrow Q_2\). Both strategies break the constraints between the second and third stages. However, experiments in Sections \ref{subsec:Perplexity-based two stage learning} and \ref{subsec:PD-based two stage learning} show that the loss change from low to high PPL is much greater than from high to low PD. Thus, we prioritize the constraints in Equation (\ref{eq:ppl}), making the first four-stage training strategy \(\mathbf{Q_3 \to Q_4 \to Q_1 \to Q_2} \) the better choice. 
% Furthermore, we implement an inner-to-outer hierarchical gradual transition method to ensure smooth data transitions between stages.

\begin{algorithm}[t]
\caption{\small Ordered dataset construction via FRAME}
\begin{algorithmic}[1]
\small
\State \textbf{Input:} pretraining data $D$, batch size $N$, steepness $a$
\State \textbf{Output:} ordered dataset $S_{\text{frame}}$
\Function{MergeDatasets}{$D_1$, $D_2$, RandomSample}
\State Total training steps $m \gets \frac{|D_1|+|D_2|}{N}$
\If{RandomSample}
    \State Randomly shuffle \(D_1\) and \(D_2\)
\EndIf
\State Initialize $l \gets 0$, $r \gets 0$, $Q_{\text{merged}} \gets []$
\For{i = 1 \textbf{to} m}
    \State Calculate completion ratio $p \gets \frac{i}{m}$
    \State Calculate proportion $f(p) \gets \frac{1}{1 + \exp(a(p - 0.5))}$
    \State $B_1 \gets D_1[l:l+f(p)N]$
    \State $B_2 \gets D_2[r:r+(1-f(p))N]$
    \State Batch $B_i = B_1 \cup B_2$
    \State $l \gets l + f(p)N$, $r \gets r + (1-f(p))N$
    \State Add all the samples in $B_i$ to $Q_{\text{merged}}$
\EndFor
\State \Return $Q_{\text{merged}}$
\EndFunction
\State Train RMs on i.i.d. subset of $D$
\State Calculate PPL and PD for all samples in $D$ using RMs
\State Divide into 4 quadrants: $Q_1$, $Q_2$, $Q_3$, $Q_4$, based on PPL and PD thresholds.
% \State \textbf{Four-Stage Training:}
\State $S_{34} \gets \textsc{MergeDatasets}(Q_3, Q_4, \text{True})$
\State $S_{12} \gets \textsc{MergeDatasets}(Q_1, Q_2, \text{True})$
\State $S_{\text{frame}} \gets \textsc{MergeDatasets}(S_{34}, S_{12}, \text{False})$
\end{algorithmic}
\label{alg:frames}
\end{algorithm}

% \begin{algorithm}[t]
% \caption{\small Four quadrant multi-stage pretraining strategy}
% \small
% \begin{algorithmic}[1]
% \State \textbf{Input:} pretraining data $D$, all training steps $K$
% \State \textbf{Output:} trained model $\theta_K$
% \State Initialize model parameters $\theta_0$
% \State Train RMs on i.i.d. subset of $D$ 
% \State Calculate PPL and PD for all samples in $D$ using RMs
% \State Divide $D$ into 4 quadrants: $Q_1$, $Q_2$, $Q_3$, and $Q_4$, based on PPL and PD median.

% \State \textbf{Inner-layer mixing:}
% \For{$t = 1$ \textbf{to} $\frac{|Q_3| + |Q_4|}{N}$}
%     \State Sample batch $B_t^{34}$ from $Q_3$ and $Q_4$ according f
% \EndFor
% \State $S_{34} \gets \bigcup_{t=1}^{\frac{|Q_3| + |Q_4|}{N}} B_t^{34}$

% \For{$t = 1$ \textbf{to} $\frac{|Q_1| + |Q_2|}{N}$}
%     \State Sample batch $B_t^{12}$ from $Q_1$ and $Q_2$
% \EndFor
% \State $S_{12} \gets \bigcup_{t=1}^{\frac{|Q_1| + |Q_2|}{N}} B_t^{12}$

% \For{$k = 1$ \textbf{to} $K$}
%     \State \textbf{Outer-layer mixing:}
%     \State Sample final batch $B_k$ from $S_{34}$ and $S_{12}$
    
%     \State Update model parameters $\theta_k$ using batch $B_k$
% \EndFor
%\State \textbf{Return:} trained model $\theta_K$
% \end{algorithmic}
% \label{alg:frames}
% \end{algorithm}

\paragraph{Formulation of FRAME with Stage Transition}  
As illustrated in Figure \ref{fig:PD ablation study}, direct stage transitions cause performance fluctuations. We aim to facilitate a smooth transition between stages. 

We start by outlining the smoothing process $f_{merge}$ for two-stage training: given the need to train on mixed data from two subsets \(D_1\) and \(D_2\), let \(i\) be the current training step and \(m\) the total number of training steps. The completion ratio is defined as \(p = \frac{i}{m}\). The sampled batch \(B_i\) should satisfy the following condition:
\begin{equation}
    % \small
    B_i = \{x \mid x \sim D_1\}_{f(p) \cdot N} \cup \{x \mid x \sim D_2\}_{(1 - f(p)) \cdot N} 
\end{equation}
where \(N\) represents the batch size, and \(f(p)\) denotes the proportion of samples from \(D_1\). Inspired by PDPC \cite{zhang2025preferencecurriculumllmspretrained}, we employ the S-shape function as \(f(p)\):

\begin{equation}
f(p) = \frac{1}{1 + \exp(a(p - 0.5))}   
\label{eq:sigmoid}
\end{equation}
where \(a\) controls the steepness of the curve. Unlike PDPC, we merely utilize the S-shape function for smoothing during stage transitions. Therefore, we use a larger \(a=35\) instead of \(a=10\) as in PDPC, to achieve a steeper function curve, as illustrated in Appendix \ref{sec:s-shape}.

In the four-stage training of FRAME, 
% we apply the above smoothing process to all data merging: 
we initially obtain $S_{34} = f_{merge}(Q_3, Q_4)$ and $S_{12} = f_{merge}(Q_1, Q_2)$. During this phase, samples are randomly selected from the two data sources for each batch. Subsequently, we construct $S_{frame} = f_{merge}(S_{34}, S_{12})$. In this phase, samples must be drawn in the original sequence of \(S_{12}\) and \(S_{34}\) to maintain the order of \(Q_3 \rightarrow Q_4 \rightarrow Q_1 \rightarrow Q_2\).

Notably, FRAME only organizes the given data without performing selection, which can serve as the final data preprocessing step before pretraining, without disrupting the continuity of training.

\section{Experiments}
\begin{table*}[t]
    \centering
    \setlength{\fboxsep}{1pt} % 调整背景框的内边距
    \scalebox{1.0}{
    \small
    \begin{tabular}{c|
    >{\centering\arraybackslash}m{1.5cm} 
    >{\centering\arraybackslash}m{1.5cm}|
    >{\centering\arraybackslash}m{1.5cm} 
    >{\centering\arraybackslash}m{1.5cm}
    >{\centering\arraybackslash}m{1cm}
    >{\centering\arraybackslash}m{1.2cm}
    >{\centering\arraybackslash}m{1cm}
    >{\centering\arraybackslash}m{1cm}
    }
    \toprule
        \textbf{Method} & \textbf{Metric} & \textbf{Order} & \textbf{ARC-E} & \textbf{ARC-C} & \textbf{SciQ} & \textbf{HellaSw.} & \textbf{PIQA} & \textbf{AVG.} \\ 
    \midrule
        \multicolumn{1}{c}{-} & - & Random & 56.5 & 23.6 & 85.8 & 34.2 & 67.3 & 53.5 \\ 
    \midrule
        \multirow{6}{*}{Sequential} & PD & High2Low & 54.7\textsuperscript{\hspace{0.3em}\raisebox{0.5ex}{\colorbox{red!10}{\tiny$\downarrow$1.8}}} & 21.8\textsuperscript{\hspace{0.3em}\raisebox{0.5ex}{\colorbox{red!10}{\tiny$\downarrow$1.8}}} & 87.1\textsuperscript{\hspace{0.3em}\raisebox{0.5ex}{\colorbox{green!10}{\tiny$\uparrow$1.3}}} & 33.7\textsuperscript{\hspace{0.3em}\raisebox{0.5ex}{\colorbox{red!10}{\tiny$\downarrow$0.5}}} & 67.8\textsuperscript{\hspace{0.3em}\raisebox{0.5ex}{\colorbox{green!10}{\tiny$\uparrow$0.5}}} & 53.0\textsuperscript{\hspace{0.3em}\raisebox{0.5ex}{\colorbox{red!10}{\tiny$\downarrow$0.5}}} \\
        & PD & Low2High & 56.1\textsuperscript{\hspace{0.3em}\raisebox{0.5ex}{\colorbox{red!10}{\tiny$\downarrow$0.4}}} & 21.3\textsuperscript{\hspace{0.3em}\raisebox{0.5ex}{\colorbox{red!10}{\tiny$\downarrow$2.3}}} & 86.2\textsuperscript{\hspace{0.3em}\raisebox{0.5ex}{\colorbox{green!10}{\tiny$\uparrow$0.4}}} & 34.4\textsuperscript{\hspace{0.3em}\raisebox{0.5ex}{\colorbox{green!10}{\tiny$\uparrow$0.2}}} & 67.6\textsuperscript{\hspace{0.3em}\raisebox{0.5ex}{\colorbox{red!10}{\tiny$\downarrow$0.2}}} & 53.1\textsuperscript{\hspace{0.3em}\raisebox{0.5ex}{\colorbox{red!10}{\tiny$\downarrow$0.4}}} \\
        & PPL & High2Low & 45.5\textsuperscript{\hspace{0.3em}\raisebox{0.5ex}{\colorbox{red!10}{\tiny$\downarrow$11.0}}} & 20.6\textsuperscript{\hspace{0.3em}\raisebox{0.5ex}{\colorbox{red!10}{\tiny$\downarrow$3.0}}} & 71.2\textsuperscript{\hspace{0.3em}\raisebox{0.5ex}{\colorbox{red!10}{\tiny$\downarrow$14.6}}} & 30.3\textsuperscript{\hspace{0.3em}\raisebox{0.5ex}{\colorbox{red!10}{\tiny$\downarrow$3.9}}} & 63.7\textsuperscript{\hspace{0.3em}\raisebox{0.5ex}{\colorbox{red!10}{\tiny$\downarrow$3.6}}} & 46.3\textsuperscript{\hspace{0.3em}\raisebox{0.5ex}{\colorbox{red!10}{\tiny$\downarrow$7.2}}} \\
        & PPL & Low2High & 47.8\textsuperscript{\hspace{0.3em}\raisebox{0.5ex}{\colorbox{red!10}{\tiny$\downarrow$8.7}}} & 17.9\textsuperscript{\hspace{0.3em}\raisebox{0.5ex}{\colorbox{red!10}{\tiny$\downarrow$5.7}}} & 72.7\textsuperscript{\hspace{0.3em}\raisebox{0.5ex}{\colorbox{red!10}{\tiny$\downarrow$13.1}}} & 29.1\textsuperscript{\hspace{0.3em}\raisebox{0.5ex}{\colorbox{red!10}{\tiny$\downarrow$5.1}}} & 62.4\textsuperscript{\hspace{0.3em}\raisebox{0.5ex}{\colorbox{red!10}{\tiny$\downarrow$4.9}}} & 46.0\textsuperscript{\hspace{0.3em}\raisebox{0.5ex}{\colorbox{red!10}{\tiny$\downarrow$7.5}}} \\
        & Qu.Edu & High2Low & 57.2\textsuperscript{\hspace{0.3em}\raisebox{0.5ex}{\colorbox{green!10}{\tiny$\uparrow$0.7}}} & 26.4\textsuperscript{\hspace{0.3em}\raisebox{0.5ex}{\colorbox{green!10}{\tiny$\uparrow$2.8}}} & 85.4\textsuperscript{\hspace{0.3em}\raisebox{0.5ex}{\colorbox{red!10}{\tiny$\downarrow$0.4}}} & 33.0\textsuperscript{\hspace{0.3em}\raisebox{0.5ex}{\colorbox{red!10}{\tiny$\downarrow$1.2}}} & 66.2\textsuperscript{\hspace{0.3em}\raisebox{0.5ex}{\colorbox{red!10}{\tiny$\downarrow$1.1}}} & 53.6\textsuperscript{\hspace{0.3em}\raisebox{0.5ex}{\colorbox{green!10}{\tiny$\uparrow$0.1}}} \\
        & Qu.Edu & Low2High & 56.8\textsuperscript{\hspace{0.3em}\raisebox{0.5ex}{\colorbox{green!10}{\tiny$\uparrow$0.3}}} & 26.0\textsuperscript{\hspace{0.3em}\raisebox{0.5ex}{\colorbox{green!10}{\tiny$\uparrow$2.4}}} & 84.1\textsuperscript{\hspace{0.3em}\raisebox{0.5ex}{\colorbox{red!10}{\tiny$\downarrow$1.7}}} & 33.5\textsuperscript{\hspace{0.3em}\raisebox{0.5ex}{\colorbox{red!10}{\tiny$\downarrow$0.7}}} & 67.9\textsuperscript{\hspace{0.3em}\raisebox{0.5ex}{\colorbox{green!10}{\tiny$\uparrow$0.6}}} & 53.7\textsuperscript{\hspace{0.3em}\raisebox{0.5ex}{\colorbox{green!10}{\tiny$\uparrow$0.2}}} \\
    \midrule
        \multirow{5}{*}{Preference CL} & PPL & S.R. & 56.1\textsuperscript{\hspace{0.3em}\raisebox{0.5ex}{\colorbox{red!10}{\tiny$\downarrow$0.4}}} & 24.1\textsuperscript{\hspace{0.3em}\raisebox{0.5ex}{\colorbox{green!10}{\tiny$\uparrow$0.5}}} & 87.8\textsuperscript{\hspace{0.3em}\raisebox{0.5ex}{\colorbox{green!10}{\tiny$\uparrow$2.0}}} & 33.9\textsuperscript{\hspace{0.3em}\raisebox{0.5ex}{\colorbox{red!10}{\tiny$\downarrow$0.3}}} & 67.4\textsuperscript{\hspace{0.3em}\raisebox{0.5ex}{\colorbox{green!10}{\tiny$\uparrow$0.1}}} & 53.9\textsuperscript{\hspace{0.3em}\raisebox{0.5ex}{\colorbox{green!10}{\tiny$\uparrow$0.4}}} \\
        & PPL & S. & 56.1\textsuperscript{\hspace{0.3em}\raisebox{0.5ex}{\colorbox{red!10}{\tiny$\downarrow$0.4}}} & 22.6\textsuperscript{\hspace{0.3em}\raisebox{0.5ex}{\colorbox{red!10}{\tiny$\downarrow$1.0}}} & 85.5\textsuperscript{\hspace{0.3em}\raisebox{0.5ex}{\colorbox{red!10}{\tiny$\downarrow$0.3}}} & 34.2\textsuperscript{\hspace{0.3em}\raisebox{0.5ex}{{\tiny~~~0.0}}} & 67.5\textsuperscript{\hspace{0.3em}\raisebox{0.5ex}{\colorbox{green!10}{\tiny$\uparrow$0.2}}} & 53.2\textsuperscript{\hspace{0.3em}\raisebox{0.5ex}{\colorbox{red!10}{\tiny$\downarrow$0.3}}} \\
        & Qu.Edu & S.R & 56.7\textsuperscript{\hspace{0.3em}\raisebox{0.5ex}{\colorbox{green!10}{\tiny$\uparrow$0.2}}} & 24.9\textsuperscript{\hspace{0.3em}\raisebox{0.5ex}{\colorbox{green!10}{\tiny$\uparrow$1.3}}} & 86.2\textsuperscript{\hspace{0.3em}\raisebox{0.5ex}{\colorbox{green!10}{\tiny$\uparrow$0.4}}} & 33.6\textsuperscript{\hspace{0.3em}\raisebox{0.5ex}{\colorbox{red!10}{\tiny$\downarrow$0.6}}} & 66.9\textsuperscript{\hspace{0.3em}\raisebox{0.5ex}{\colorbox{red!10}{\tiny$\downarrow$0.4}}} & 53.7\textsuperscript{\hspace{0.3em}\raisebox{0.5ex}{\colorbox{green!10}{\tiny$\uparrow$0.2}}} \\
        & Qu.Edu & S. & 55.5\textsuperscript{\hspace{0.3em}\raisebox{0.5ex}{\colorbox{red!10}{\tiny$\downarrow$1.0}}} & 24.8\textsuperscript{\hspace{0.3em}\raisebox{0.5ex}{\colorbox{green!10}{\tiny$\uparrow$1.2}}} & 87.8\textsuperscript{\hspace{0.3em}\raisebox{0.5ex}{\colorbox{green!10}{\tiny$\uparrow$2.0}}} & 34.0\textsuperscript{\hspace{0.3em}\raisebox{0.5ex}{\colorbox{green!10}{\tiny$\uparrow$0.2}}} & 67.4\textsuperscript{\hspace{0.3em}\raisebox{0.5ex}{\colorbox{green!10}{\tiny$\uparrow$0.1}}} & 53.9\textsuperscript{\hspace{0.3em}\raisebox{0.5ex}{\colorbox{green!10}{\tiny$\uparrow$0.4}}} \\
        & PD & S.R. & 56.7\textsuperscript{\hspace{0.3em}\raisebox{0.5ex}{\colorbox{green!10}{\tiny$\uparrow$0.2}}} & 24.9\textsuperscript{\hspace{0.3em}\raisebox{0.5ex}{\colorbox{green!10}{\tiny$\uparrow$1.3}}} & 86.2\textsuperscript{\hspace{0.3em}\raisebox{0.5ex}{\colorbox{green!10}{\tiny$\uparrow$0.4}}} & 33.6\textsuperscript{\hspace{0.3em}\raisebox{0.5ex}{\colorbox{red!10}{\tiny$\downarrow$0.6}}} & 67.4\textsuperscript{\hspace{0.3em}\raisebox{0.5ex}{\colorbox{green!10}{\tiny$\uparrow$0.1}}} & 53.8\textsuperscript{\hspace{0.3em}\raisebox{0.5ex}{\colorbox{green!10}{\tiny$\uparrow$0.3}}} \\
    \midrule
        PDPC & PD & S. & 57.3\textsuperscript{\hspace{0.3em}\raisebox{0.5ex}{\colorbox{green!10}{\tiny$\uparrow$0.8}}} & 
        26.6\textsuperscript{\hspace{0.3em}\raisebox{0.5ex}{\colorbox{green!10}{\tiny$\uparrow$3.0}}} & 
        87.9\textsuperscript{\hspace{0.3em}\raisebox{0.5ex}{\colorbox{green!10}{\tiny$\uparrow$2.1}}} & 
        33.7\textsuperscript{\hspace{0.3em}\raisebox{0.5ex}{\colorbox{red!10}{\tiny$\downarrow$0.5}}} & 
        68.0\textsuperscript{\hspace{0.3em}\raisebox{0.5ex}{\colorbox{green!10}{\tiny$\uparrow$0.7}}} & 
        54.7\textsuperscript{\hspace{0.3em}\raisebox{0.5ex}{\colorbox{green!10}{\tiny$\uparrow$1.2}}} \\
    \midrule
        FRAME & PPL\&PD & - & \textbf{62.9}\textsuperscript{\hspace{0.3em}\raisebox{0.5ex}{\colorbox{green!10}{\tiny$\uparrow$6.4}}} & \textbf{26.5}\textsuperscript{\hspace{0.3em}\raisebox{0.5ex}{\colorbox{green!10}{\tiny$\uparrow$2.9}}} & \textbf{90.5}\textsuperscript{\hspace{0.3em}\raisebox{0.5ex}{\colorbox{green!10}{\tiny$\uparrow$4.7}}} & \textbf{38.4}\textsuperscript{\hspace{0.3em}\raisebox{0.5ex}{\colorbox{green!10}{\tiny$\uparrow$4.2}}} & \textbf{69.6}\textsuperscript{\hspace{0.3em}\raisebox{0.5ex}{\colorbox{green!10}{\tiny$\uparrow$2.3}}} & \textbf{57.6}\textsuperscript{\hspace{0.3em}\raisebox{0.5ex}{\colorbox{green!10}{\tiny$\uparrow$4.1}}} \\
    \bottomrule
    \end{tabular}
    }
    \caption{Downstream tasks results for different settings on \textbf{1.3B} models. We report accuracy for each task, and the best performances are marked in bold. Abbreviations: AVG. = Average, S.=S-shape Function, S.R.=S-shape Reverse Function.}
    \label{tab:1B main results}
\end{table*}

\begin{table*}[t]
    \centering
    \small
    \setlength{\fboxsep}{1pt} % 调整背景框的内边距
    \scalebox{0.85}{
    \begin{tabular}{l|cccccccccc}
    \toprule
        \textbf{Method} & \textbf{MMLU} & \textbf{CMMLU} & \textbf{CEVAL} & \textbf{BBH} & \textbf{ARC-E} & \textbf{ARC-C} & \textbf{HellaSw.} & \textbf{PIQA} & \textbf{AVG.} \\ 
    \midrule
        Random &  27.7  &  27.5  &  27.2  &  27.9  & 68.6 & 33.7 & 49.4 & 76.0 & 42.3  \\ 
    \midrule 
        PDPC  &  35.8\textsuperscript{\hspace{0.3em}\raisebox{0.5ex}{\colorbox{green!10}{\tiny$\uparrow$8.1}}}  &  35.6\textsuperscript{\hspace{0.3em}\raisebox{0.5ex}{\colorbox{green!10}{\tiny$\uparrow$8.1}}}  &  36.1\textsuperscript{\hspace{0.3em}\raisebox{0.5ex}{\colorbox{green!10}{\tiny$\uparrow$8.9}}}  &  25.7\textsuperscript{\hspace{0.3em}\raisebox{0.5ex}{\colorbox{red!10}{\tiny$\downarrow$2.2}}}  & 
        69.7\textsuperscript{\hspace{0.3em}\raisebox{0.5ex}{\colorbox{green!10}{\tiny$\uparrow$1.1}}} & 
        35.8\textsuperscript{\hspace{0.3em}\raisebox{0.5ex}{\colorbox{green!10}{\tiny$\uparrow$2.1}}} & 
        49.9\textsuperscript{\hspace{0.3em}\raisebox{0.5ex}{\colorbox{green!10}{\tiny$\uparrow$0.5}}} & 
        76.3\textsuperscript{\hspace{0.3em}\raisebox{0.5ex}{\colorbox{green!10}{\tiny$\uparrow$0.3}}} & 
        45.6\textsuperscript{\hspace{0.3em}\raisebox{0.5ex}{\colorbox{green!10}{\tiny$\uparrow$3.3}}}  \\ 
    \midrule
        $Q_3 \to Q_1 \to Q_4 \dashrightarrow Q_2$ &  25.8\textsuperscript{\hspace{0.3em}\raisebox{0.5ex}{\colorbox{red!10}{\tiny$\downarrow$1.9}}}  &  26.4\textsuperscript{\hspace{0.3em}\raisebox{0.5ex}{\colorbox{red!10}{\tiny$\downarrow$1.1}}}  &  25.5\textsuperscript{\hspace{0.3em}\raisebox{0.5ex}{\colorbox{red!10}{\tiny$\downarrow$1.6}}}  &  \textbf{28.9}\textsuperscript{\hspace{0.3em}\raisebox{0.5ex}{\colorbox{green!10}{\tiny$\uparrow$1.0}}}  & 
        68.3\textsuperscript{\hspace{0.3em}\raisebox{0.5ex}{\colorbox{red!10}{\tiny$\downarrow$0.3}}} &
        35.2\textsuperscript{\hspace{0.3em}\raisebox{0.5ex}{\colorbox{green!10}{\tiny$\uparrow$1.5}}} & 
        48.3\textsuperscript{\hspace{0.3em}\raisebox{0.5ex}{\colorbox{red!10}{\tiny$\downarrow$1.1}}} &
        75.4\textsuperscript{\hspace{0.3em}\raisebox{0.5ex}{\colorbox{red!10}{\tiny$\downarrow$0.6}}} & 
        41.7\textsuperscript{\hspace{0.3em}\raisebox{0.5ex}{\colorbox{red!10}{\tiny$\downarrow$0.6}}}  \\
    \midrule
        FRAME &  \textbf{43.0}\textsuperscript{\hspace{0.3em}\raisebox{0.5ex}{\colorbox{green!10}{\tiny\textbf{$\uparrow$15.3}}}}  &  \textbf{45.7}\textsuperscript{\hspace{0.3em}\raisebox{0.5ex}{\colorbox{green!10}{\tiny\textbf{$\uparrow$18.2}}}}  &  \textbf{44.0}\textsuperscript{\hspace{0.3em}\raisebox{0.5ex}{\colorbox{green!10}{\tiny\textbf{$\uparrow$16.8}}}}  &  27.9\textsuperscript{\hspace{0.3em}\raisebox{0.5ex}{{\tiny~~~0.0}}}  &
        \textbf{71.0}\textsuperscript{\hspace{0.3em}\raisebox{0.5ex}{\colorbox{green!10}{\tiny\textbf{$\uparrow$2.4}}}} &
        \textbf{36.5}\textsuperscript{\hspace{0.3em}\raisebox{0.5ex}{\colorbox{green!10}{\tiny\textbf{$\uparrow$2.8}}}} & 
        \textbf{50.2}\textsuperscript{\hspace{0.3em}\raisebox{0.5ex}{\colorbox{green!10}{\tiny\textbf{$\uparrow$0.8}}}} &
        \textbf{76.9}\textsuperscript{\hspace{0.3em}\raisebox{0.5ex}{\colorbox{green!10}{\tiny\textbf{$\uparrow$0.9}}}} & \textbf{49.4}\textsuperscript{\hspace{0.3em}\raisebox{0.5ex}{\colorbox{green!10}{\tiny\textbf{$\uparrow$6.1}}}}  \\ 
    \bottomrule
    \end{tabular}
    }
    \caption{Results of downstream tasks for different methods using \textbf{3B} models on \textbf{1T} tokens. "\(\dashrightarrow\)" indicates that the model's accuracy has significantly decreased before reaching this stage, so we stopped at the third stage.}
    \label{tab:3B main results}
\end{table*}

% \begin{figure}[ht]
%     \centering
    
%     \vspace{0.5cm}
    
%     \begin{subfigure}{0.49\linewidth}
%         \centering
%         \includegraphics[width=\linewidth]{figure/CEVAL_performance.pdf}
%         \caption{Accuracy on CEVAL.}
%         \label{fig:sciq}
%     \end{subfigure} 
%     \hfill
%     \begin{subfigure}{0.49\linewidth}
%         \centering
%         \includegraphics[width=\linewidth]{figure/average_performance.pdf}
%         \caption{Accuracy on average.}
%         \label{fig:piqa}
%     \end{subfigure}
%     \caption{Few-shot downstream performance on various benchmarks with respect to pretraining iterations for Random and FRAME.}
%     \label{fig:overall}
% \end{figure}

\begin{figure}[t]
    \centering
    \begin{subfigure}{0.49\linewidth}
        \centering
        \includegraphics[width=\linewidth]{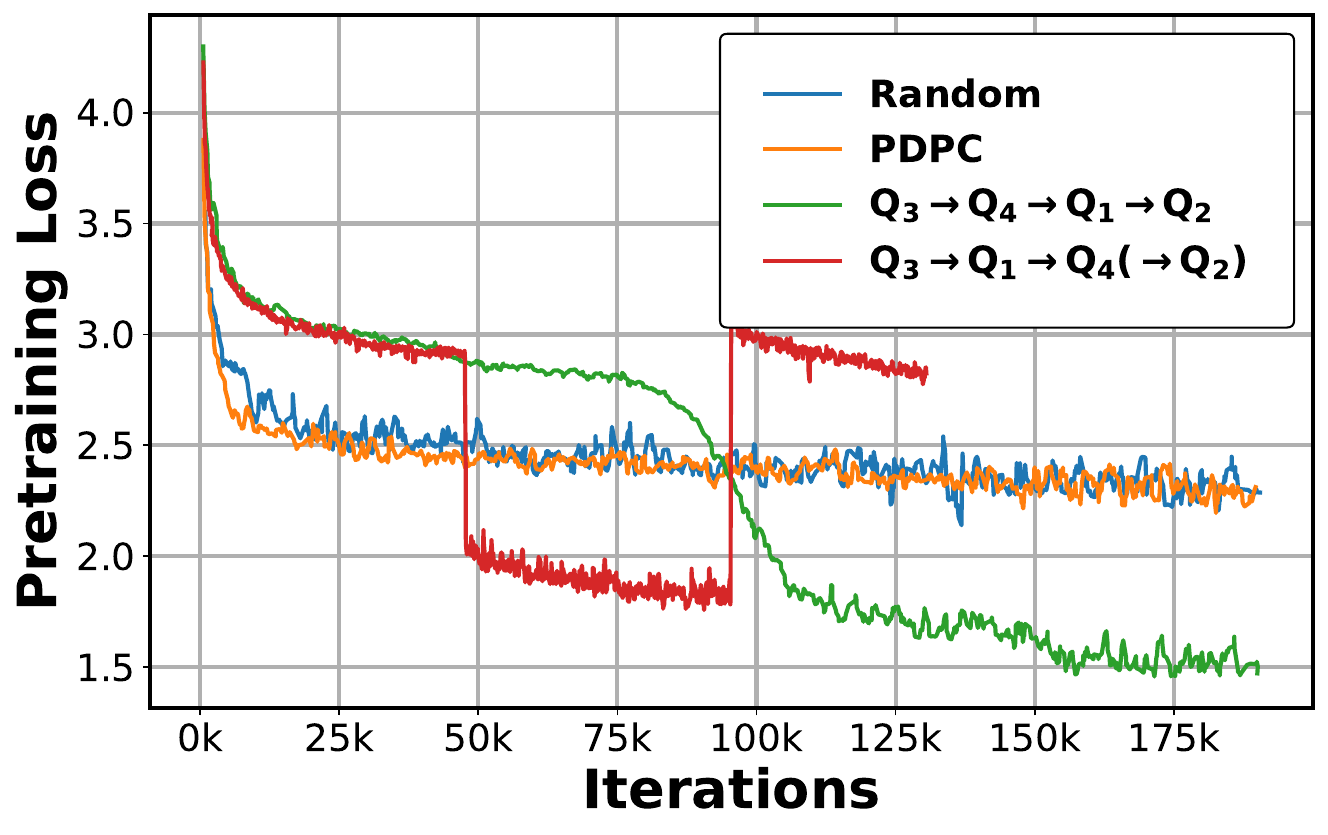}
        \caption{}
        % \label{fig:sciq}
    \end{subfigure} 
    \hfill
    \begin{subfigure}{0.49\linewidth}
        \centering
        \includegraphics[width=\linewidth]{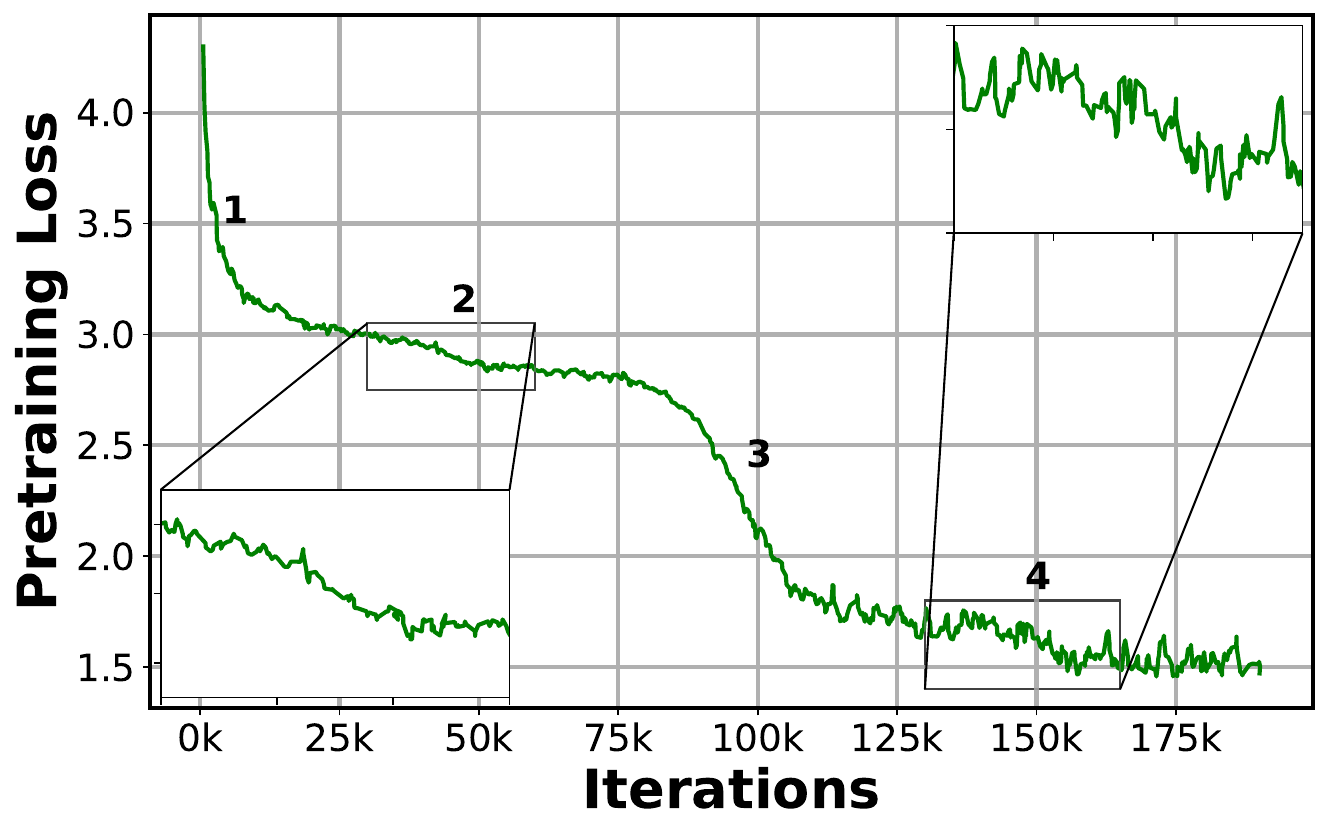}
        \caption{}
        % \label{fig:piqa}
    \end{subfigure}
    \caption{Pretraining losses of main results. (a) Comparison of different methods. (b) Details of pretraining loss. The marked numbers indicate the process of the loss decreasing four times.}
    \label{fig:Main training loss}
\end{figure}

\subsection{Settings}
\label{sec:settings}

\paragraph{Data Source} For 3B models, our pretraining data is derived from various domains, including books \cite{gao2020pile}, blogs \cite{baumgartner2020pushshift}, patents \cite{sharma2019bigpatent}, Common Crawl \cite{penedo2024fineweb}, and Wikipedia. It comprises a total of 1T tokens, with 500B tokens each for Chinese and English, akin to the Matrix dataset \cite{zhang2024mapneo}. For 1.3B models, we randomly select 100B tokens from the SlimPajama dataset \cite{cerebras2023slimpajama}.

\paragraph{Pretraining Setting}
We train 100M and 1.3B models on i.i.d subsets of the collected dataset, comprising 500B tokens, to compute PPL and PD. And we test on 3B models for the main experiments, with a batch size of 640 and a context window length of 8192. The Adam optimizer is used for training within the Megatron framework. In addition, we validate our approach on a 1.3B model, employing the same configuration. More details of model structure and training could be found in Appendix \ref{subsec:Experimental settings}.

\paragraph{Baselines}
We compare FRAME with random data sequence and PDPC \cite{zhang2025preferencecurriculumllmspretrained}. In addition we also validate on a 1.3B model, comparing it with other models based on PPL, PD, and QuRating metrics. More details of baselines could be found in Appendix \ref{subsec:Experimental settings}.

\paragraph{Evaluation}
For 1.3B models, we evaluate the models' performance on the following benchmarks: ARC-E \cite{clark2018think}, ARC-C \cite{clark2018think}, SciQ \cite{welbl2017crowdsourcing}, HellaSwag \cite{zellers2019hellaswag}, and PIQA \cite{bisk2020piqa}. 
% which include tasks such as knowledge question answering and commonsense reasoning. 
For 3B models, we additionally use benchmarks including MMLU \cite{hendrycks2020measuring}, CMMLU \cite{li2023cmmlu}, BBH \cite{suzgun2022challengingbigbenchtaskschainofthought}, and CEVAL \cite{huang2023cevalmultilevelmultidisciplinechinese}, covering more complicated tasks.
% covering tasks that assess multi-domain knowledge and complex reasoning. 
We apply in-context learning and select examples based on task characteristics. Standard accuracy serves as the final metric for all tasks.

\subsection{Main Results}
Table \ref{tab:3B main results} shows our experimental results. FRAME significantly outperforms the Random baseline, with a 6.1\% improvement in average performance. For specific tasks, it achieves notable gains of 15.3\% on MMLU and 18.2\% on CMMLU. For the 1.3B model, as illustrated in Table \ref{tab:1B main results}, FRAME also surpasses all the baselines, with 4.1\% improvement in average, and a maximum of 6.4\% gain in a single task. These findings validate the effectiveness of FRAME.  
Moreover, in our experiments with the 1.3B and 3B models, distinct pretraining datasets are utilized for each model. Notably, FRAME consistently yield substantial performance gains across all evaluated benchmarks. These results highlight the robustness and generalizability of our method, demonstrating its effectiveness across varying data sources and model architectures.

Figure \ref{fig:AVG} shows the model's benchmark performance over training steps. The model remains stable initially, with a significant performance increase after 90K steps. Figure \ref{fig:Main training loss} illustrates the training loss, showing four declines that align with benchmark performance improvements. This confirms the strong link between training loss and benchmark performance: when training stages are equally divided, lower training loss results in better performance. Additionally, FRAME effectively enhances the model's emergent capabilities, allowing it to acquire foundational skills quickly and better understand data patterns in later stages, improving overall performance. Through the loss smoothness analysis in Appendix \ref{subsec:Smoother LOSS Curve from FRAMES}, we found that the loss curve of FRAME has the lowest high-frequency energy proportion of only 0.02\%, significantly lower than Random and PDPC. This indicates that FRAME can make the model converge more stably and reduce the impact of gradient fluctuations during training.

We find that the \(Q_3 \to Q_4 \to Q_1 \to Q_2\) strategy significantly outperforms the \(Q_3 \to Q_1 \to Q_4 \to Q_2\) strategy, with the latter showing a performance decline in the third stage. This supports our argument that prioritizing the PPL dimension over the PD dimension is beneficial.

\subsection{Ablation Study}

% \begin{figure}[t]
%     \centering
%     \includegraphics[width=\linewidth]{figure/3B_4quadrant_training_loss.pdf}
%     \caption{Single quadrant training loss.}
%     \label{fig:Single quadrant training loss}
% \end{figure}
% \begin{figure}[t]
%     \centering
%     \includegraphics[width=\linewidth]{figure/main_3B_train_loss_curve.pdf}
%     \caption{Main training loss.}
%     \label{fig:Main training loss}
% \end{figure}
We investigate the differential effects of data from distinct quadrants and to characterize their intrinsic properties. Figure \ref{fig:Single_quadrant_training_loss} presents the training loss trajectories across different quadrants. The results demonstrate a significant disparity in loss magnitude, with high PPL data ($Q_3$ and $Q_4$) exhibiting substantially greater loss values compared to their low PPL counterparts ($Q_1$ and $Q_2$). Furthermore, under equivalent PPL conditions, data with higher PD ($Q_2$ and $Q_4$) consistently enabled the model to converge to lower final loss values than data with lower PD ($Q_1$ and $Q_3$). Notably, our analysis reveals that PPL exerts a more substantial influence on training loss than PD.

Furthermore, we could observe a slower convergence rate when training on high PD data. This phenomenon can be attributed to the difficulty of high PD samples, which are particularly challenging for the model in the early stage of training when its capacity is still limited. However, as training progresses, the model's capabilities gradually improve. Since the weak and strong models fit low PD data similarly, even as the model enhances, the training loss will not further decrease, leading to faster convergence. On the other hand, high PD data, with the improvement of model abilities, can further guide the model to learn new features and continue reducing training loss.

\begin{figure}[t]
    \centering
    \begin{subfigure}{0.49\linewidth}
        \centering
        \includegraphics[width=\linewidth]{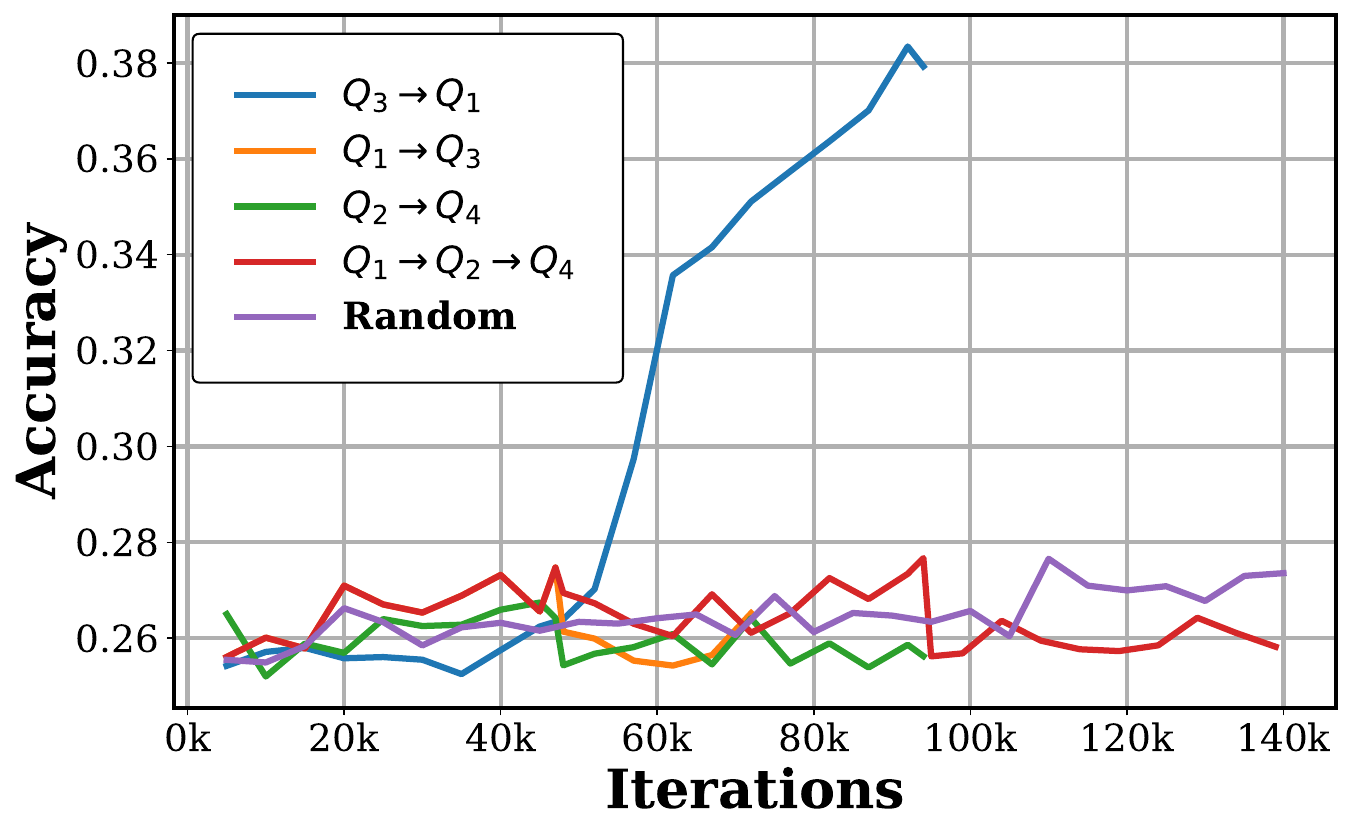}
        \caption{PPL ablation study.}
        \label{fig:PPL ablation study}
    \end{subfigure}
    \hfill
    \begin{subfigure}{0.49\linewidth}
        \centering
        \includegraphics[width=\linewidth]{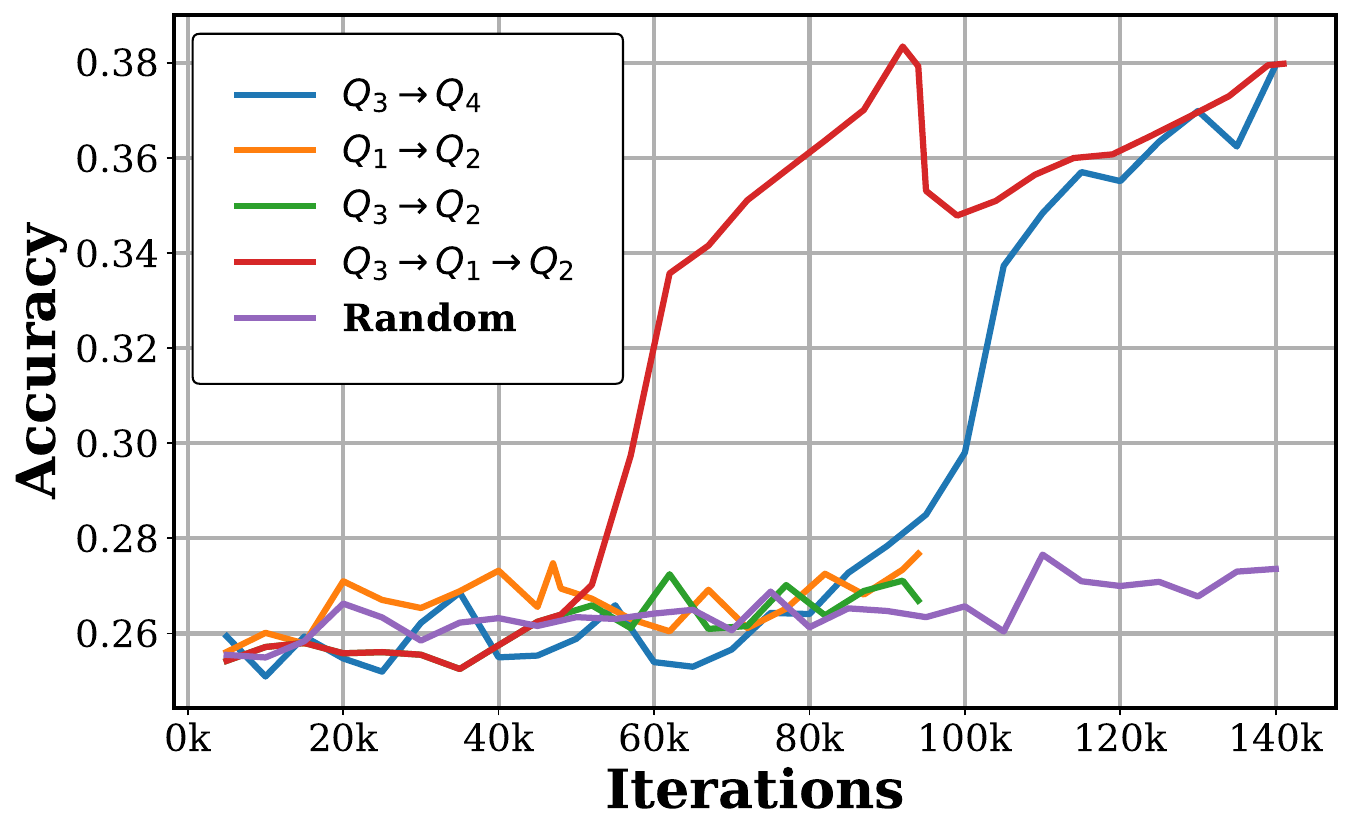}
        \caption{PD ablation study.}
        \label{fig:PD ablation study}
    \end{subfigure}
    \caption{Ablation studies of different combinations.}
\end{figure}

% From the standpoint of four quadrants, do the conclusions of Sections \ref{subsec:Perplexity-based two stage learning} and \ref{subsec:PPL-PD based four area framework} still apply? We ran tests with different quadrant combinations, and here's what we found:
From the perspective of four quadrants, do the conclusions of Sections \ref{subsec:Perplexity-based two stage learning} and \ref{subsec:PD-based two stage learning} still hold? We test various quadrant combinations and find that:

% (1) The principle that it's better for ppl to decrease still stands, unaffected by whether PD is large or small. Figure \ref{} illustrates the outcomes for each combination. We see a significant performance boost when transitioning from larger PPL data to smaller ones (A3->A1). However, moving in the opposite direction (A1->A3 and A2->A4) causes a certain degree of accuracy drop, regardless of whether PD is large or small. In a longer training period, moving from A1->A2 to A4 also results in performance decline.
% \paragraph{Training first on high PPL data followed by low PPL data proves to be a superior strategy, unaffected by whether PD is high or low.} 
\paragraph{Training first on high PPL data followed by low PPL data proves to be a superior strategy} 
Figure \ref{fig:PPL ablation study} illustrates the outcomes for each combination. We see a significant performance boost when transitioning from larger PPL data to smaller ones \(Q_3 \rightarrow Q_1\). However, moving in the opposite direction \(Q_1 \rightarrow Q_3\) and \( Q_2 \rightarrow Q_4\) causes a certain degree of accuracy drop, regardless of whether PD is large or small. Over a longer training period, moving from \(Q_1 \rightarrow Q_2\) to $Q_4$ also results in a performance decline.

% (2) The concept that it's better for PD to increase still applies, regardless of the size of PPL. As shown in Figure \ref{}, no matter the size of PPL, when PD grows (A3->A4, A1->A2 and A3->A2), the model performs better on the benchmark. Furthermore, in a longer training phase, moving from A3->A1 to A2 can enhance the model's performance even more. This aligns with the findings in \cite{}.
% \paragraph{Starting with low PD data followed by high PD data is more effective, regardless of whether PPL is high or low.} 
\paragraph{Starting with low PD data followed by high PD data is more effective}
As shown in Figure \ref{fig:PD ablation study}, when PD increases (\(Q_3 \rightarrow Q_4\), \(Q_1 \rightarrow Q_2\), and \(Q_3 \rightarrow Q_2\)), the model performs better on the benchmark. Furthermore, in a longer training phase, transitioning from \(Q_3 \rightarrow Q_1 \rightarrow Q_2\) can further enhance the model's performance.

In addition, the \(Q_3 \rightarrow Q_1 \rightarrow Q_2\) setup, where $f_{merge}$ is not applied, shows a performance drop between the second and third stages. This suggests that $f_{merge}$ is necessary. More results can be found in Appendix \ref{subsec:Ablation Studies}.
% \noindent \textbf{(3) Data with larger PPL is ideal for priming the model training.} Comparing A3->A2 and A1->A2, the former exhibits a greater accuracy increase, suggesting that data with larger PPL is suitable for the initial warm-up stage of model training.
% \paragraph{High PPL data is suitable for the initial warm-up stage of training.} Comparing A3 \(\rightarrow\) A2 and A1 \(\rightarrow\) A2, the former shows a greater accuracy increase, indicating that data with larger PPL is beneficial for the initial warm-up stage of model training.

\subsection{Analysis}
\label{subsec:analysis}

\begin{figure}[t]
    \centering
    \includegraphics[width=0.9\linewidth]{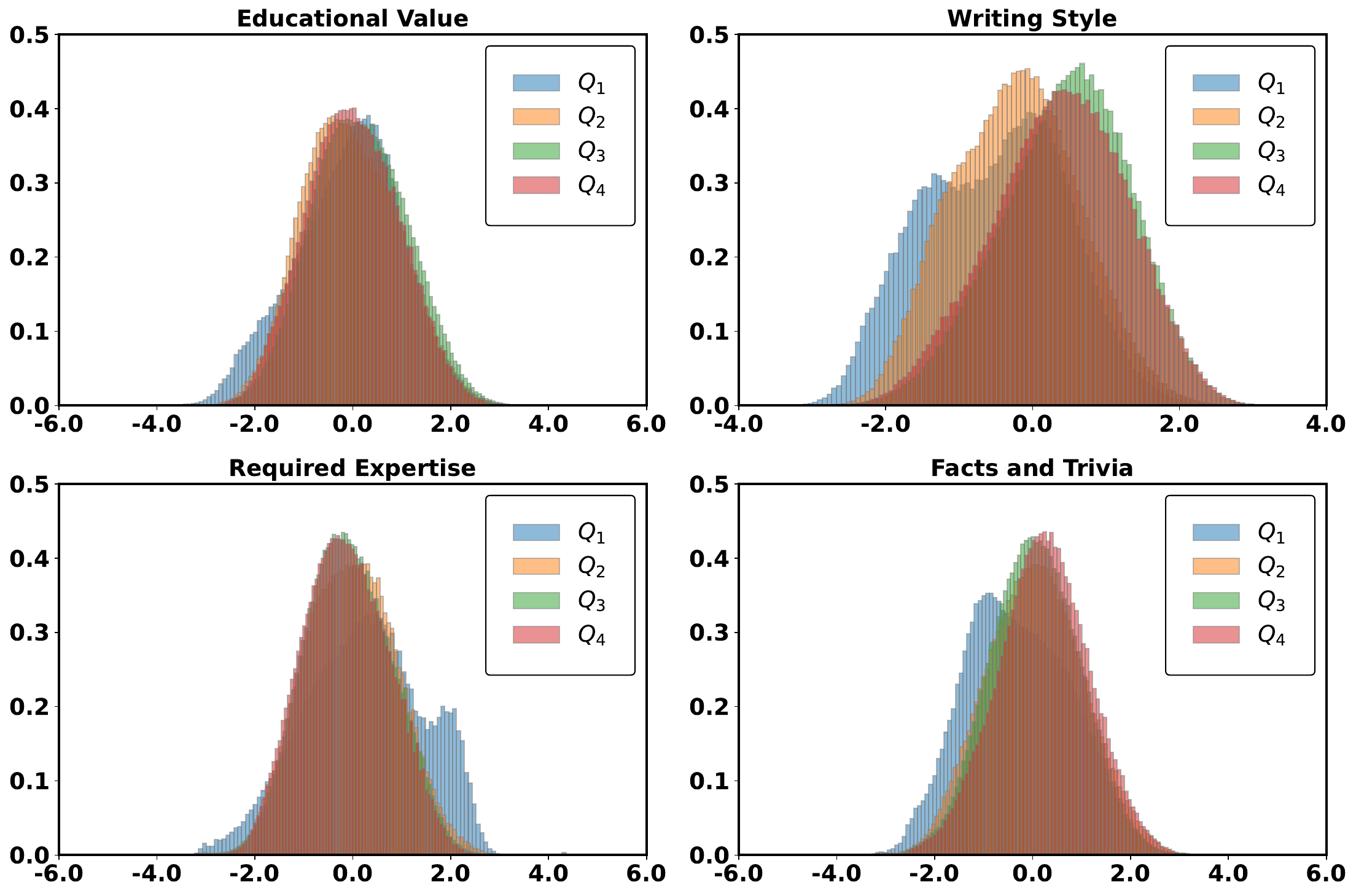}
    \caption{Human intuitive quality metric distribution.}
    \label{fig:quality_distribution}
\end{figure}

\paragraph{Data in Quadrants Without Inherent Human Intuitive Cognition Favorability}
% We aim to investigate the characteristics of data within the four quadrants. To this end, we extracted 1,000 samples from each quadrant and performed a quality analysis using four distinct scoring functions from QuRating. As shown in Figure \ref{}, we observe that the quality distributions across all dimensions are similar for samples from the four quadrants. This indicates that PPL and PD, as model-perceived data measures, do not exhibit a significant correlation with data measures based on human intuitive cognition, such as knowledge, quality, and diversity. Our training framework ensures the homogeneity of data quality throughout the model training process, preventing scenarios where the model learns only low-quality data during any specific stage.
We investigate the characteristics of data within the four quadrants by extracting 1,000 samples from each and analyzing their quality using four raters from QuRating. As shown in Figure \ref{fig:quality_distribution}, the quality distributions are similar across all dimensions for samples from each quadrant, which suggests that model-perceived measures like PPL and PD don't significantly correlate with human cognition-based measures, such as knowledge, quality, and diversity. FRAME maintains consistent data quality throughout the training process, preventing the model from focusing solely on low-quality data at any stage.

\begin{figure}[t]
    \centering
    \includegraphics[width=1.0\linewidth]{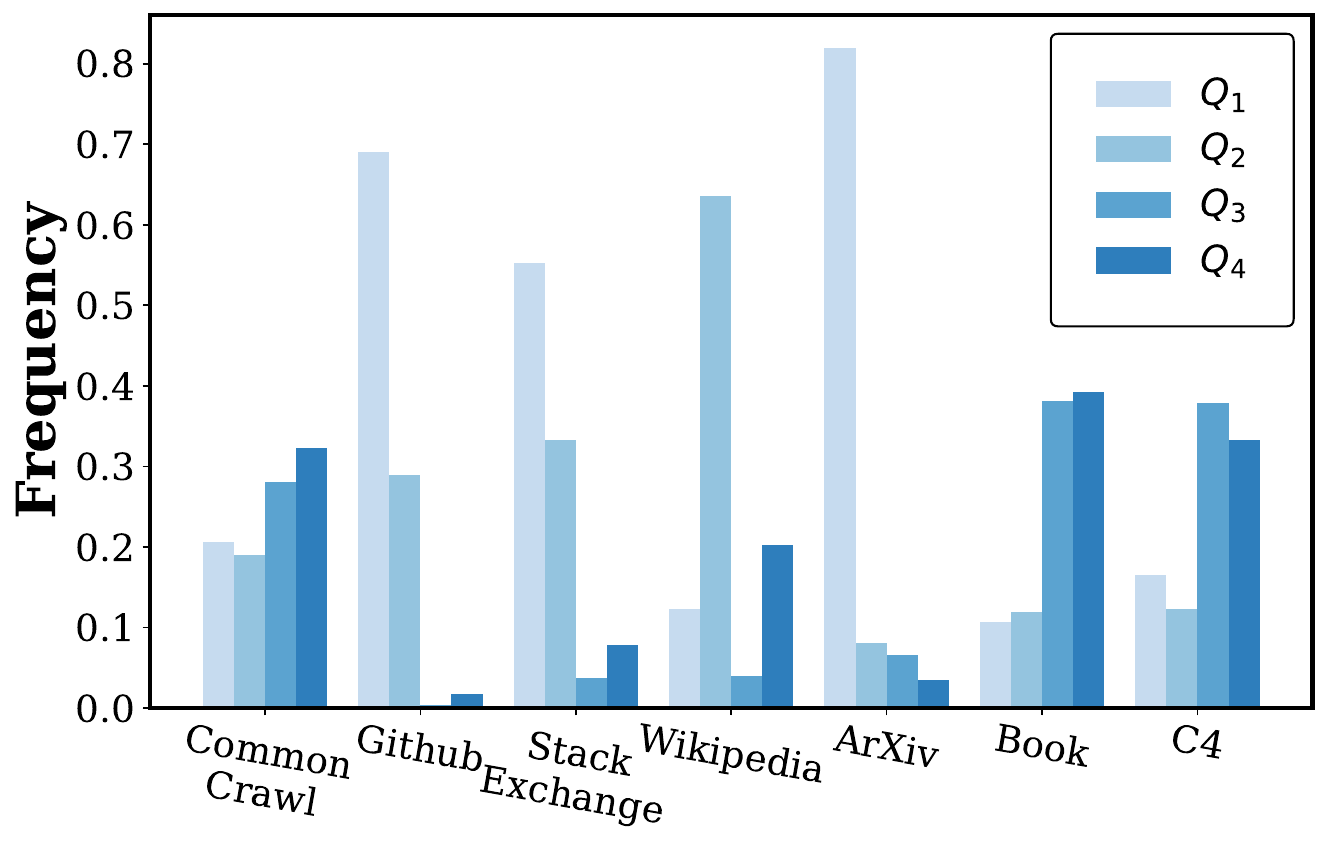}
    \caption{Data distribution across different sources.}
    \label{fig:Data distribution}
\end{figure}

% \begin{figure*}[t]
%     \centering
%     \includegraphics[width=1.0\linewidth]{figure/zh_data_distribution.png}
%     \caption{ZH Data distribution.}
%     \label{fig:ZH Data distribution}
% \end{figure*}
% \begin{figure}[t]
%     \centering
%     \begin{subfigure}{0.49\linewidth}
%         \centering
%         \includegraphics[width=\linewidth]{figure/en_data_distribution.png}
%         \caption{EN Data distribution.}
%         \label{fig:EN Data distribution}
%     \end{subfigure}
%     \hfill
%     \begin{subfigure}{0.49\linewidth}
%         \centering
%         \includegraphics[width=\linewidth]{figure/zh_data_distribution.png}
%         \caption{ZH Data distribution.}
%         \label{fig:ZH Data distribution}
%     \end{subfigure}
%     \caption{Comparison of EN and ZH Data distributions.}
%     \label{fig:Data distributions}
% \end{figure}

\paragraph{PPL and PD Distribution}

We extract 70K samples from 7 domains within our training dataset, covering areas such as Arxiv, Law, Code, and Math. Using the 1.3B reference model, we calculate the PPL of all samples and present the PPL distributions in Figure \ref{fig:PPL Distribution}. Significant differences are observed across various domains, with mean values ranging from 5 to 22 and varying degrees of variance. Data from domains such as Wikipedia typically exhibit lower PPL, whereas data from Reddit show higher PPL. This implies that sorting data by PPL could cause obvious shifts in domain representation between initial and later training stages, potentially exposing the model to overly homogeneous data during each training stage. \citeauthor{sachdeva2024train} (\citeyear{sachdeva2024train}) highlight that data diversity enhances model performance, and an uneven distribution may lead to a decline in performance.

We calculate the PD of the sampled data from the 100M and 1.3B reference models and visualize its distribution across various sources, as shown in Figure \ref{fig:PD Distribution}. We find that the PD distribution is quite similar across different sources. This similarity implies that using PD as a metric for data partitioning allows each training phase to include data from various sources, ensuring data diversity throughout the training process and thereby maintaining pretraining efficiency.

\paragraph{Data Distribution of the Four Quadrants}

% We extracted 1 million samples from each of the four quadrant distributions and visualized these samples based on their domains in the original dataset, as shown in Figure \ref{}. We observed that almost all domains are represented within each quadrant. This indicates that sorting data solely by domain is not an optimal approach. Instead, PPL and PD, as model-perceived metrics, provide more precise data ordering signals.
% We extract 1 million samples from each of the four quadrant distributions and visualize them, as shown in Figure \ref{}. We observe that each quadrant contains data from a wide range of sources, indicating that FRAME effectively ensures data diversity throughout the training process. 

% Furthermore, there are distinct differences among the quadrants. For instance, A2 contains a large number of domain-specific datasets such as arXiv, Wikipedia, and PubMedCentral. This aligns with common cognitive understanding, as these datasets may feature more complex sentence structures and terminology, which large models can learn effectively while small models may struggle to master. Additionally, the data distribution in A3 is relatively balanced. According to this research, the initial training phase of the model is better suited for learning sufficiently diverse data. Therefore, it is reasonable to present data in A3 at the beginning of the training process.
We extract 1 million samples from each of the four quadrant distributions and visualize them, as shown in Figure \ref{fig:Data distribution}. We observe that each quadrant contains data from a wide range of sources, indicating that FRAME effectively ensures data diversity throughout the training process. Notably, there are distinct differences among the quadrants. For instance, $Q_2$ contains a large number of domain-specific datasets such as Wikipedia, and StackExchange. This aligns with common cognitive understanding, as these datasets may feature more complex sentence structures and terminology, which large models can learn effectively while small models may struggle to master, thus suitable for learning in the later stages of training. This also explains the excellent performance in knowledge-intensive tasks, as the texts in Wikipedia contain a wealth of factual knowledge. Additionally, the data distribution in $Q_3$ is relatively balanced, which is advantageous for the initial training phase. Presenting data from $Q_3$ at the beginning of the training process allows the model to learn from sufficiently diverse data, thereby establishing a strong foundation for subsequent learning stages.

\begin{figure}
    \centering
    \includegraphics[width=0.8\linewidth]{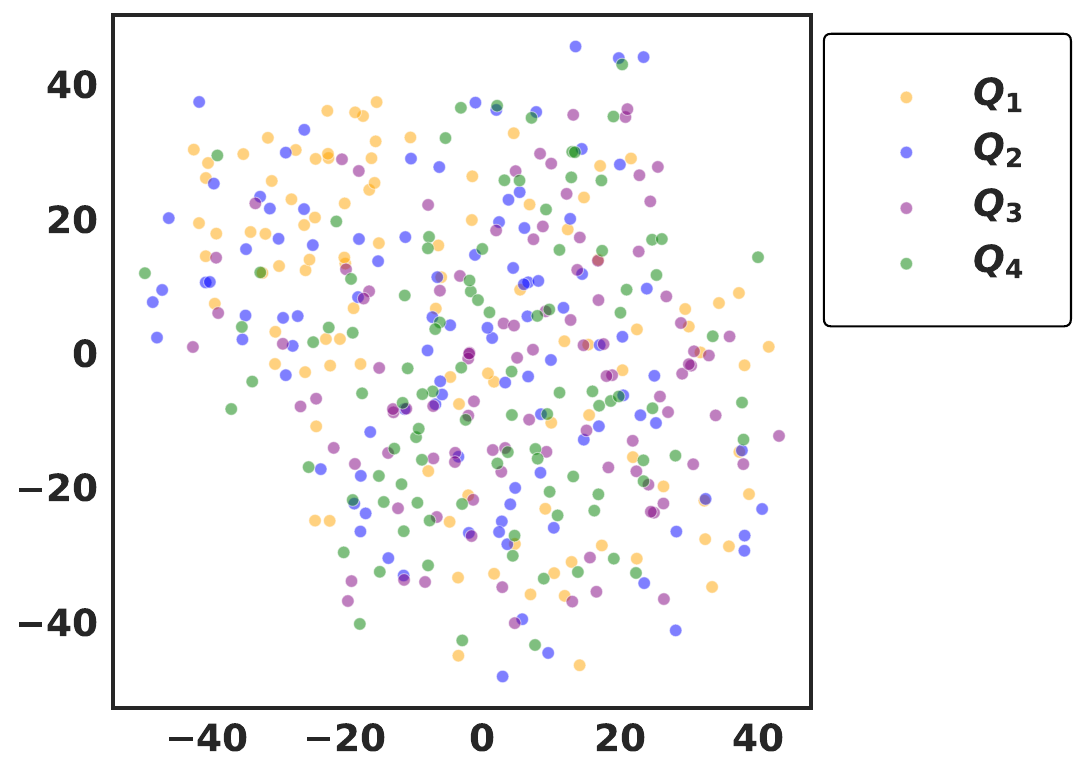}
    \caption{Semantic distribution of different quadrants.}
    \label{fig:Semantic diversity}
\end{figure}

\paragraph{Semantic Properties Analysis}
To assess distinct semantic properties within each quadrant's data, we randomly select 1K samples from each quadrant and devise 10 language-text related traits for GPT-4o evaluation. For simplicity and precision, all traits are formulated as yes-or-no questions, and we calculate the percentage of samples meeting each trait's criteria. Details of all traits are listed in Appendix \ref{sec:prompt}. As shown in Figure \ref{fig:GPT analyses}, the data across all four quadrants exhibit similar semantic properties. In addition, we use T5 \cite{raffel2023exploringlimitstransferlearning} to obtain dense vectors of the samples and perform dimensionality reduction using t-SNE, as shown in Figure \ref{fig:Semantic diversity}. We could observe that the data in the four quadrants do not have obvious distinctions at the semantic level. This indicates no obvious difference in the semantic distribution of data between the initial and later training stages, ensuring a diverse range of data is consistently encountered throughout the training process, and avoiding the collapse of the model into a certain preference, thereby damaging generalization performance.
% \begin{figure}
%     \centering
%     \includegraphics[width=1.0\linewidth]{figure/gpt_analysis1.pdf}
%     \caption{GPT analysis1}
%     \label{fig:GPT analysis1}
% \end{figure}

% \begin{figure}
%     \centering
%     \includegraphics[width=1.0\linewidth]{figure/gpt_analysis2.pdf}
%     \caption{GPT analysis2}
%     \label{fig:GPT analysis2}
% \end{figure}
\begin{figure}[t]
    \centering
    \begin{subfigure}{0.45\linewidth}
        \centering
        \includegraphics[width=\linewidth]{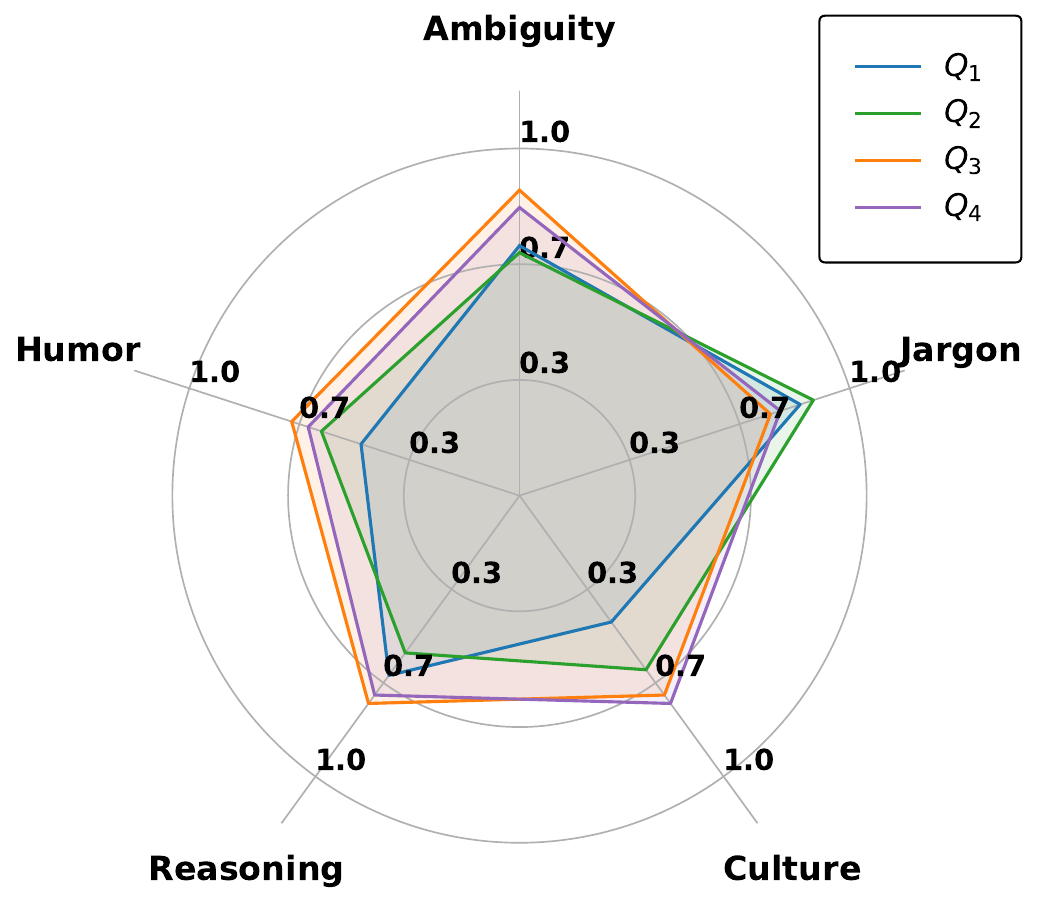}
        % \caption{GPT analysis1}
        \label{fig:GPT analysis1}
    \end{subfigure}
    \hspace{0.02\linewidth}
    \begin{subfigure}{0.47\linewidth}
        \centering
        \includegraphics[width=\linewidth]{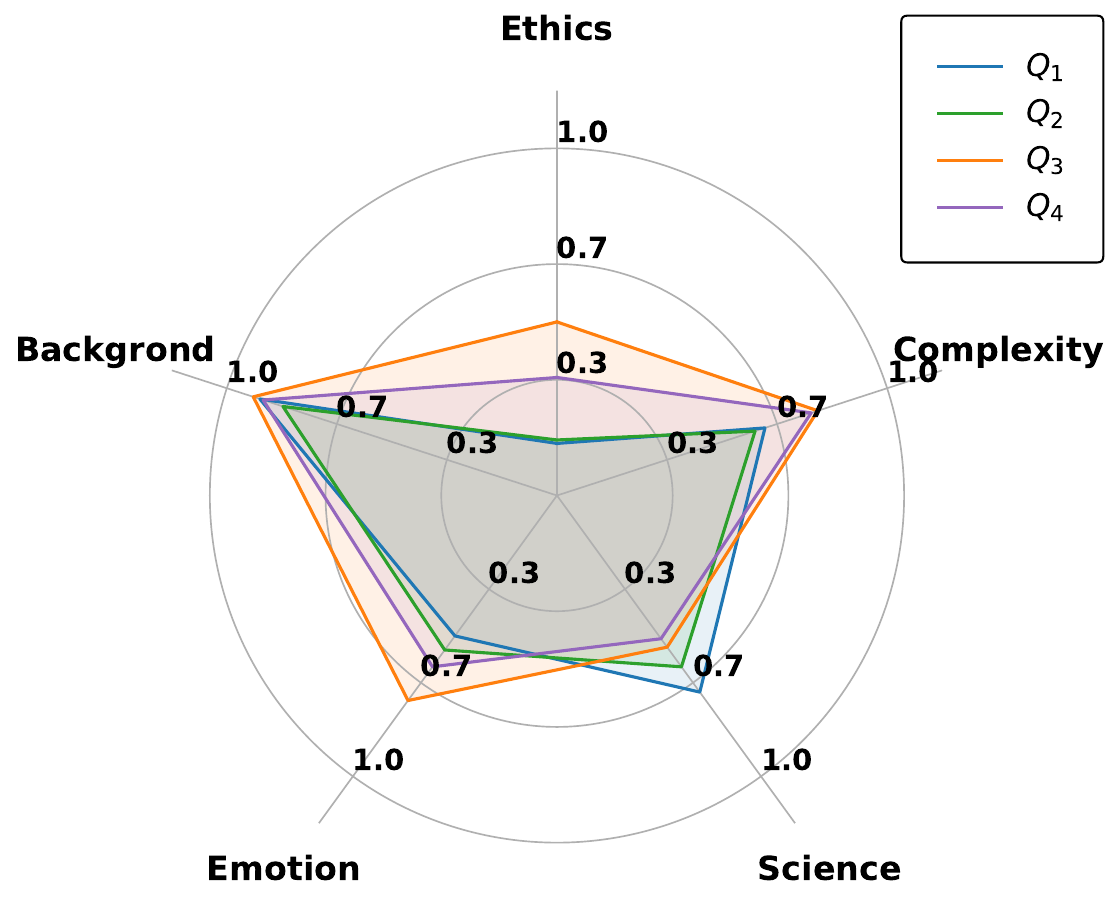}
        % \caption{GPT analysis2}
        \label{fig:GPT analysis2}
    \end{subfigure}
    \caption{Semantic analysis of different quadrants.}
    \label{fig:GPT analyses}
\end{figure}
\section{Related Works}
Multi-stage pretraining has emerged as a pivotal strategy in the development of LLMs, enabling models to better capture and utilize diverse data characteristics by dividing the training process into distinct phases\cite{pavlova2025multi,zhao2024slit,DBLP:journals/corr/abs-2107-14596,tan-etal-2022-msp}. \citeauthor{DBLP:journals/corr/abs-2107-14596} (\citeyear{DBLP:journals/corr/abs-2107-14596}) propose a multi-stage pretraining method that leverages various granularities of information, significantly boosting model performance. \citeauthor{tan-etal-2022-msp} (\citeyear{tan-etal-2022-msp}) explore the use of multi-stage prompting to improve translation tasks, demonstrating its effectiveness in enhancing downstream applications.

% In the realm of multimodal learning, models employ multi-stage pretraining to align and integrate different modalities, such as text and images, achieving superior representation learning\cite{JI2025129138}. Furthermore, the Instruction Pretraining framework illustrates the flexibility of multi-stage pretraining in supervised multitask environments, showcasing its capacity to augment massive raw corpora with instruction-response pairs\cite{cheng2024instructionpretraininglanguagemodels}.

% \subsection{Data Preprocessing for LLM Pretraining}
In LLM pretraining, data preprocessing is key to ensuring dataset quality\cite{duan2025enhancing}. Traditional methods use expert rules to filter low-quality data \cite{raffel2020exploring, rae2021scaling, laurenccon2022bigscience, together2023redpajama, penedo2024fineweb} and remove duplicates \cite{lee2022deduplicating, sorscher2022beyond, abbas2023semdedup, cerebras2023slimpajama, tirumala2024d4}. These approaches enhance quality but may lack semantic depth. To improve semantic selection, strategies involve using targeted sources or proxy models \cite{wenzek2020ccnet, xie2023data, marion2023less, thakkar2023self, engstrom2024dsdm, yu2024mates}. Classifiers automate selection, like the logistic regression model used by \citeauthor{du2022glam} (\citeyear{du2022glam}), with others employing complex scoring \cite{zhang2024autonomous, sachdeva2024train}. QuRating \cite{wettig2024qurating} uses multiple raters for nuanced evaluation.

Current methods focus on selection but overlook aligning data characteristics with learning stages, missing opportunities in data organization and sequencing to boost pretraining effectiveness.

\section{Conclusion}
% In this study, we introduce the \textbf{F}our-quad\textbf{RA}nt \textbf{M}ulti-stage pr\textbf{E}training \textbf{S}trategy (FRAME), a novel approach designed to boost the performance of LLMs by systematically organizing the pretraining process into four stages. Guided by the principle of achieving significant loss reductions four times, FRAME employs a strategic partitioning of pretraining data into four quadrants based on PPL and PD. The experimental results, showing a 9.03\% average improvement over random sampling, underscore the effectiveness of FRAME in optimizing pretraining data organization. 
% This work highlights the importance of a structured approach to data organization in the pretraining of LLMs, paving the way for further research into more refined and efficient pretraining strategies.
In this study, we propose the \textbf{F}our-quad\textbf{RA}nt \textbf{M}ulti-stage pr\textbf{E}training strategy (FRAME), a novel approach designed to boost the performance of LLMs by systematically organizing the pretraining process into four stages. Guided by the principle of achieving significant loss reductions four times, FRAME employs a strategic partitioning of pretraining data into four quadrants based on PPL and PD. The experimental results, showing a 16.8\% average improvement across MMLU and CMMLU over random sampling, underscore the effectiveness of FRAME in optimizing pretraining data organization.

\section{Limitations and Future Works}

In this study, we simply attempt to split the data into two parts based on the size of PPL and PD. Future research can be more detailed, such as subdividing a dimension into three parts or more, to explore more training stages. In addition, we plan to expand the scope of work to verify the applicability of the two-stage pretraining method in more language model architectures, such as Mamba\cite{gu2023mamba}, and Mixture of Experts\cite{wang2024scaling}. Although our evaluation criteria are already quite comprehensive, it is still possible to extend to a wider range of evaluations, including more detailed domain-specific or interactive tasks, such as the evaluation of analogy reasoning ability\cite{hu2023context}. Despite these potential limitations, we firmly believe that our research provides valuable insights and practical contributions to the academic community.

\bibliography{main}

\appendix

\section{Ethical Considerations}
We utilized publicly available training corpora from the internet to train our models, which inevitably included biased or harmful content, raising concerns about the safety of the model-generated content. To mitigate this issue, we prioritized the selection of high-quality datasets and implemented rigorous data-cleansing processes to remove harmful elements. Additionally, considering the computationally intensive nature of LLM training and its potential environmental impact, this paper explores a multi-stage training approach, aiming to enhance resource efficiency and reduce environmental pollution during the training process.

\begin{table}[b]
    \centering
    \small
    \begin{tabular}{l|c}
    \toprule
        \textbf{Hyperparameter} & \textbf{Value} \\ 
    \midrule
        Precision & bfloat16 \\ 
        Layers & 30 \\ 
        Hidden dimension & 1920 \\ 
        Attention heads & 15 \\ 
        Vocab size & 131072 \\ 
        Sequence length & 8192 \\ 
        Activation & SiLU \\ 
        Position embedding & RoPE \\ 
    \bottomrule
    \end{tabular}
    \caption{Model structure of 1.3B model}
    \label{tab:Model structure of 1.3B model}
\end{table}

\begin{table}[t]
    \centering
    \small
    \begin{tabular}{l|c}
    \toprule
        \textbf{Hyperparameter} & \textbf{Value} \\ 
    \midrule
        Precision & bfloat16 \\ 
        Layers & 32 \\ 
        Hidden dimension & 2560 \\ 
        Attention heads & 32 \\ 
        Vocab size & 131072 \\ 
        Sequence length & 8192 \\ 
        Activation & SiLU \\ 
        Position embedding & RoPE \\ 
    \bottomrule
    \end{tabular}
    \caption{Model structure of 3B model}
    \label{tab:Model structure of 3B model}
\end{table}

\section{Experimental Details}
\subsection{Experimental settings}
\label{subsec:Experimental settings}
\paragraph{Model configuration and training} We train reference models with 100M and 1.3B parameters on 500B tokens randomly sampled from the collected dataset, utilizing the Llama architecture \cite{touvron2023llama}. We use the 1.3B model to compute PPL and both the 100M and 1.3B models to calculate PD. For the main experiments, we train a 3B model. We set the batch size to 640 and the context window length to 8192. The initial learning rate is \(2 \times 10^{-4}\), with a warm-up phase of 375M tokens. We apply cosine learning rate scheduling with a weight decay of 0.1. We use the Adam optimizer to train the model within the Megatron framework \cite{shoeybi2019megatron}. Furthermore, we validate our approach on a 1.3B model, using the same settings as the 3B model. We use Ascend 910B NPU for training. For the 3B model, we use 512 NPUs for training, each model takes over 180 hours; For the 1.3B models, we use 96 NPUs, each model takes 6 hours.

In Table \ref{tab:Model structure of 1.3B model} and \ref{tab:Model structure of 3B model}, we present the model configuration of the 1.3B and 3B models.

\begin{table}[b]
    \centering
    \small
    \begin{tabular}{cccc}
    \toprule
        \textbf{Methods} & \textbf{Steps} & \textbf{MMLU} & \textbf{CMMLU} \\
    \midrule
        Random & \multirow{3}{*}{30K} & 25.4 & 25.9 \\ 
        $\text{A}_{\text{PPL}}^{\text{low}} \to \text{A}_{\text{PPL}}^{\text{high}}$ & ~ & \textbf{26.7} & \textbf{26.2} \\
        $\text{A}_{\text{PPL}}^{\text{high}} \to \text{A}_{\text{PPL}}^{\text{low}}$ & ~ & 25.3 & 25.5 \\ 
    \midrule
        Random & \multirow{3}{*}{60K} & 25.9 & 25.8 \\ 
        $\text{A}_{\text{PPL}}^{\text{low}} \to \text{A}_{\text{PPL}}^{\text{high}}$ & ~ & 25.5 & 25.5 \\ 
        $\text{A}_{\text{PPL}}^{\text{high}} \to \text{A}_{\text{PPL}}^{\text{low}}$ & ~ & \textbf{33.4} & \textbf{35.4} \\ 
    \midrule
        Random & \multirow{3}{*}{95K} & 24.8 & 25.6 \\ 
        $\text{A}_{\text{PPL}}^{\text{low}} \to \text{A}_{\text{PPL}}^{\text{high}}$ & ~ & 26.0 & 26.0 \\ 
        $\text{A}_{\text{PPL}}^{\text{high}} \to \text{A}_{\text{PPL}}^{\text{low}}$ & ~ & \textbf{39.6} & \textbf{42.6} \\ 
    \bottomrule
    \end{tabular}
    \caption{Accuracy on MMLU and CMMLU for two-stage pretraining across different steps based on PPL with 3B models.}
    \label{tab:PPL_MMLU_CMMLU}
\end{table}

\paragraph{Baselines}
% In 3B model setting, we compare FRAME with several baselines:
% \begin{itemize}
%     \item \textbf{Random}: % 以完全随机的形式组织数据顺序。
%     \item \textbf{PDPC \cite{zhang2025preferencecurriculumllmspretrained}}: % 使用PD作为模型感知的课程学习指标，并利用探索出的符合模型偏好的学习曲线对数据顺序进行组织。
%     \item \textbf{\(Q_3 \rightarrow Q_1 \rightarrow Q_4 \rightarrow Q_2\)}: % 采用与FRAME相同的平滑操作，但是优先满足PD相关的约束条件，来组织数据顺序。
% \end{itemize}
In the 3B model setting, we compare FRAME with several baselines:
\begin{itemize}
    \item \textbf{Random}: Data is organized in a completely random order, meaning there is no specific sequence or strategy for data input
    \item \textbf{PDPC} \cite{zhang2025preferencecurriculumllmspretrained}: Utilizes PD as a model-aware curriculum learning indicator, organizing the data sequence through a preference function that aligns with the model's inherent preferences.
    \item \textbf{\(Q_3 \rightarrow Q_1 \rightarrow Q_4 \rightarrow Q_2\)}: Adopts the same smoothing process as FRAME but prioritizes satisfying PD-related constraints to organize the data order.
\end{itemize}

In the 1.3B model setting, we also compare FRAME with other baselines:
\begin{itemize}
    \item \textbf{Sequential}: Organizes data by directly sorting according to PPL, PD, and QuRating \cite{wettig2024qurating}, in either ascending or descending order. % 按照PPL、PD以及QuRating指标直接对数据进行排序，以增大或减少的顺序
    \item \textbf{Preference CL}: Utilizes the same preference function as PDPC but replaces the indicators with PPL or QuRating. We adopt both S-shape and S-shape reverse function.% 采用和PDPC一样的偏好函数，但是将指标换为PPL或PD
\end{itemize}
% we also compare it with models trained by sequencing based on the PPL, PD, and QuRating \cite{wettig2024qurating} metrics.

\begin{table}[t]
    \centering
    \small
    \begin{tabular}{cccc}
    \toprule
        \textbf{Methods} & \textbf{Steps} & \textbf{MMLU} & \textbf{CMMLU} \\ 
    \midrule
        Random & \multirow{2}{*}{30K} & 25.4 & 25.9 \\ 
        $\text{A}_{\text{PD}}^{\text{low}} \to \text{A}_{\text{PD}}^{\text{high}}$ & ~ & \textbf{26.7} & \textbf{26.2} \\ 
    \midrule
        Random & \multirow{2}{*}{60K} & 25.9 & \textbf{25.8} \\ 
        $\text{A}_{\text{PD}}^{\text{low}} \to \text{A}_{\text{PD}}^{\text{high}}$ & ~ & \textbf{26.2} & 25.1 \\
    \midrule
        Random & \multirow{2}{*}{95K} & 24.8 & 25.6 \\ 
        $\text{A}_{\text{PD}}^{\text{low}} \to \text{A}_{\text{PD}}^{\text{high}}$ & ~ & \textbf{26.9} & \textbf{27.2} \\ 
    \bottomrule
    \end{tabular}
    \caption{Accuracy on MMLU and CMMLU for two-
stage pretraining based on PD with 3B models.}
    \label{tab:PD_MMLU_CMMLU}
\end{table}

\begin{figure}[b]
    \centering
    \includegraphics[width=0.7\linewidth]{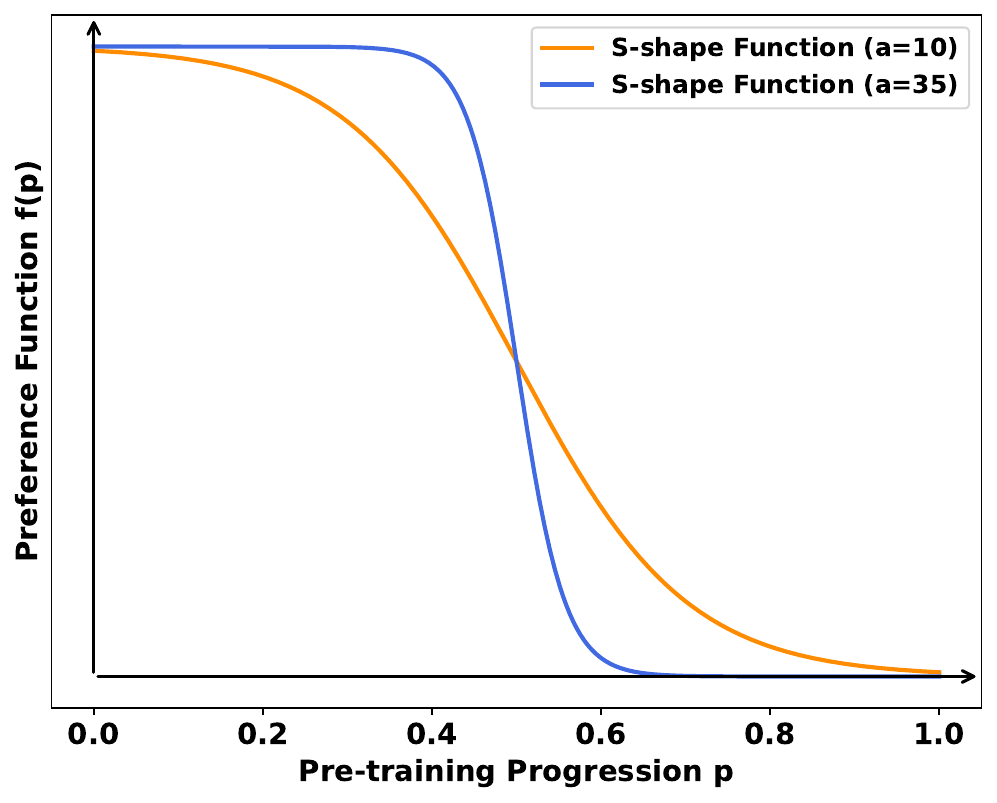}
    \caption{S-shape functions with a=10 and a=35.}
    \label{fig:S-shape}
\end{figure}

\begin{figure*}[t]
    \centering
    \includegraphics[width=1.0\linewidth]{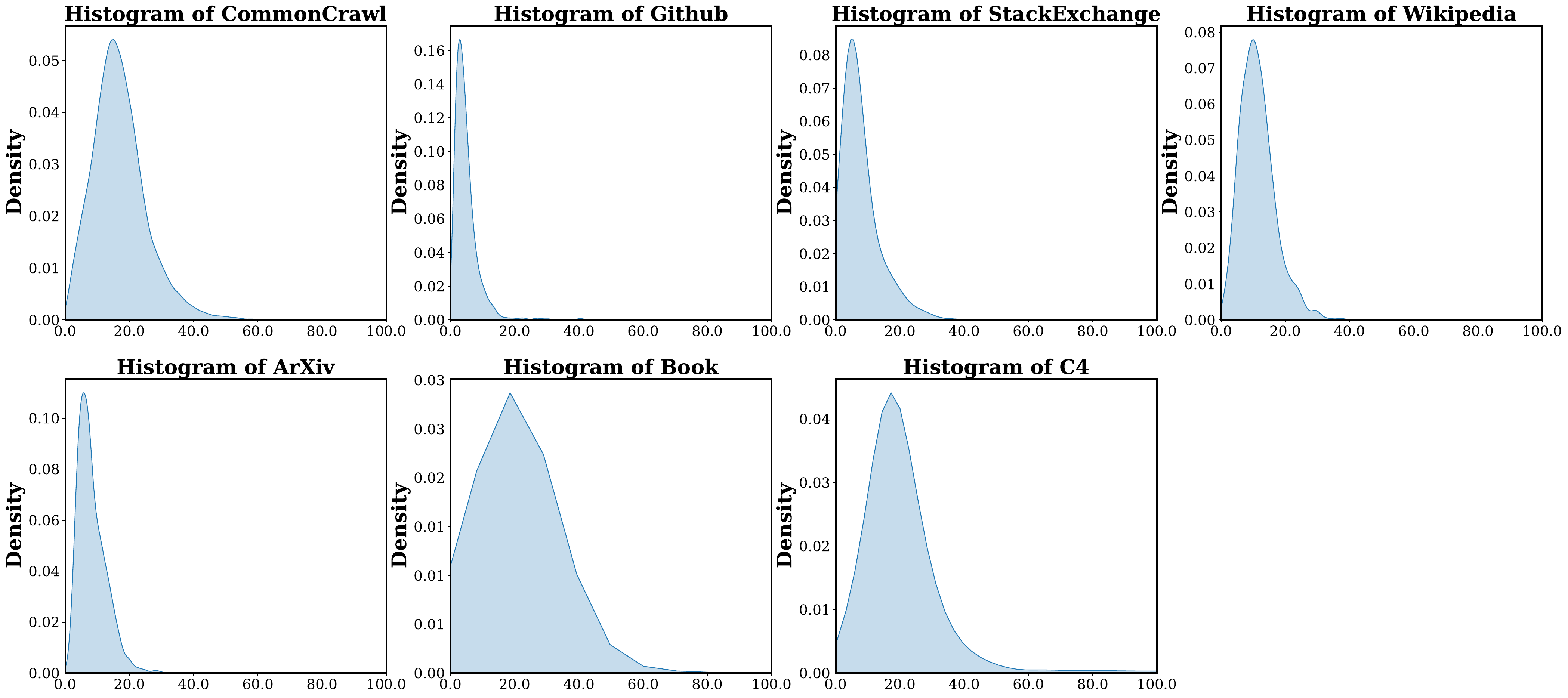}
    \caption{PPL distribution across different sources.}
    \label{fig:PPL Distribution}
\end{figure*}

\begin{figure*}[t]
    \centering
    \includegraphics[width=1.0\linewidth]{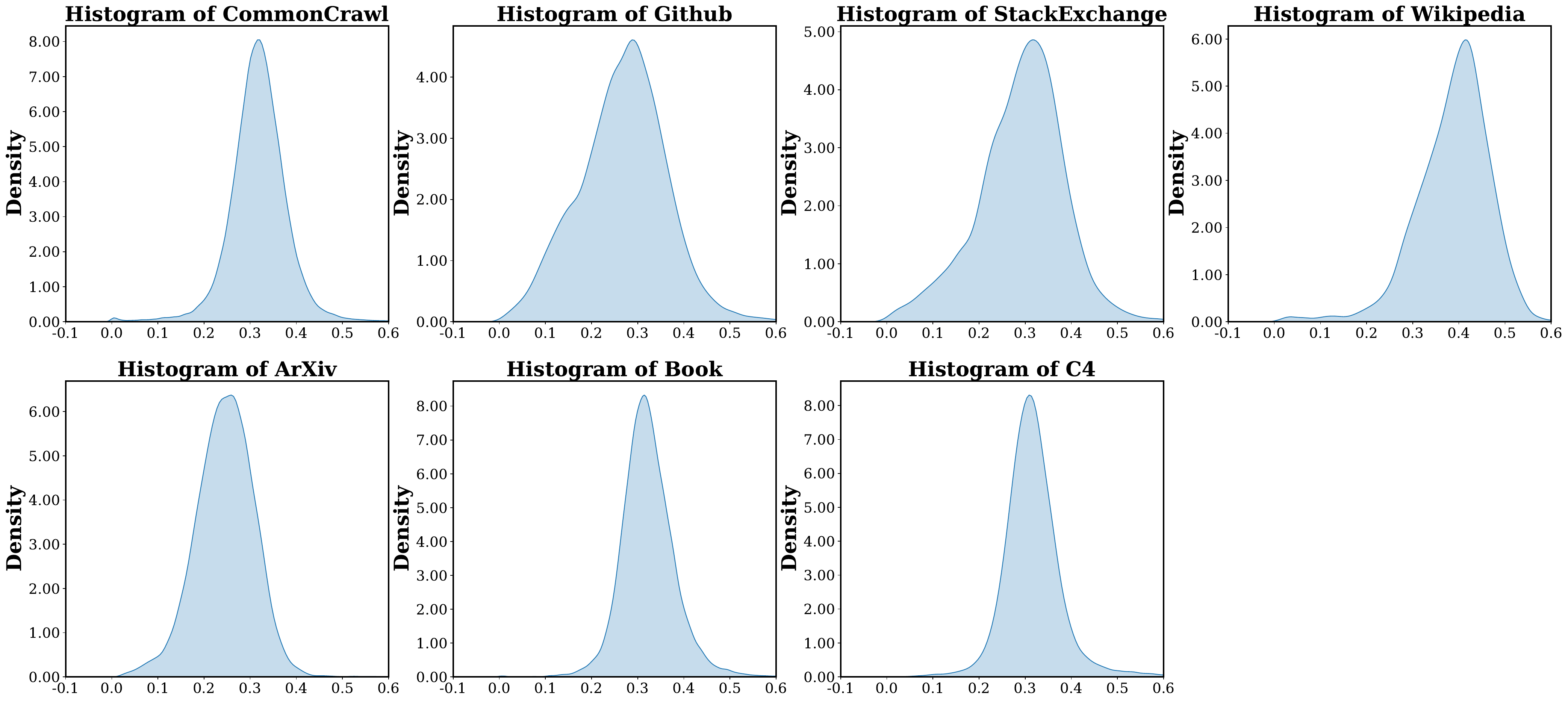}
    \caption{PD Distribution across different sources.}
    \label{fig:PD Distribution}
\end{figure*}

\subsection{Benchmark Accuracy of Two-stage Pretraining Based on PPL} \label{PPL_2stage}
Table \ref{tab:PPL_MMLU_CMMLU} shows the model's benchmark accuracy at various steps for the settings \( \text{A}_{\text{PPL}}^{\text{low}} \rightarrow \text{A}_{\text{PPL}}^{\text{high}} \) and \( \text{A}_{\text{PPL}}^{\text{high}} \rightarrow \text{A}_{\text{PPL}}^{\text{low}} \). Training on high PPL data followed by low PPL data initially yields lower performance than random training at 30K steps, but accuracy improves significantly in the later phases at 60K and 95K steps. In contrast, in the \( \text{A}_{\text{PPL}}^{\text{low}} \rightarrow \text{A}_{\text{PPL}}^{\text{high}} \) setting, the model shows slight improvement initially but eventually aligns with the random sequence results. This leads to our first key finding: training first on high PPL data followed by low PPL data can cause the loss to drop significantly twice, ultimately boosting model performance.

\subsection{Benchmark Accuracy of Two-stage Pretraining Based on PD}
\label{PD_2stage}
Table \ref{tab:PD_MMLU_CMMLU} shows the model's accuracy as training steps increase under the \( \text{A}_{\text{PD}}^{\text{low}} \rightarrow \text{A}_{\text{PD}}^{\text{high}} \) setting. The model consistently outperforms the Random model at most steps. Notably, at 95K steps, it exceeds the Random setting by 2.1\% on MMLU and 1.6\% on CMMLU, which validates PD as an effective metric. This leads to our second key finding: training first on low PD data followed by high PD data can cause the loss to drop significantly twice, ultimately boosting model performance.

\subsection{S-shape Functions with a=10 and a=35}\label{sec:s-shape}
By applying the S-shape function, we gradually decrease the proportion of \(Q_3\) data while increasing the proportion of \(Q_4\) data, resulting in the mixed data \(S_{34}\). Figure \ref{fig:S-shape} illustrates the two forms of Equation \ref{eq:sigmoid} with \(a=10\) and \(a=35\). In this paper, we adopt the steeper form with \(a=35\) for smoothing transitions between different stages. In the early stages of model training, the proportion of data from the first stage is close to 1. As training progresses, this proportion approaches 0. During the mid-training phase, there is a gradual transition between the data from the first and second stages.

\begin{figure*}[t]
    % \centering
    \begin{subfigure}{0.48\linewidth}
        \centering
        \includegraphics[width=\linewidth]{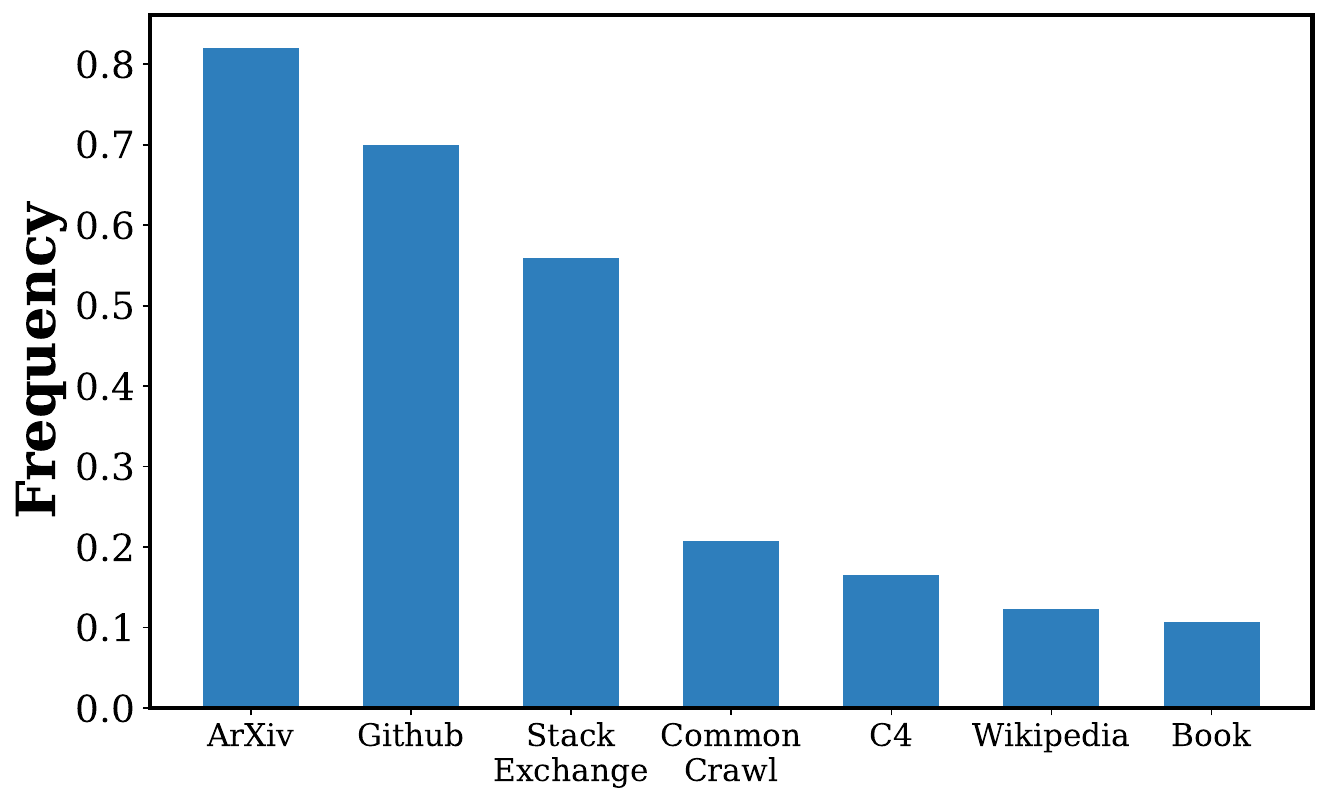}
        \caption{$Q_1$}
        \label{fig:data_distribution_A1}
    \end{subfigure}
    \hfill
    \begin{subfigure}{0.48\linewidth}
        \centering
        \includegraphics[width=\linewidth]{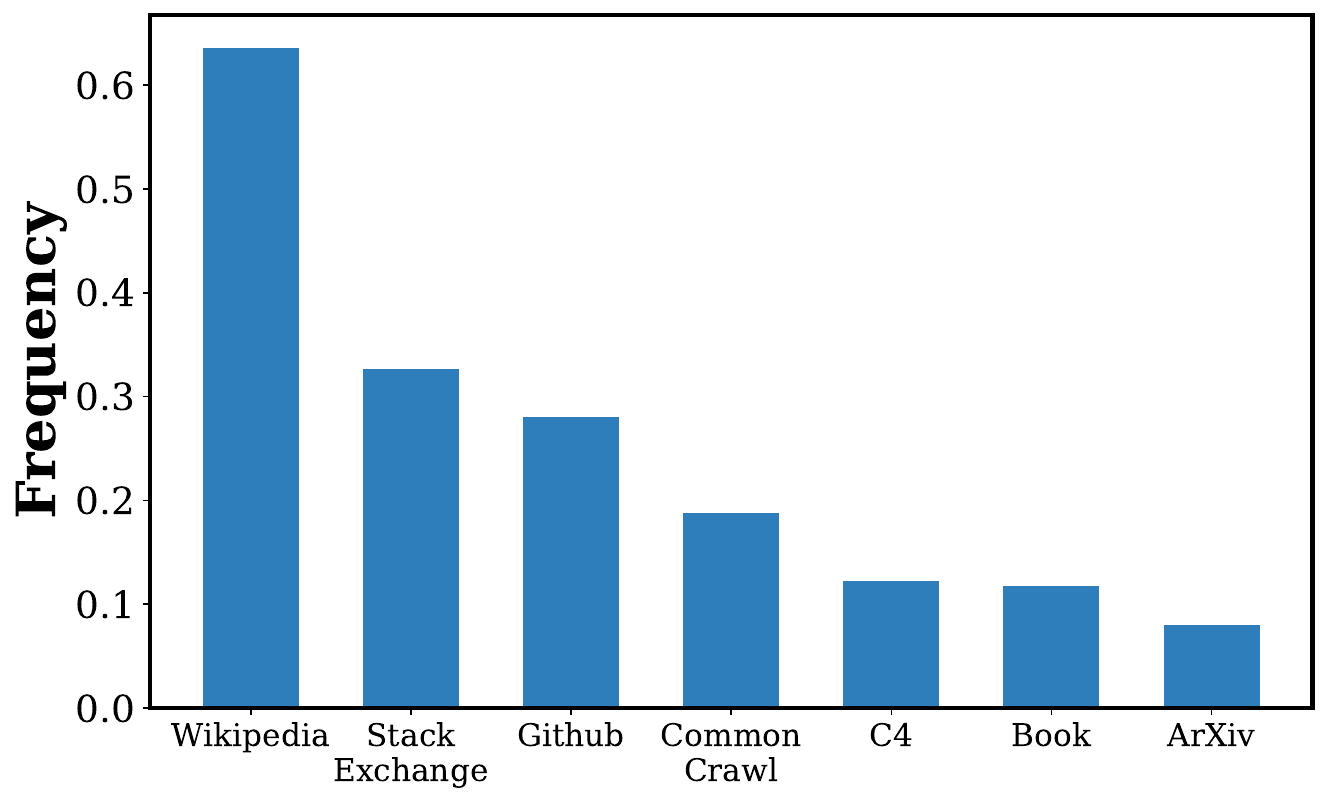}
        \caption{$Q_2$}
        \label{fig:data_distribution_A2}
    \end{subfigure}
    \\
    \begin{subfigure}{0.48\linewidth}
        \centering
        \includegraphics[width=\linewidth]{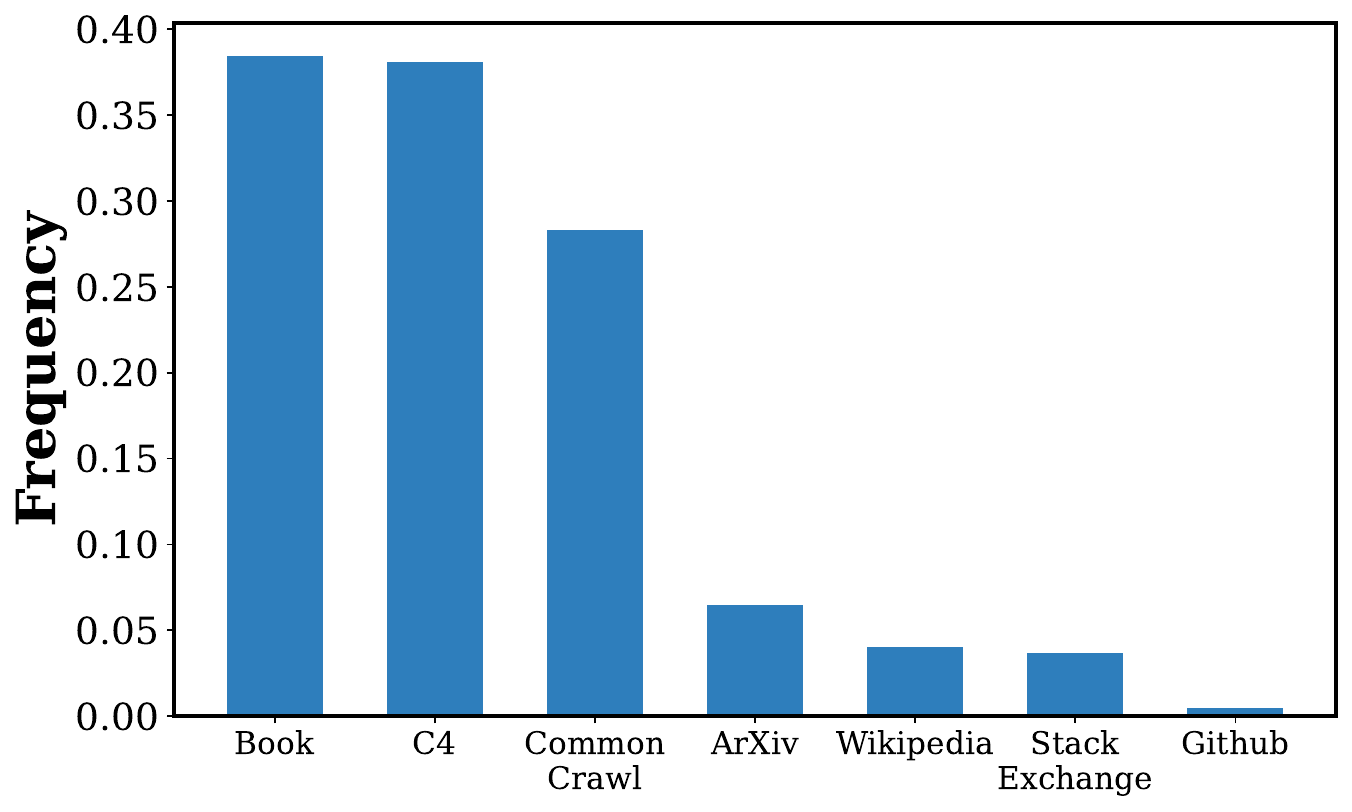}
        \caption{$Q_3$}
        \label{fig:data_distribution_A3}
    \end{subfigure}
    \hfill
    \begin{subfigure}{0.48\linewidth}
        \centering
        \includegraphics[width=\linewidth]{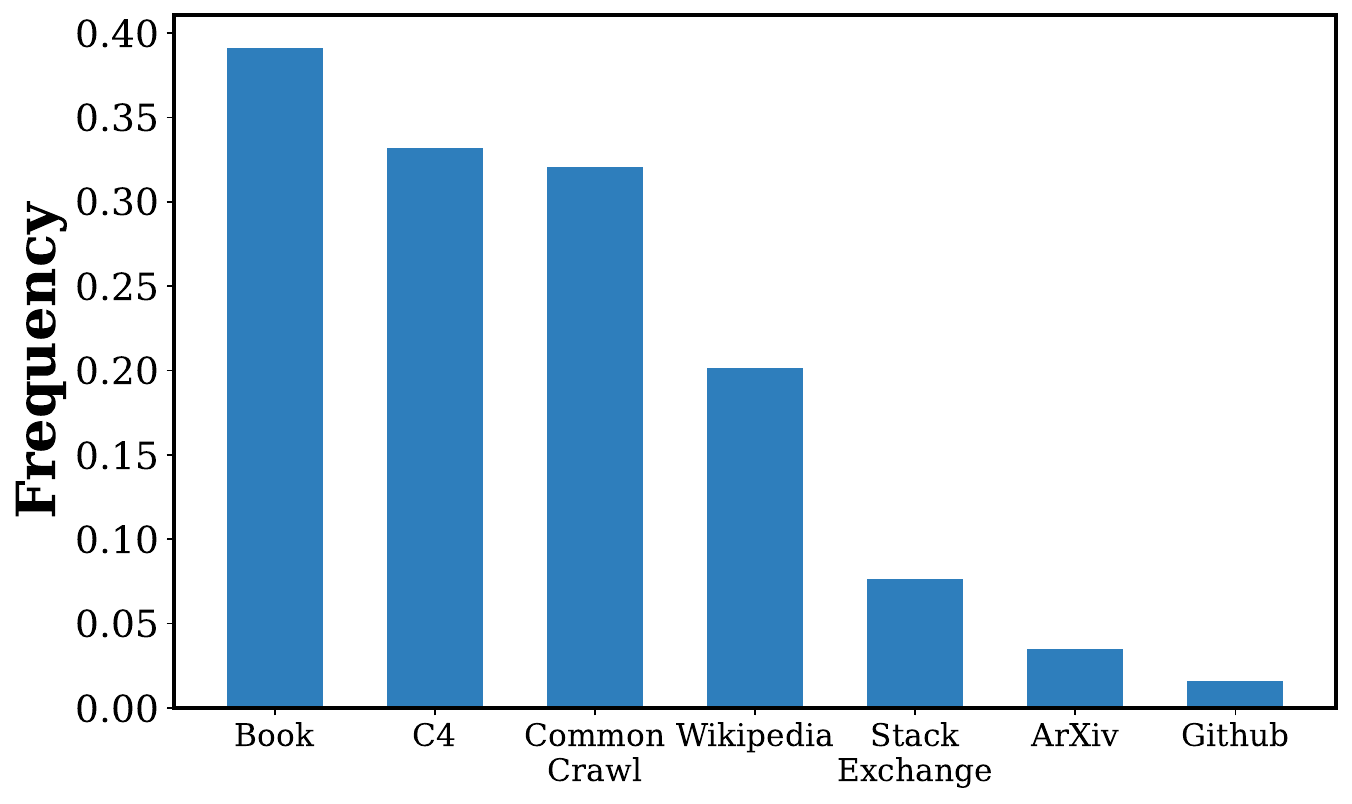}
        \caption{$Q_4$}
        \label{fig:data_distribution_A4}
    \end{subfigure}
    \caption{Data distribution details. We've rearranged the order based on the frequency of different sources appearing in each quadrant.}
    \label{fig:data distribution}
\end{figure*}

\subsection{Data Distribution} 
\label{sec:Data Distribution}
Figure \ref{fig:PPL Distribution} and \ref{fig:PD Distribution} show the PPL and PD distribution of samples in our training set. The PPL of data from different domains shows significant variations, with mean values ranging from 5 to 22 and varying degrees of variance. Sorting data by PPL may lead to imbalanced domain representation between early and late training stages, exposing the model to overly homogeneous data at each stage and potentially degrading performance. In contrast, the distribution of PD is quite similar across different data sources, ensuring that each training phase includes diverse data, thereby maintaining pretraining efficiency and enhancing model performance.

Figure \ref{fig:data distribution} illustrates the domain distribution of data across the four quadrants, arranged in descending order of proportion. Quadrants 1 and 2 predominantly contain data related to code (sourced from Github and Stack Exchange) and knowledge (sourced from Arxiv and Wikipedia). In contrast, the data in Quadrants 3 and 4 primarily originate from books and the Common Crawl.

\subsection{Additional evaluation results}
\begin{figure*}[t]
    % \centering
    \begin{subfigure}{0.48\linewidth}
        \centering
        \includegraphics[width=\linewidth]{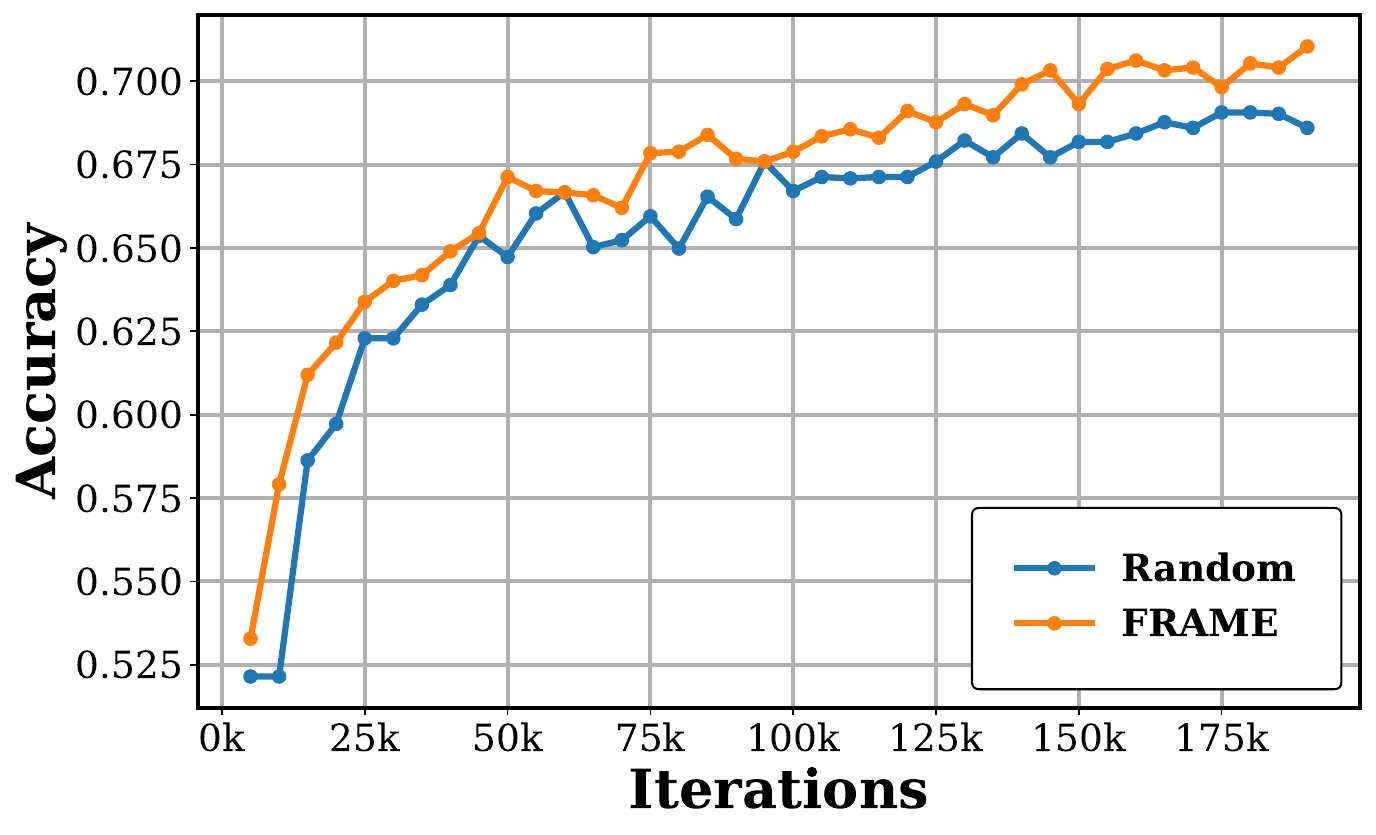}
        \caption{ARC-E}
        \label{fig:3b_arc-e}
    \end{subfigure}
    \hfill
    \begin{subfigure}{0.48\linewidth}
        \centering
        \includegraphics[width=\linewidth]{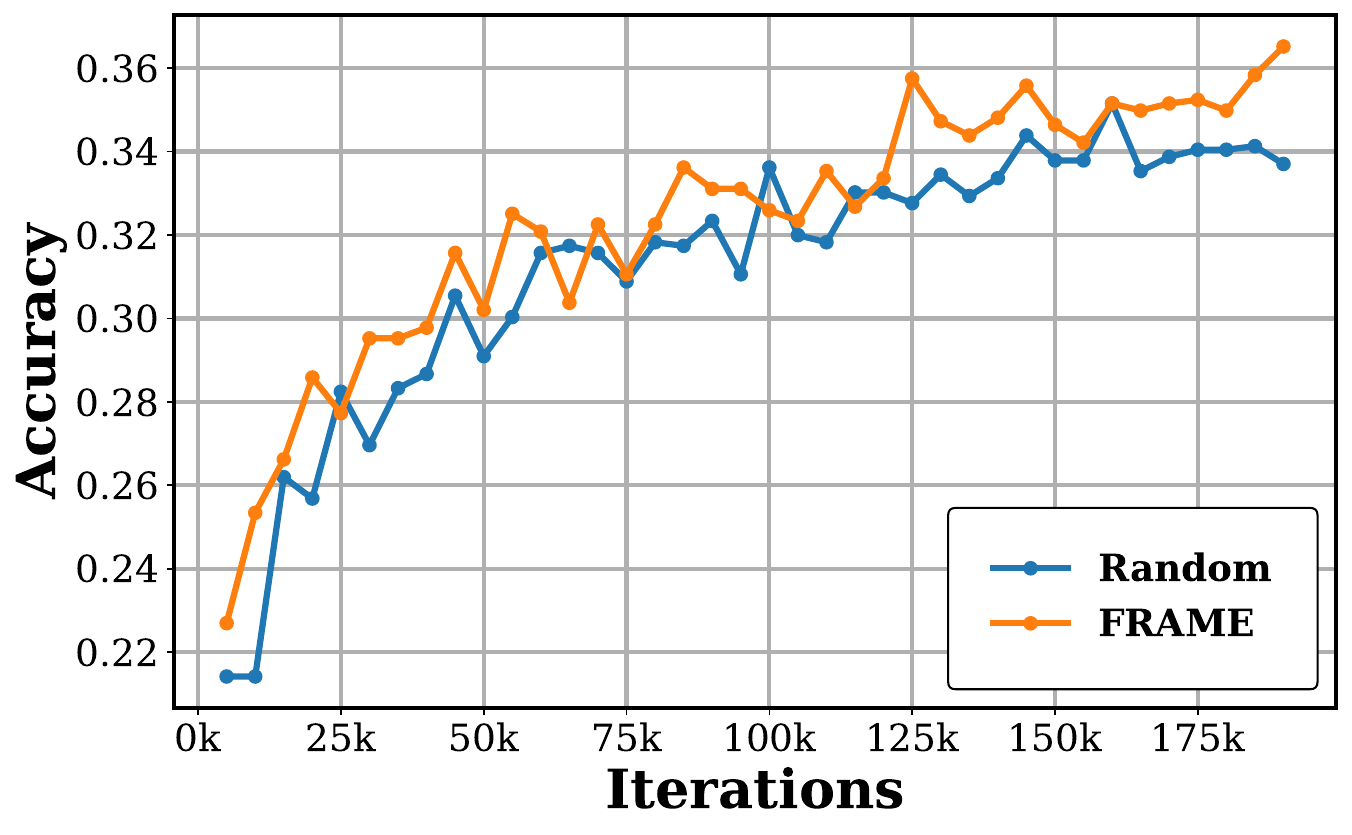}
        \caption{ARC-C}
        \label{fig:3b_arc-c}
    \end{subfigure}
    \\
    \begin{subfigure}{0.48\linewidth}
        \centering
        \includegraphics[width=\linewidth]{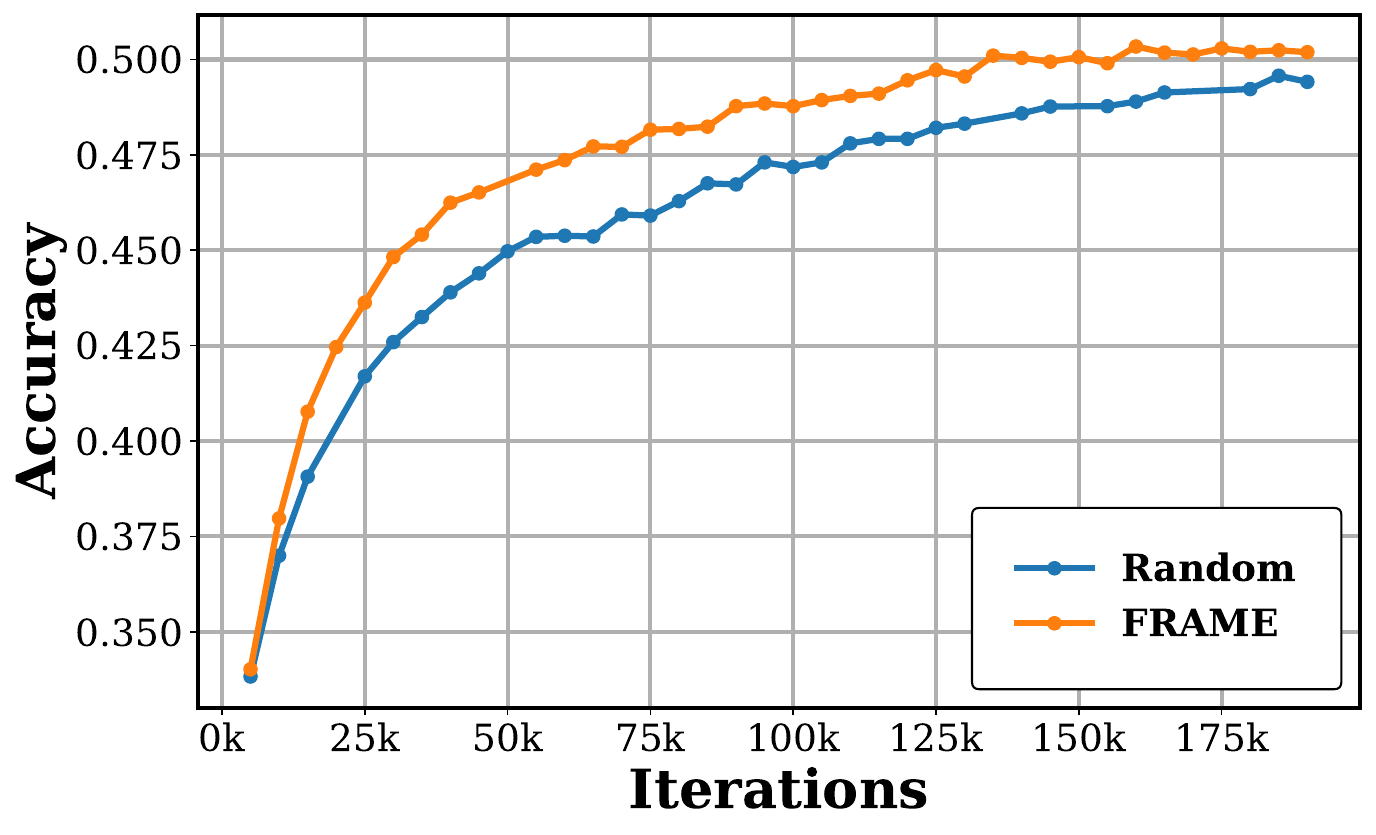}
        \caption{HellaSwag}
        \label{fig:3b_hellaswag}
    \end{subfigure}
    \hfill
    \begin{subfigure}{0.48\linewidth}
        \centering
        \includegraphics[width=\linewidth]{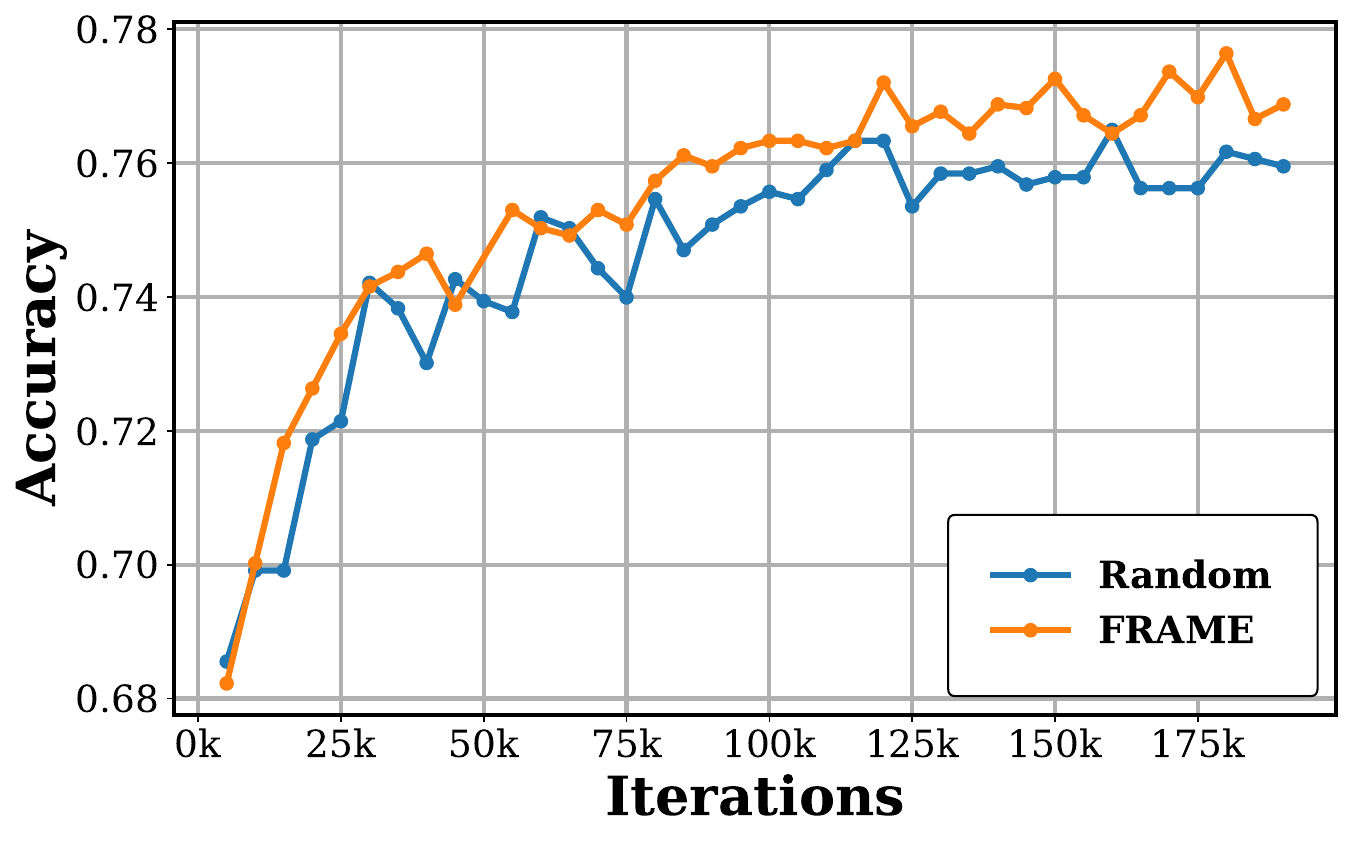}
        \caption{PIQA}
        \label{fig:3b_piqa}
    \end{subfigure}
    \\
    \begin{subfigure}{0.48\linewidth}
        \centering
        \includegraphics[width=\linewidth]{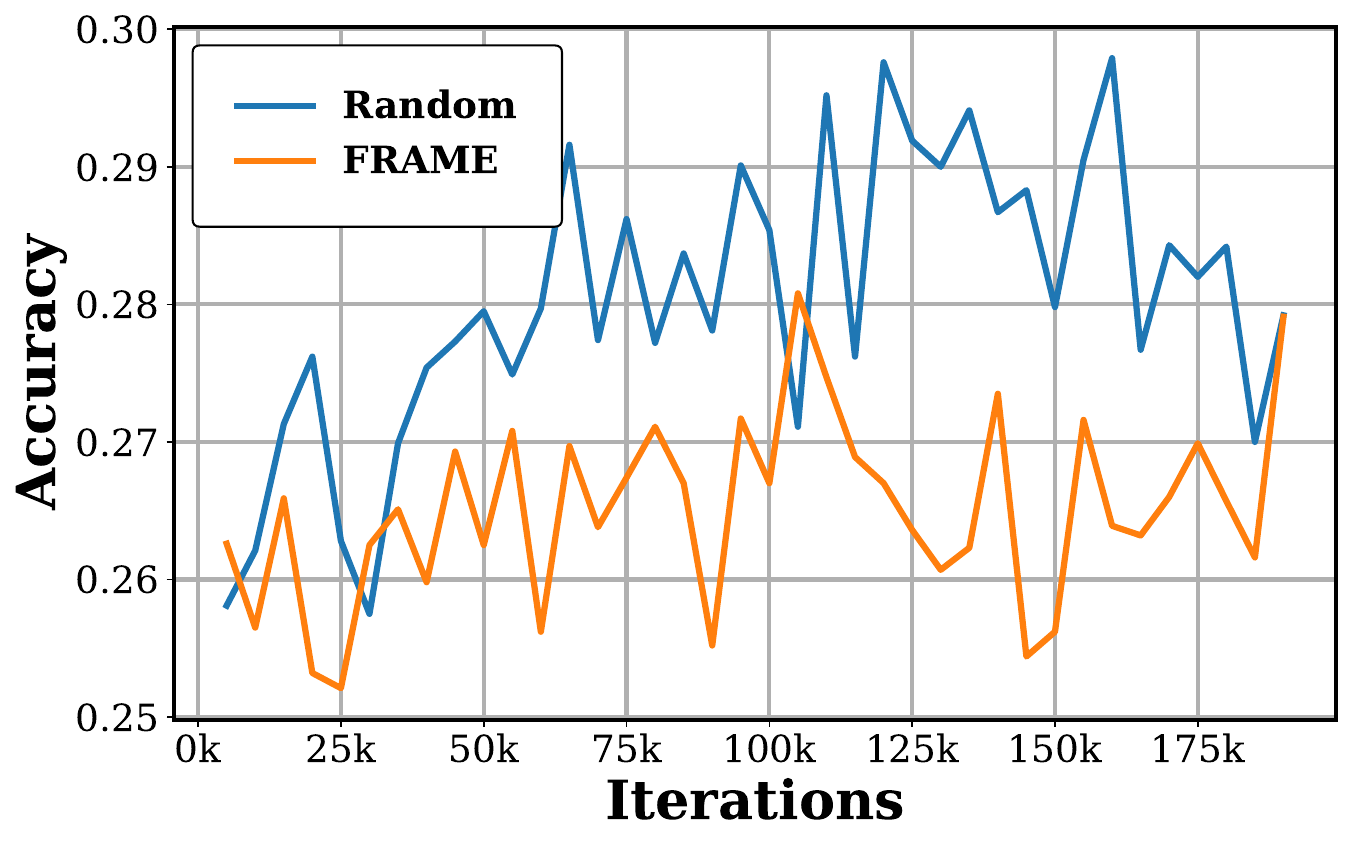}
        \caption{BBH}
        \label{fig:3b_bbh}
    \end{subfigure}
    \caption{Performance of downstream tasks on 3B models with respect to training iterations. We compare FRAME with Random baseline.}
    \label{fig:more_benchmark_details}
\end{figure*}

In Figure \ref{fig:more_benchmark_details}, we present additional evaluation results on 3B models. It is evident that in the majority of benchmarks, FRAME significantly outperforms Random, underscoring the effectiveness of our proposed method.

% \twocolumn
\subsection{Smoother LOSS Curve from FRAME}
\label{subsec:Smoother LOSS Curve from FRAMES}
% To quantitatively analyze the characteristics of the training loss curve, we applied spectral analysis to assess its smoothness. Specifically, we performed a Fast Fourier Transform (FFT) on the loss curve \( l[n] \) as \( Y[k] = \text{FFT}{ l[n] } \) to convert it to the frequency domain and computed the Power Spectral Density (PSD) \( PSD[k] = \frac{|Y[k]|^2}{N} \) to analyze the energy distribution across different frequency components. We selected a cutoff frequency \( f_c \) to divide the spectrum into low-frequency and high-frequency parts and calculated the proportion of high-frequency energy \( R = \frac{\sum_{f > f_c} PSD[k]}{\sum_{k=0}^{N/2} PSD[k]} \). A smaller ratio \( R \) indicates fewer high-frequency components and a smoother temporal curve.
To analyze the training loss curve's smoothness, we use spectral analysis. We perform a Fast Fourier Transform (FFT) on the loss curve \( l[n] \) as \( Y[k] = \text{FFT}(l[n]) \) to convert it to the frequency domain and compute the Power Spectral Density \( PSD[k] = \frac{|Y[k]|^2}{N} \) to examine energy distribution across frequencies. By selecting a cutoff frequency \( f_c \), we divide the spectrum into low and high frequencies and calculate the high-frequency energy proportion \( R = \frac{\sum_{f > f_c} PSD[k]}{\sum_{k=0}^{N/2} PSD[k]} \). A smaller \( R \) indicates a smoother temporal curve with fewer high-frequency components.

% Our analysis revealed that the loss curve from the four-quadrant learning approach has the lowest high-frequency energy proportion at 0.02\%, which is significantly lower than that of single-dimension learning and random training methods. This finding suggests that our proposed training method enables the model to converge with a more stable trend, reduces the impact of gradient fluctuations during training, maintains result stability, and enhances generalization. This improvement is attributed to our four-stage training strategy, which groups data with certain commonalities into consecutive batches and incorporates a clustering operation within each batch, thereby stabilizing the model training process.
Our analysis reveals that the loss curve from FRAME has the lowest high-frequency energy proportion at 0.02\%, significantly lower than that of PDPC and Random. This suggests that FRAME allows the model to converge more stably, and reduces the impact of gradient fluctuations during training. This improvement is due to our four-stage training strategy, which organizes similar data into consecutive batches, thus stabilizing the model training process.

\subsection{Ablation Studies}
\label{subsec:Ablation Studies}
\begin{figure*}[t]
    % \centering
    \begin{subfigure}{0.48\linewidth}
        \centering
        \includegraphics[width=\linewidth]{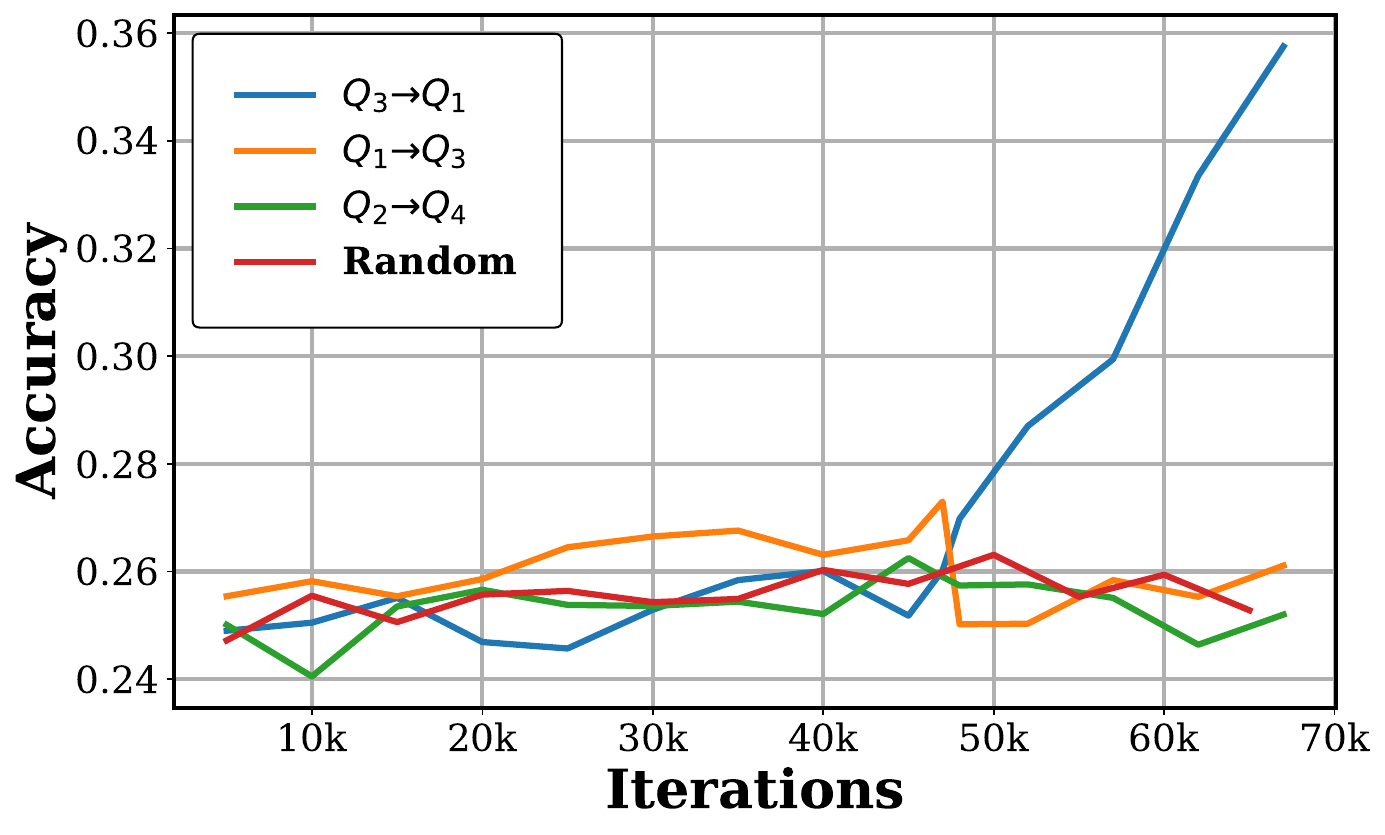}
        \caption{MMLU}
        \label{fig:PPL ablation study1_2}
    \end{subfigure}
    \hfill
    \begin{subfigure}{0.48\linewidth}
        \centering
        \includegraphics[width=\linewidth]{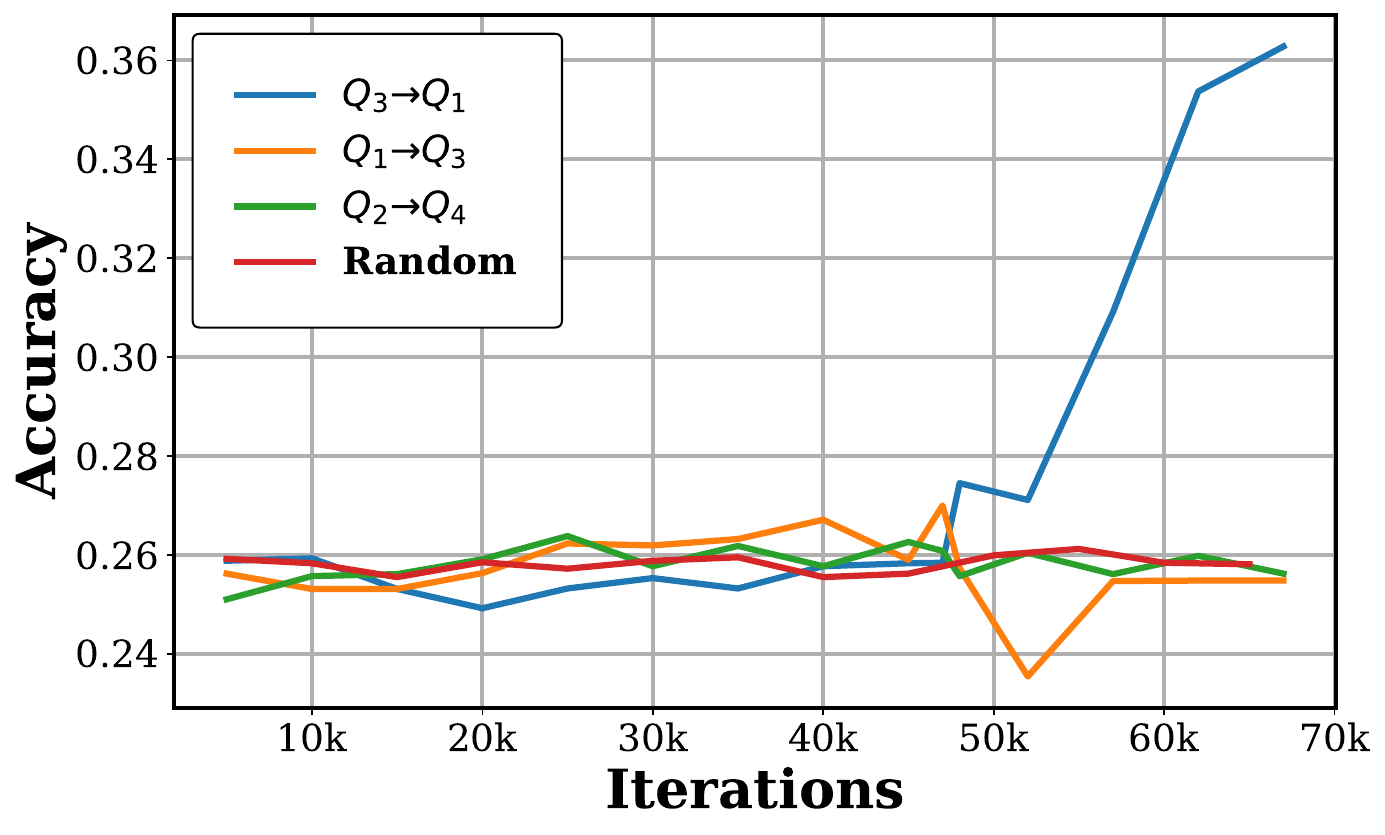}
        \caption{CMMLU}
        \label{fig:PPL ablation study1_3}
    \end{subfigure}
    \\
    \begin{subfigure}{0.48\linewidth}
        \centering
        \includegraphics[width=\linewidth]{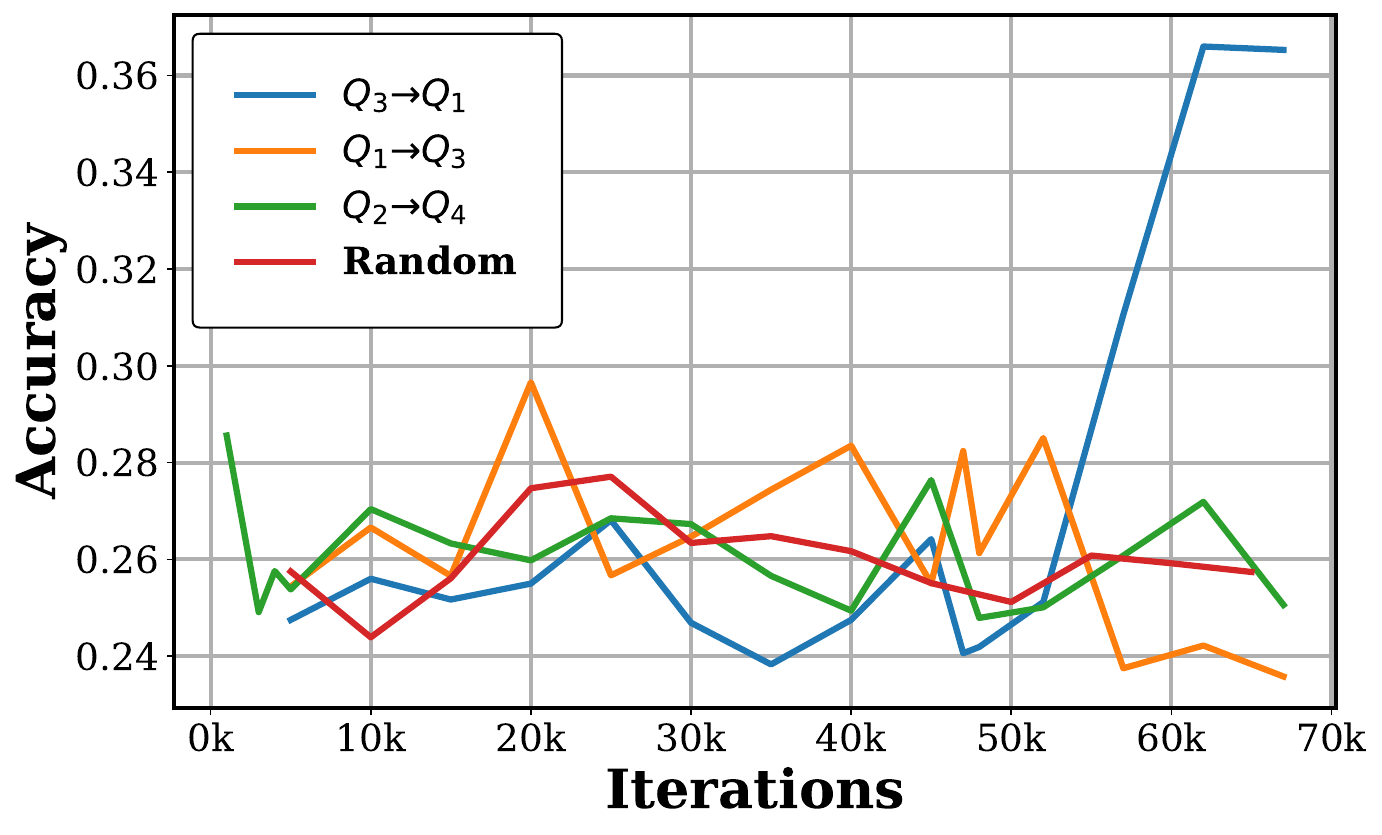}
        \caption{CEVAL}
        \label{fig:PPL ablation study1_4}
    \end{subfigure}
    \hfill
    \begin{subfigure}{0.48\linewidth}
        \centering
        \includegraphics[width=\linewidth]{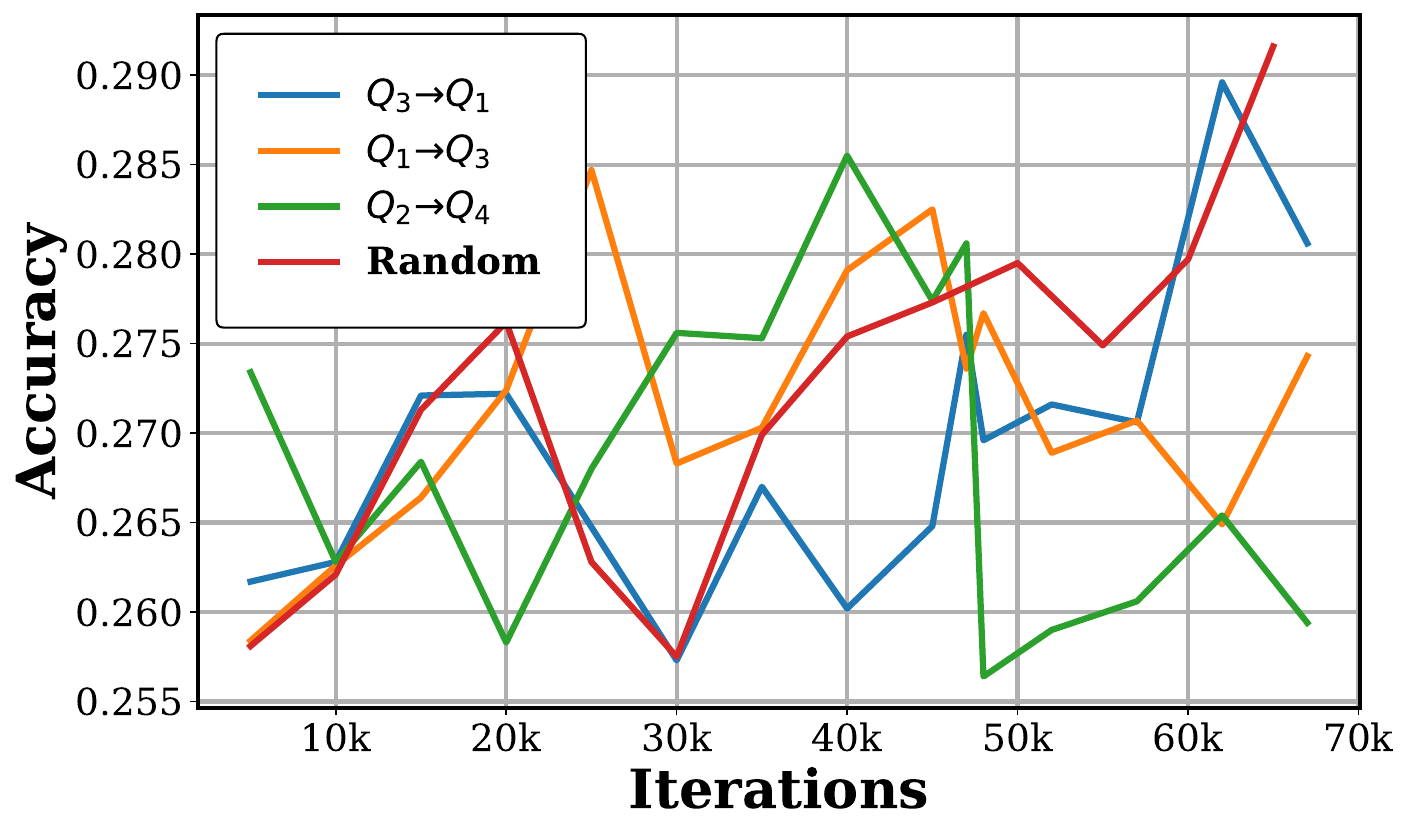}
        \caption{BBH}
        \label{fig:PPL ablation study1_1}
    \end{subfigure}
    \caption{PPL related ablation study (2 quadrants).}
    \label{fig:PPL ablation study1}
\end{figure*}

\begin{figure*}[t]
    % \centering
    \begin{subfigure}{0.48\linewidth}
        \centering
        \includegraphics[width=\linewidth]{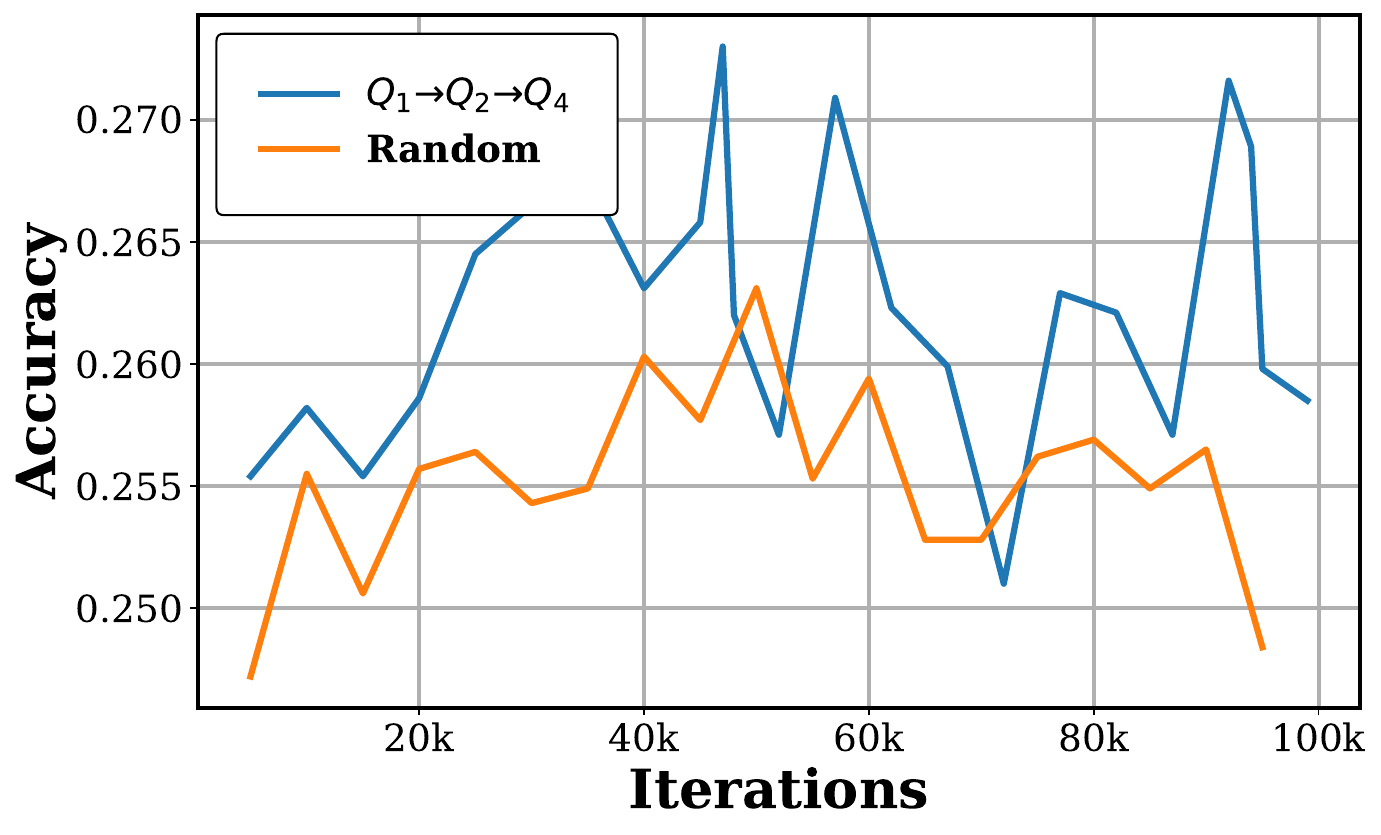}
        \caption{MMLU}
        \label{fig:PPL ablation study2_2}
    \end{subfigure}
    \hfill
    \begin{subfigure}{0.48\linewidth}
        \centering
        \includegraphics[width=\linewidth]{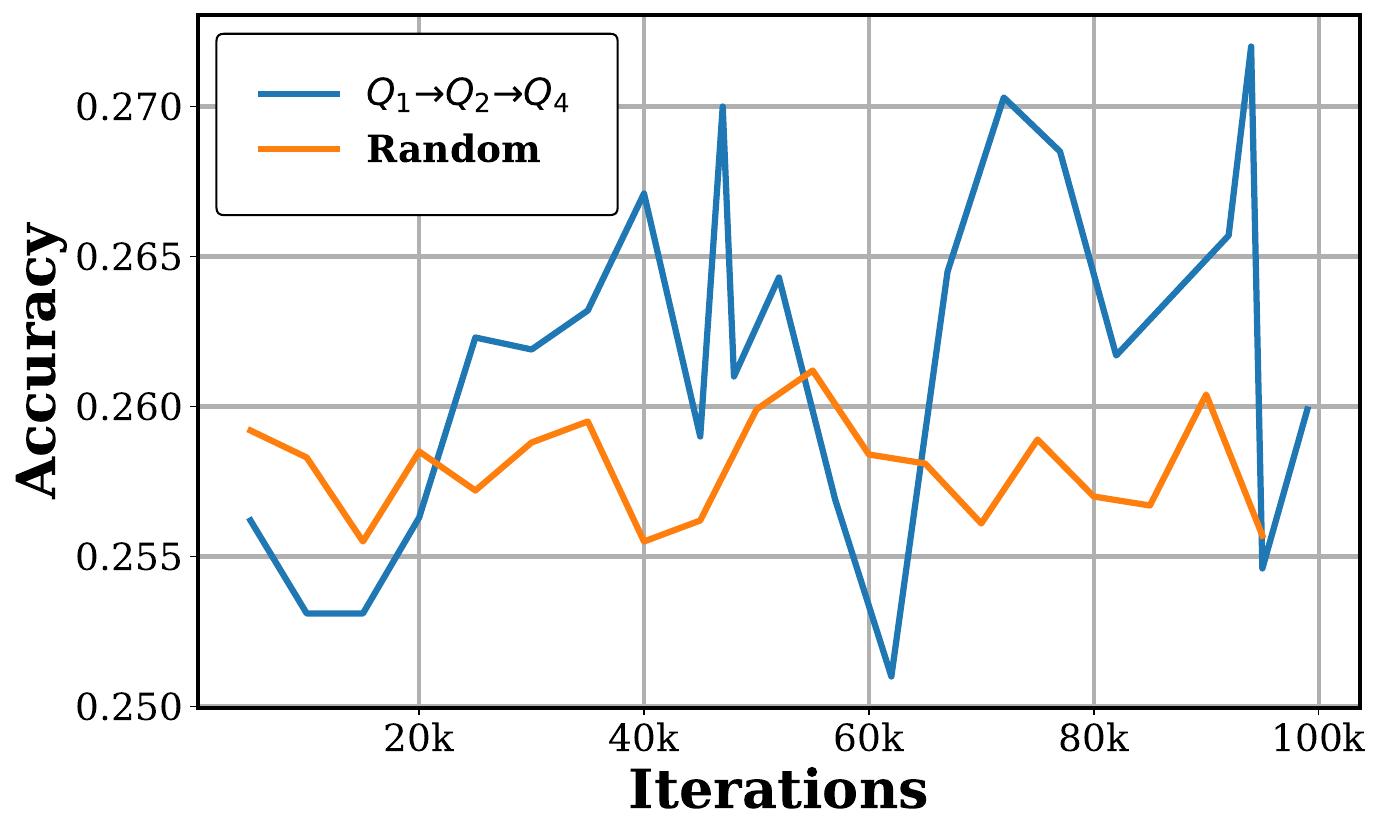}
        \caption{CMMLU}
        \label{fig:PPL ablation study2_3}
    \end{subfigure}
    \\
    \begin{subfigure}{0.48\linewidth}
        \centering
        \includegraphics[width=\linewidth]{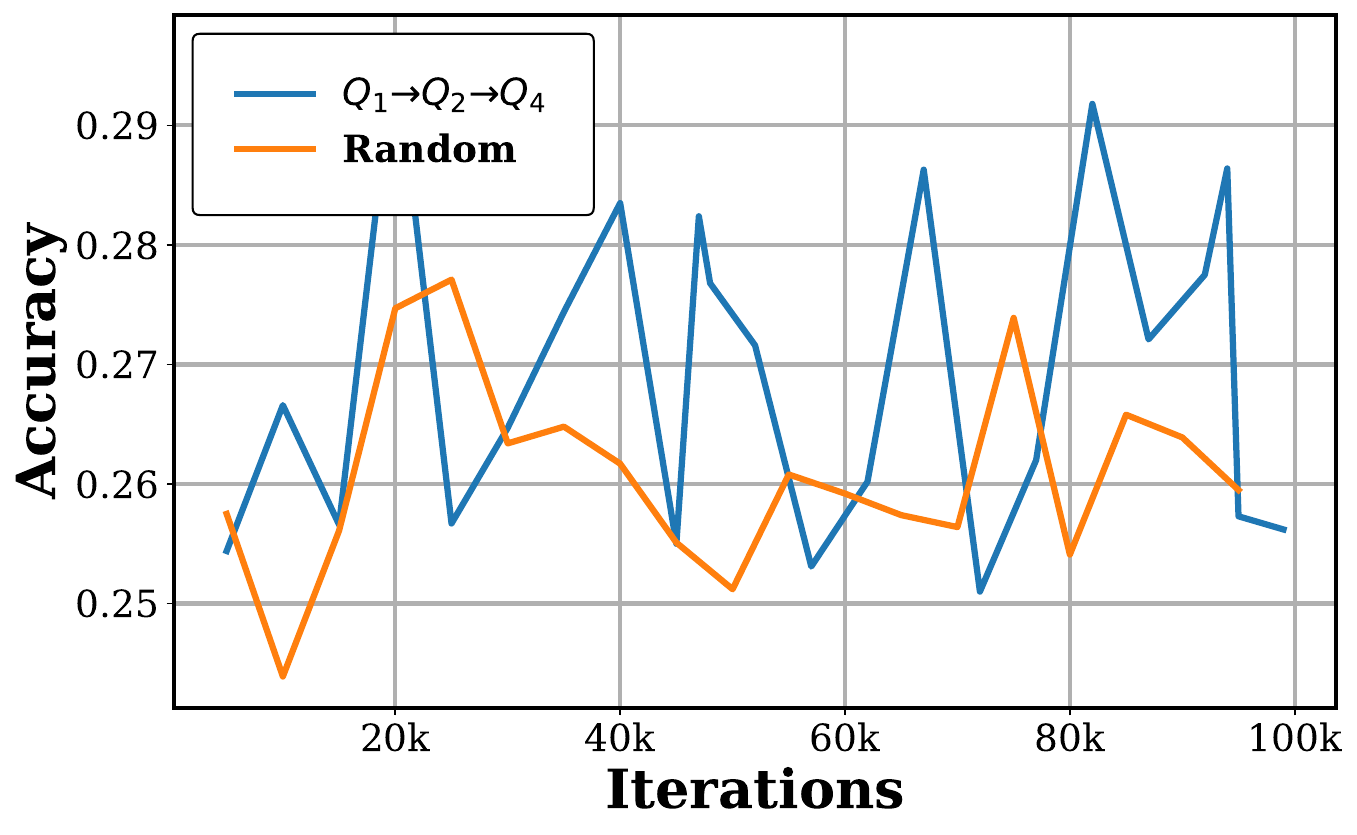}
        \caption{CEVAL}
        \label{fig:PPL ablation study2_4}
    \end{subfigure}
    \hfill
    \begin{subfigure}{0.48\linewidth}
        \centering
        \includegraphics[width=\linewidth]{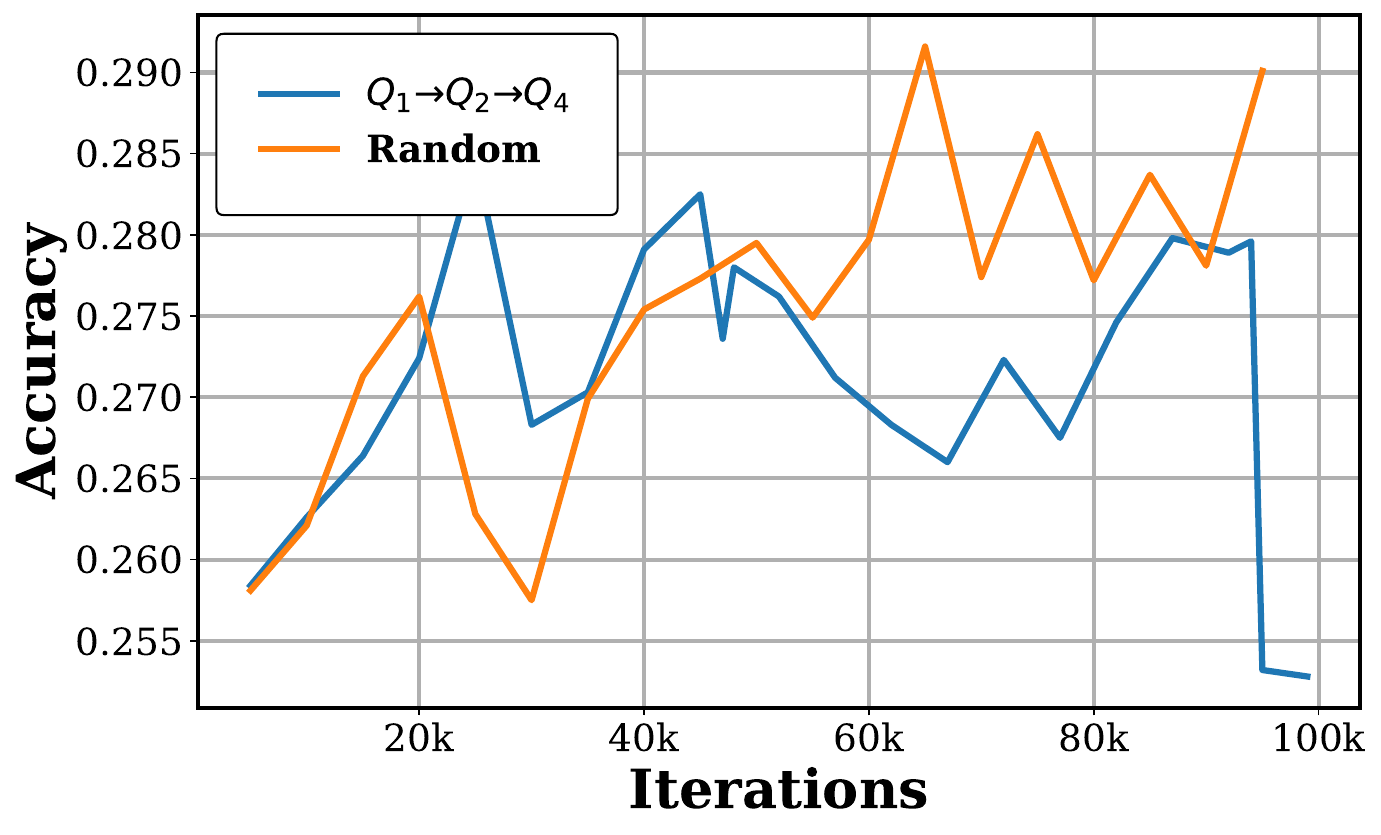}
        \caption{BBH}
        \label{fig:PPL ablation study2_1}
    \end{subfigure}
    \caption{PPL related ablation study (3 quadrants).}
    \label{fig:PPL ablation study2}
\end{figure*}

\begin{figure*}[t]
    % \centering
    \begin{subfigure}{0.48\linewidth}
        \centering
        \includegraphics[width=\linewidth]{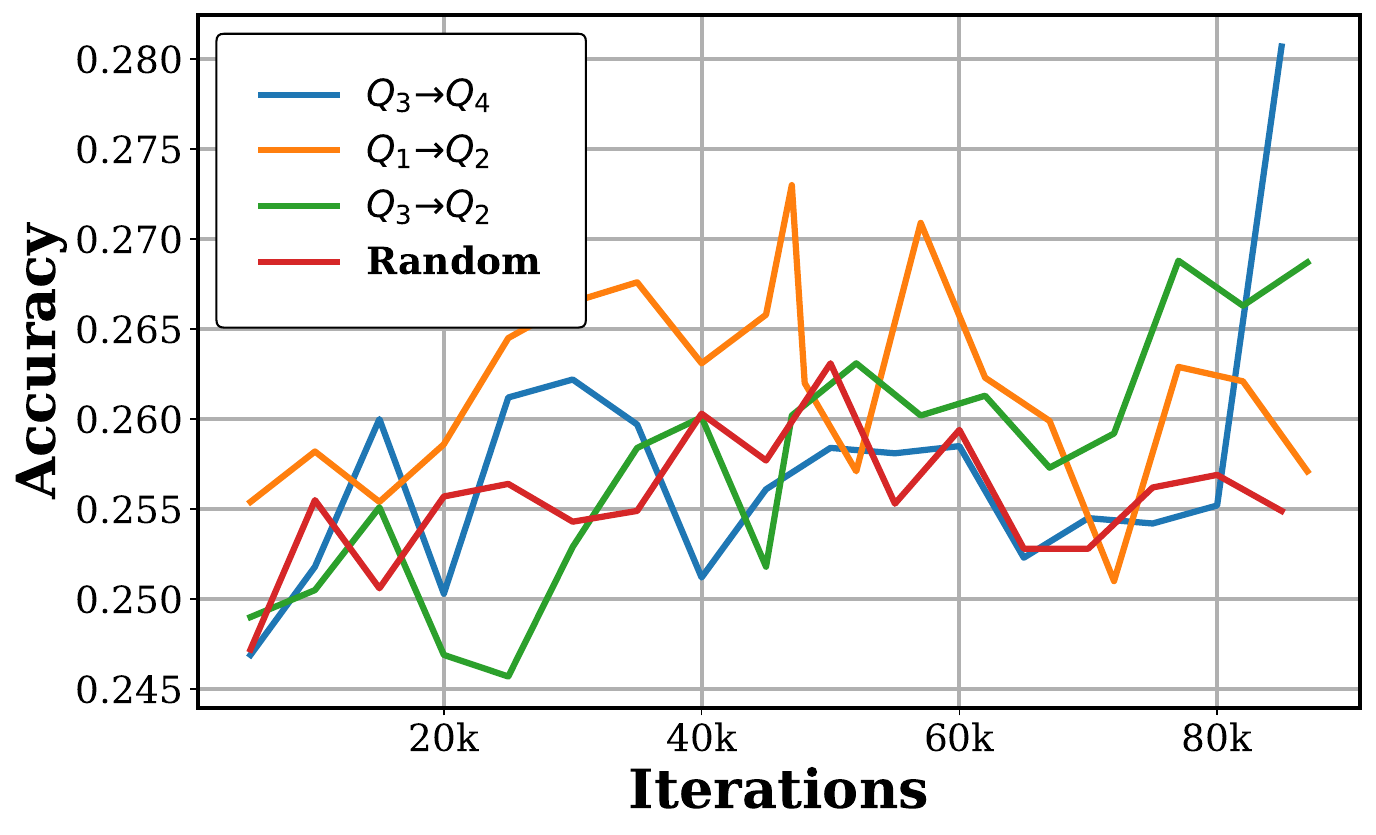}
        \caption{MMLU}
        \label{fig:PD ablation study1_2}
    \end{subfigure}
    \hfill
    \begin{subfigure}{0.48\linewidth}
        \centering
        \includegraphics[width=\linewidth]{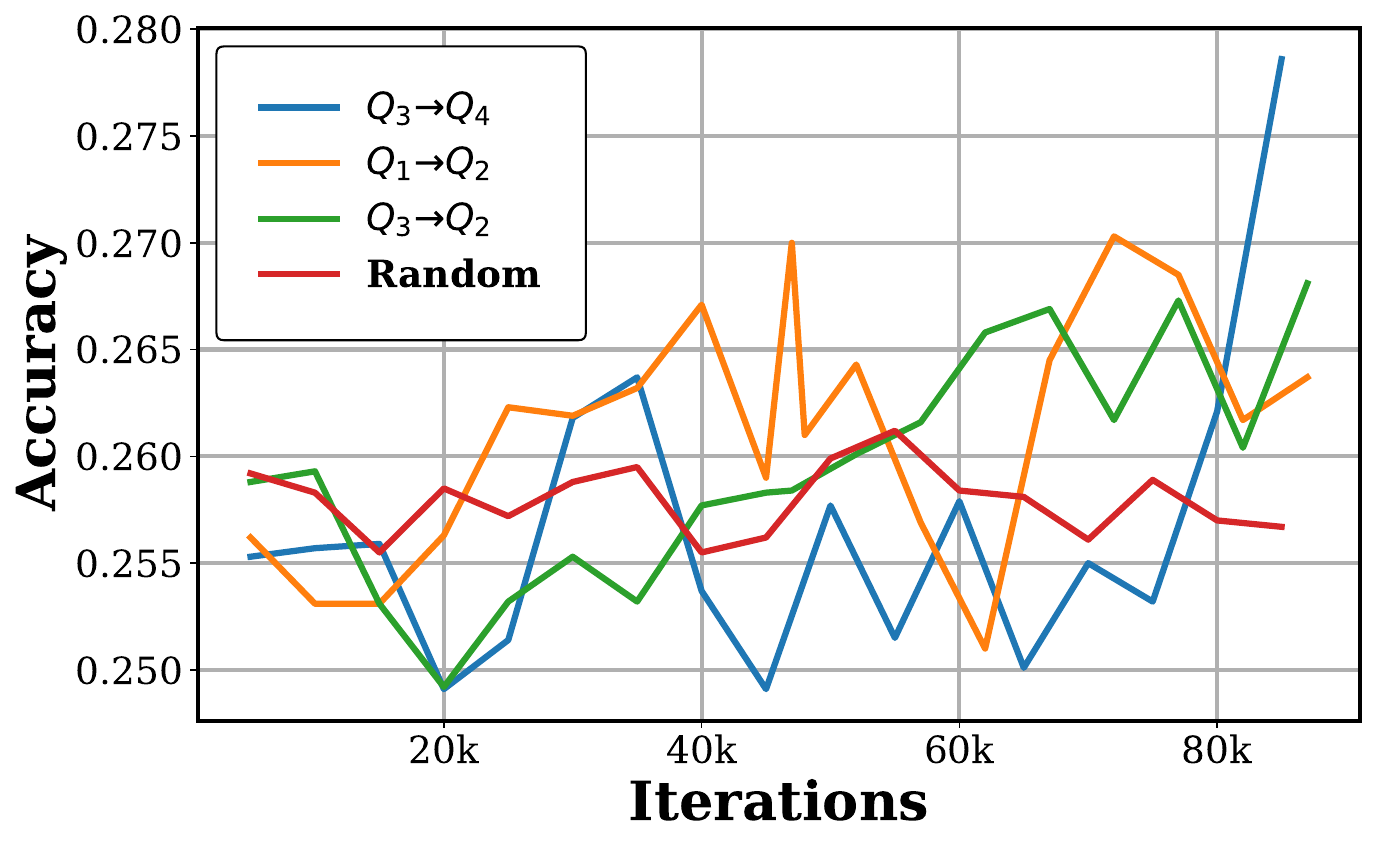}
        \caption{CMMLU}
        \label{fig:PD ablation study1_3}
    \end{subfigure}
    \\
    \begin{subfigure}{0.48\linewidth}
        \centering
        \includegraphics[width=\linewidth]{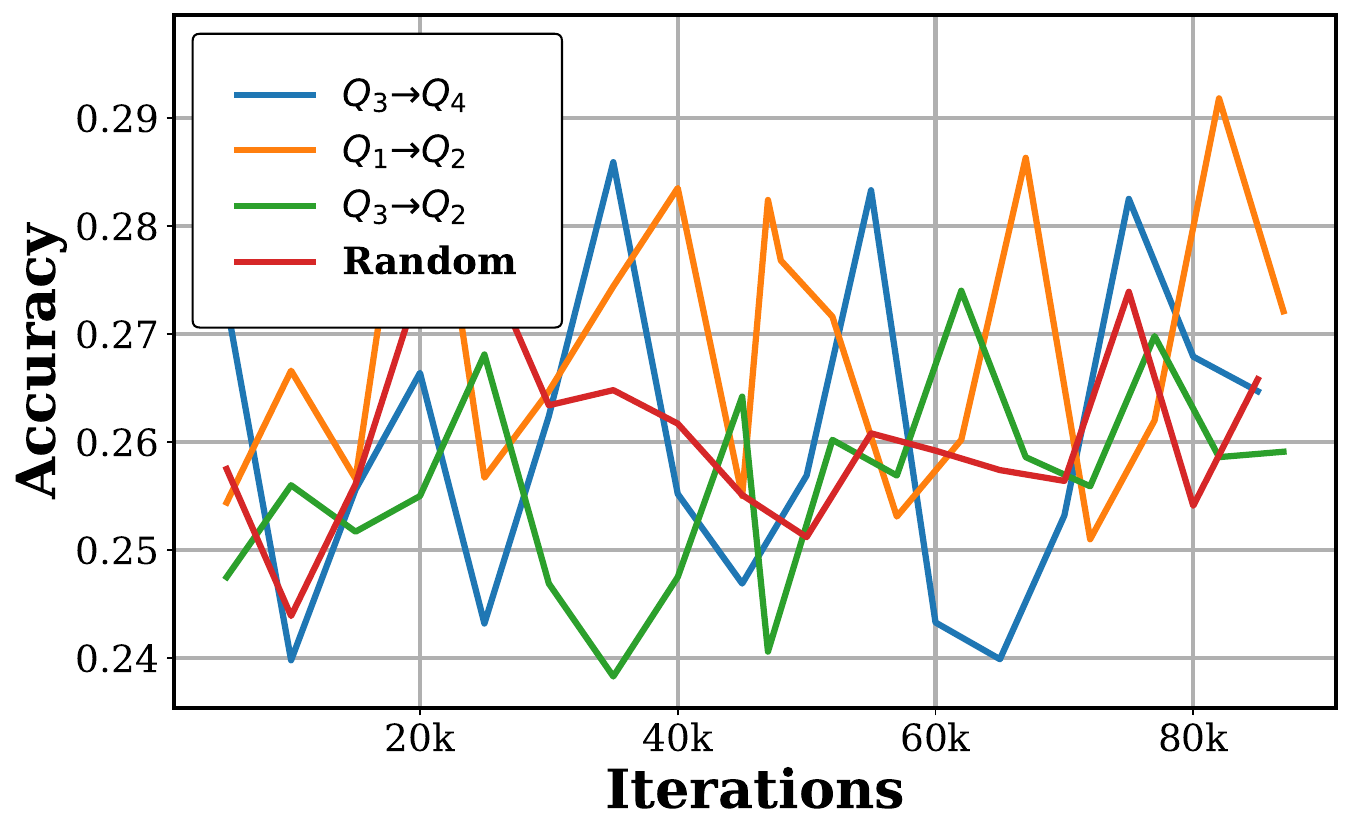}
        \caption{CEVAL}
        \label{fig:PD ablation study1_4}
    \end{subfigure}
    \hfill
    \begin{subfigure}{0.48\linewidth}
        \centering
        \includegraphics[width=\linewidth]{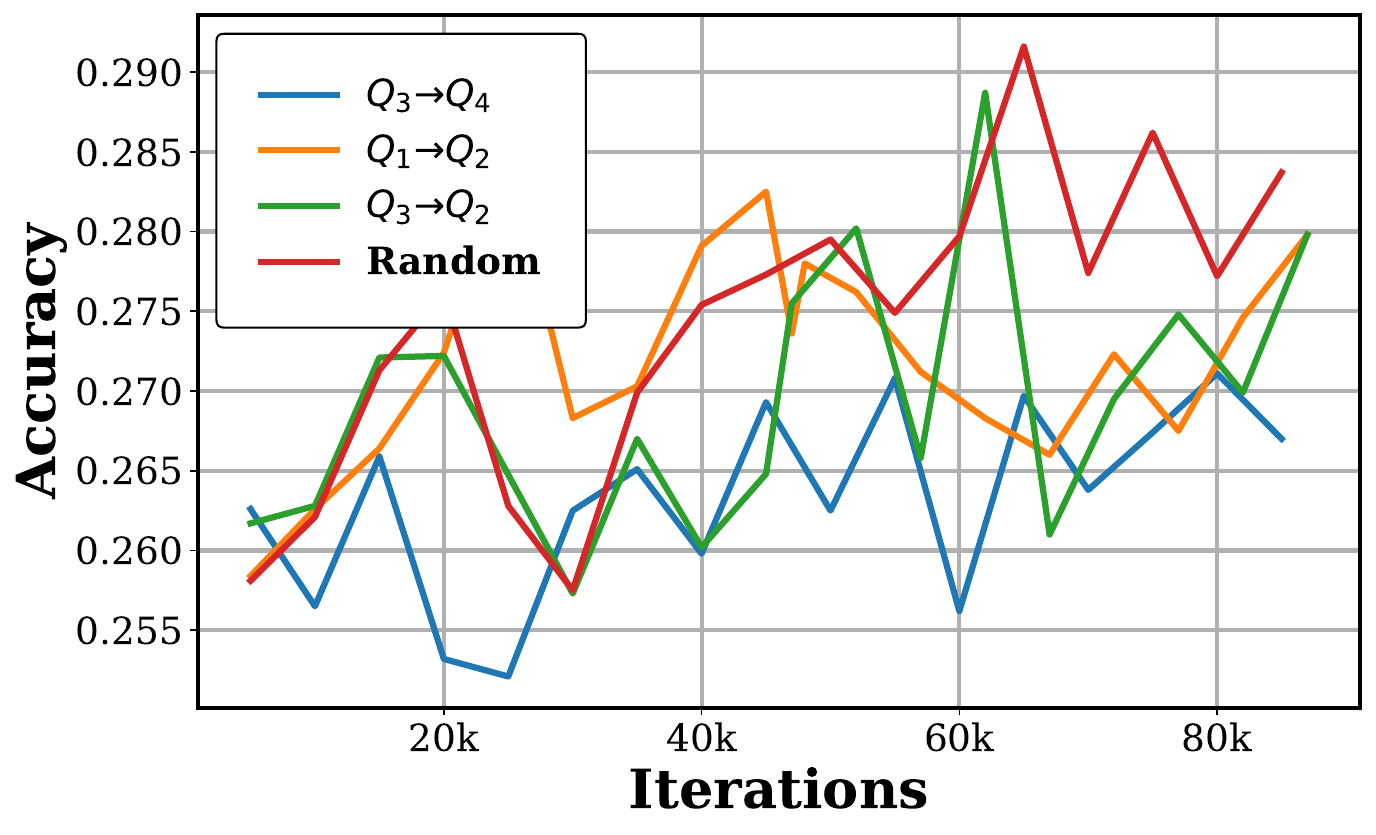}
        \caption{BBH}
        \label{fig:PD ablation study1_1}
    \end{subfigure}
    \caption{PD related ablation study (2 quadrants).}
    \label{fig:PD ablation study1}
\end{figure*}

\begin{figure*}[b]
    % \centering
    \begin{subfigure}{0.48\linewidth}
        \centering
        \includegraphics[width=\linewidth]{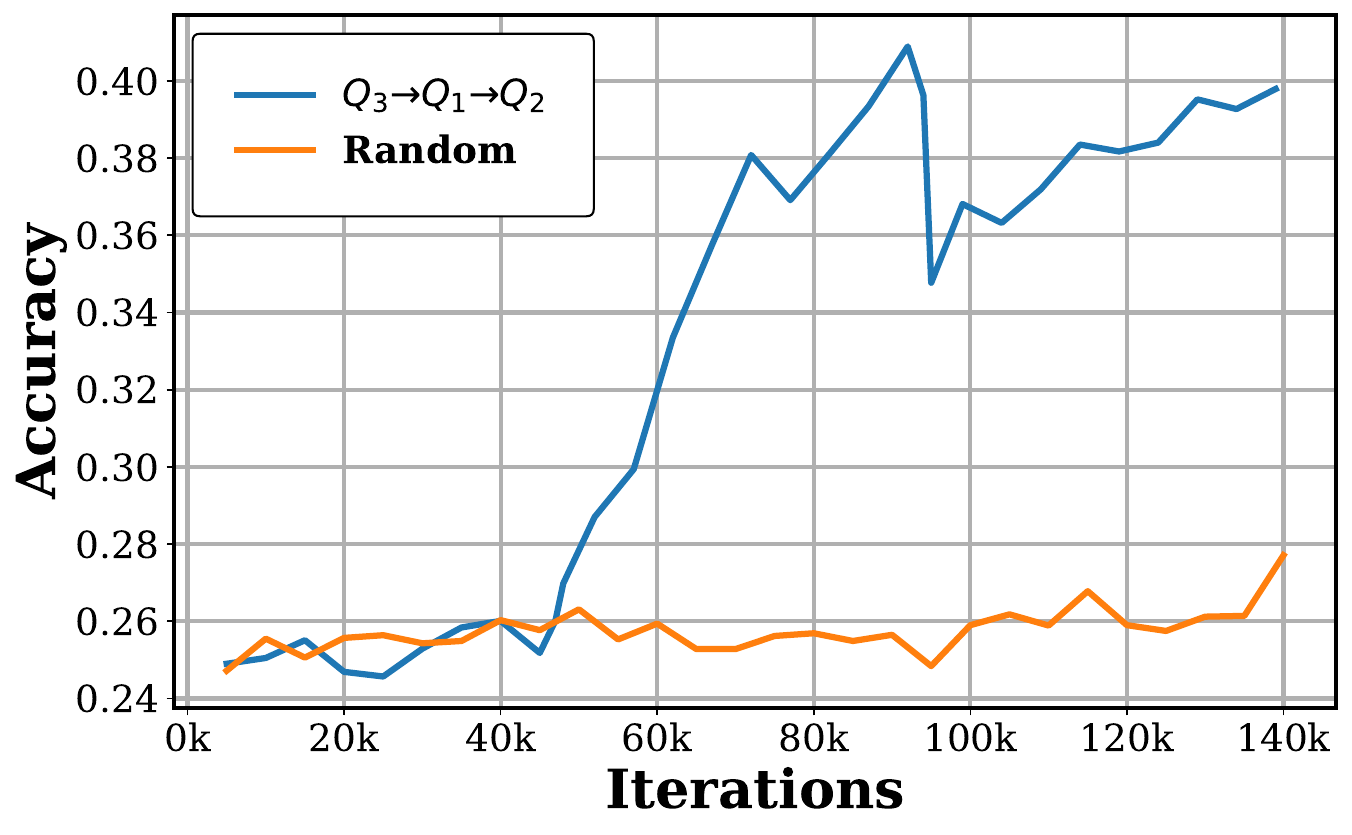}
        \caption{MMLU}
        \label{fig:PD ablation study2_2}
    \end{subfigure}
    \hfill
    \begin{subfigure}{0.48\linewidth}
        \centering
        \includegraphics[width=\linewidth]{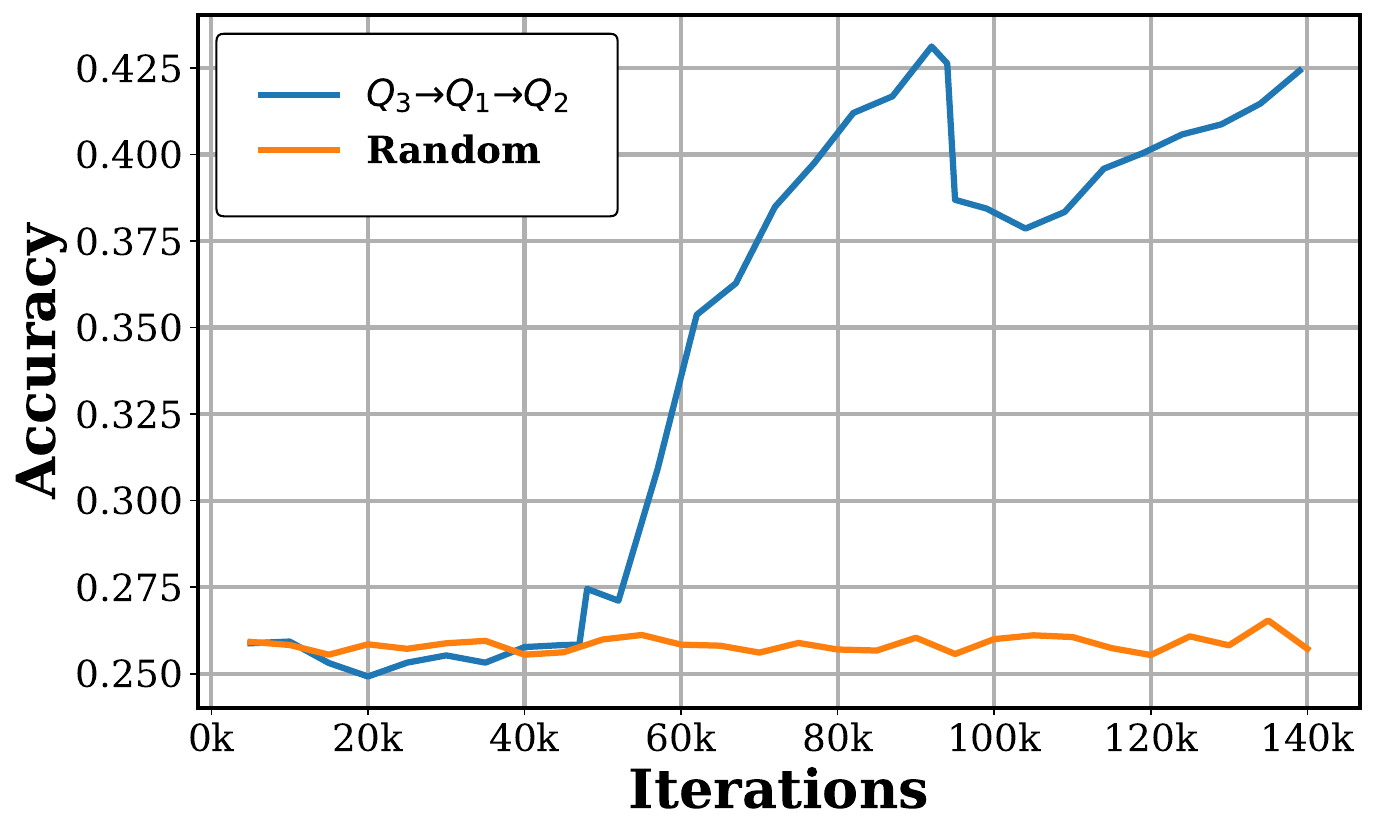}
        \caption{CMMLU}
        \label{fig:PD ablation study2_3}
    \end{subfigure}
    \\
    \begin{subfigure}{0.48\linewidth}
        \centering
        \includegraphics[width=\linewidth]{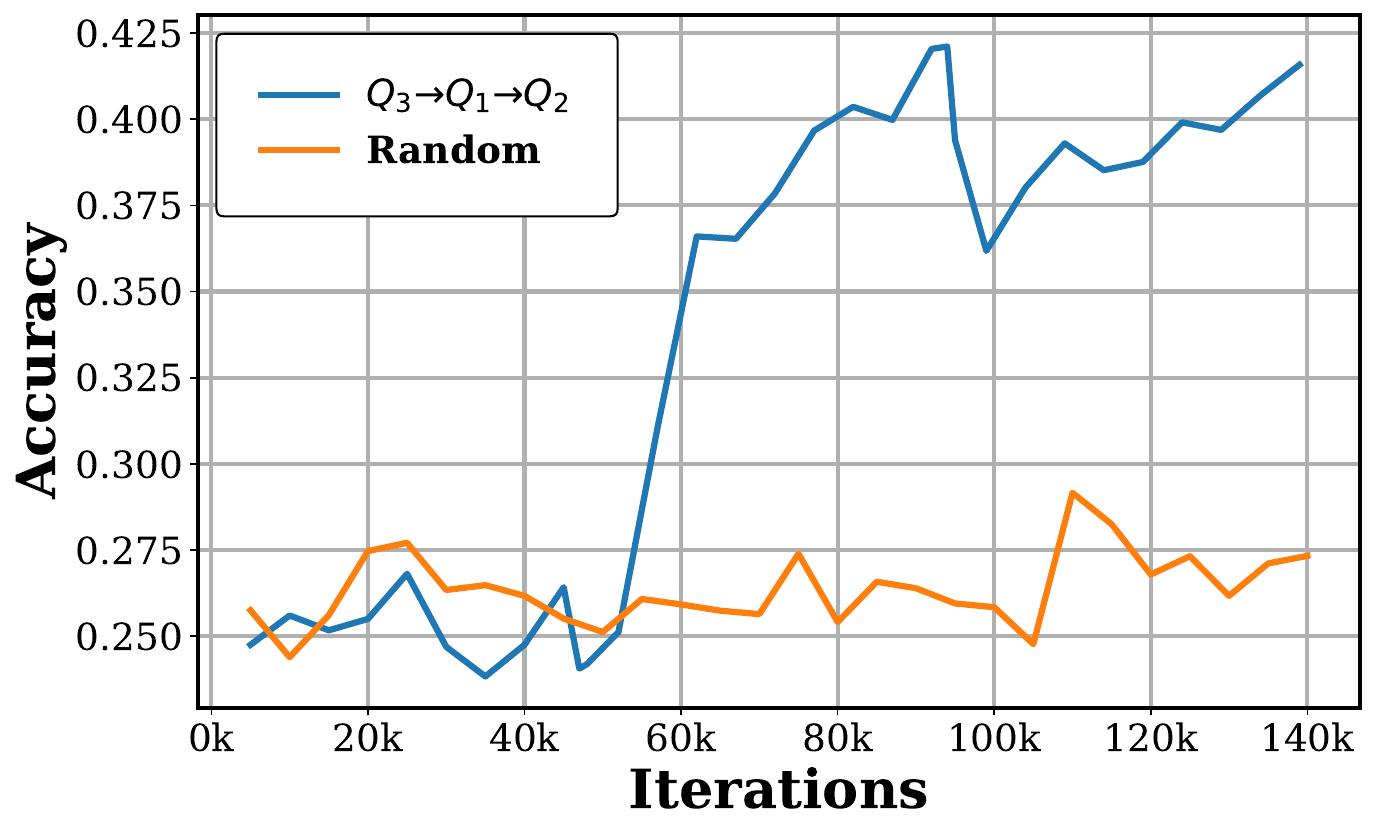}
        \caption{CEVAL}
        \label{fig:PD ablation study2_4}
    \end{subfigure}
    \hfill
    \begin{subfigure}{0.48\linewidth}
        \centering
        \includegraphics[width=\linewidth]{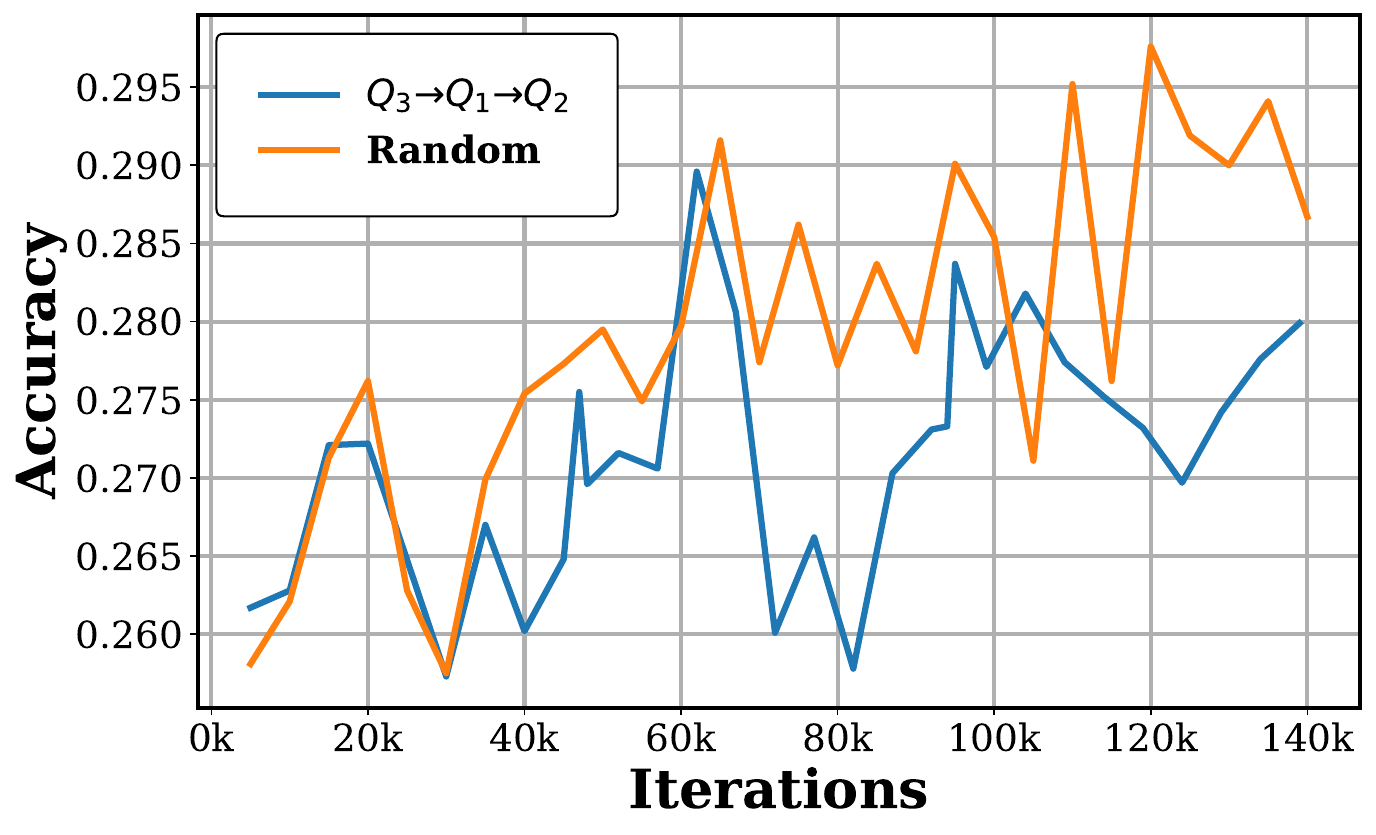}
        \caption{BBH}
        \label{fig:PD ablation study2_1}
    \end{subfigure}
    \caption{PD related ablation study (3 quadrants).}
    \label{fig:PD ablation study2}
\end{figure*}

In Figures \ref{fig:PPL ablation study1}, \ref{fig:PPL ablation study2}, \ref{fig:PD ablation study1} and \ref{fig:PD ablation study2}, we provide a more detailed presentation of the performance of our ablation experiments across various subsets.

\paragraph{PPL Ordering} During two-stage training, we observed that training in the order from high PPL to low PPL enhances the model's emergent capabilities. This training strategy shows significant advantages across multiple datasets, such as MMLU and CMMLU. Following the training path from $Q_3$ to $Q_1$ can significantly improve accuracy on benchmarks. However, if the opposite training order is adopted (i.e., from $Q_1$ to $Q_3$ or from $Q_2$ to $Q_4$), the model's performance is similar to that of a randomly initialized model, and in some cases, even worse. This indicates that the impact of training order on model performance is asymmetric, highlighting the importance of properly arranging the training stages.

In three-stage training, the same trend persists. Specifically, after completing training from $Q_1$ to $Q_2$ and then switching to $Q_4$, following the order of data PPL from small to large, the model's accuracy on multiple benchmarks is even lower than that of Random.

\paragraph{PD Ordering} From the perspective of two-stage training, the training order from small PD to large PD is beneficial for model training. We found that whether in the same PPL region (e.g., $Q_3$ to $Q_4$ or $Q_1$ to $Q_2$) or between different PPL regions (e.g., $Q_3$ to $Q_2$), the model can ultimately achieve performance superior to that of Random.

This conclusion is also applicable in three-stage training. During the transition from the second stage ($Q_1$) to the third stage ($Q_2$) (Figure \ref{fig:PD ablation study2}), the model's final accuracy still improved. This further supports the effectiveness of the small PD to large PD order, emphasizing the importance of properly arranging training steps in multi-stage training.

% \onecolumn


\clearpage

\onecolumn
\section{Prompts for Case Study}
\label{sec:prompt}
The prompt used in Section \ref{subsec:analysis} to analyze the linguistic features of data across different qhadrants is as follows.

% \onecolumn
\begin{tcolorbox}[colback=white!95!gray,colframe=gray!50!black,rounded corners,label={prompt-dot}, title={Prompts for Property Recognition}]
\begin{lstlisting}[breaklines=true, xleftmargin=0pt, breakindent=0pt, columns=fullflexible, mathescape, numbers=none]
You are a language model training data annotator. Your task is to identify whether the given text possesses the following characteristic: {Property}

The text to be annotated is:
{text}

Please determine whether the given text possesses this characteristic according to the above rules. 

The output format should be "Because..., my answer is 'X'." where X must be either "yes" or "no." 

You should remain objective and refrain from adding any further comments after making your choice.
\end{lstlisting}
\end{tcolorbox}

The property comes from the following rules:
\begin{enumerate}
    \item Does the text contain polysemous words? Polysemous words may make understanding more difficult.
    \item Does the text use specialized terminology? Specialized terminology may require specific domain knowledge to understand.
    \item Does understanding the text require specific cultural background knowledge? Cultural background dependence may increase the complexity of understanding.
    \item Does the text require logical reasoning to understand? Logical reasoning adds depth to understanding.
    \item Does the text contain elements of humor? Humor may affect the way the text is understood.
    \item Does the text explore ethical or moral issues? This may increase the depth of thought.
    \item Does the text use complex sentence structures? Complex sentence structures may increase the difficulty of understanding.
    \item Does the text contain scientific or technical concepts? These concepts may require specific knowledge to understand.
    \item Does the text express obvious emotional tones? Emotional tones may affect the understanding of the text.
    \item Does understanding the text require additional background knowledge? Background knowledge requirements may affect the comprehensibility of the text.
\end{enumerate}

\onecolumn
\section{Data Cases}
Table \ref{PD_quantiles} presents samples extracted from each quadrant. 
% The data in the $Q_1$ quadrant primarily consists of samples with low PPL and low PD. These samples exhibit homogeneity and a high degree of duplication. The $Q_2$ quadrant mainly contains data with low PPL but high PD, featuring text expressions that conform to standard written language conventions, with samples being relatively longer and more complex. The data in $Q_3$ and $Q_4$, with high PPL and a mix of low and high PD, often feature creative, informal, or unexpected language that doesn't follow typical grammar rules or standard phrasing, differing significantly from usual writing or speaking styles.

% \renewcommand{\arraystretch}{0.5} % 设置行间距为1.5倍
\begin{longtable}{p{15.5cm}}
\toprule
\midrule
\endfirsthead

\caption[]{(continued)} \\
\toprule
\midrule
\endhead

\midrule
\endfoot

% \bottomrule
\endlastfoot
\textbf{Quadrant 1} \\
\midrule
\small \textit{\textbf{Sample 1}:} ... 112. Ra4+ Kd5 113. Ra5+ Kc4 114. Ra6 Qd7+ 115. Kb6 Kb4 116. Ra7 Qd6+ 117. Kb7 Kb5 118. Kc8 Qf8+ 119. Kb7 Qd8 120. Ra1 Qe7+ 121. Kc8 Qe8+ 122. Kb7 Qe4+ 123. Kc8 Qc4+ 124. Kb7 Qf7+ 125. Kc8 Qg8+ 0-1 Analysis of Utegaliyev - Goh from the Baku Olympiad - Life is tough! ***Addendum: The Rook \& Bishop vs Rook ending at move 102 was covered specifically by Karsen Muller \& Frank Lamprecht on page 301 of Fundamental Chess Endings. My annotations are slightly confusing there ... \\
\small \textit{\textbf{Sample 2}:} ... Women With an Alcoholic Parent Have More Risk Factors Ray Kachatorian / Photographer's Choice / Getty Images There are differences in how parental alcoholism affects daughters as opposed to how it affects sons, particularly when it comes to psychopathology, or mental health disorders, in each gender. Daughters of alcoholics are affected by a parent's alcoholism in many of the same ways that sons are. Both are at higher risk of developing alcohol abuse disorders compared to children of non-alcoholic parents. But there are some differences in how women are influenced, scientists say ... \\
\midrule
\textbf{Quadrant 2} \\
\midrule
\small \textit{\textbf{Sample 1}:} ... pension advice be sought in the case of South Africa. Residents in South Africa are subject to tax (up to 40\%) on their total income no matter the source-country. Foreign pensions are exempt from this. Any UK pension, as long as it was not 'received or accrued' from or in South Africa, and is 'in consideration of past employment' elsewhere, is exempt from income tax. This specifically excludes anyone whose pension has arisen from time in public office in South Africa. For residents of South Africa, ... \\
\small \textit{\textbf{Sample 2}:} ... , Bengaluru (As Prepared for Delivery) Honorable President Pranab Mukherjee, Ministers, Excellencies, Distinguished guests: Thank you, it is such a great honor to be here today, accepting this award among such distinguished company. Let me first thank His Excellency the Honorable President Mukherjee, Prime Minister Modi, and the Government of India\u2013 \u2026 By U.S. Mission India | 9 January, 2017 | Topics: Chennai, Press Releases, Speeches | Tags: Bilateral relationship Remarks by Ambassador Richard R. Verma at the inaugural plenary of Indo-Asia Connectivity for Shared Prosperity ... \\
\midrule
\textbf{Quadrant 3} \\
\midrule
\small \textit{\textbf{Sample 1}:} ... information may be obtained from CPO. The Anti-Kickback Act of 1986 (41.U.S.C.51-58) was passed to deter subcontractors from making payments and contractors from accepting payments for the purpose of improperly obtaining or rewarding favorable treatment in connection with a prime contract or subcontract. Imposes criminal penalties on any person who knowingly and willfully engages in the prohibited conduct addressed in paragraph (a) of this subsection. Provides for the recovery of civil penalties by the United States from any person who knowingly engages in such prohibited conduct and from any person whose employee, subcontractor, or subcontractor employee provides, accepts, or charges a kickback ... \\
\small \textit{\textbf{Sample 2}:} ... also wrote Brave New World and numerous essays. Consider some of his words as follows: \"We are in a process of developing a whole series of techniques which will enable the controlling oligarchy to get people to actually love their servitude. A really efficient totalitarian state would be one in which the all-powerful executive of political bosses and their army of managers control a population of slaves who do not have to be coerced, because they love their servitude. To make them love it is the task assigned, in present-day totalitarian states, to ministries of propaganda ... \\
\midrule
\textbf{Quadrant 4} \\
\midrule
\small \textit{\textbf{Sample 1}:} ... in which 3000 people died in a single atrocity, to one's horror at the deaths of ten and perhaps sometimes twenty times as many in each of the bombings of such places as Hamburg, Dresden, Tokyo, Hiroshima and Nagasaki. This way of grasping the purport of what area bombing meant, really meant, is vital to making a difference to how we behave and what we accept today in the conduct of conflicts. There is nothing abstract or theoretical about the mass murder in which bombing consists: it is real and terrible, and anything that drives the point home has its place in the debate ... \\
\small \textit{\textbf{Sample 2}:} ... and China in Africa. Ba'ath Party Dominance and Mistakes. Iraq. A Fragile Country. Burkina Faso. The Church under attack. DR Congo. Insecurity and bad governance contribute to the spread of Ebola. Drug Trafficking. RD.Congo. Beatification of twenty martyr missionaries on track. Africa. Start-ups rolling out. Between Maras and Pandilla El Salvador. A Country In Transition. Nigeria. Troubles in Kano. A New President. The Church of Africa. A Return To Its Origins. Mexico. \"Until dignity becomes the custom\". Philippine. Modern Day Missionaries of the World. The Gumuz People, their culture. The Religious Universe of the Gumuz. The Gumuz in front of a changing world. Africa. The art of food. Ancient flavours and genuine ingredients ... \\
\bottomrule
\caption{Samples from different quadrants.}
\label{PD_quantiles} \\
\end{longtable}

\end{document}